\documentclass{article}

\usepackage[final]{neurips_2025}

\usepackage[utf8]{inputenc} 
\usepackage[T1]{fontenc}    
\usepackage{hyperref}       
\usepackage{url}            
\usepackage{booktabs}       
\usepackage{amsfonts}       
\usepackage{nicefrac}       
\usepackage{microtype}      
\usepackage{xcolor}         

\usepackage{lipsum}
\usepackage{amsmath}
\usepackage{amsthm}
\usepackage{amssymb}   

\usepackage{algorithm}
\usepackage{algpseudocode}
\usepackage{wrapfig}
\usepackage{siunitx }
\usepackage{bigints}
\usepackage{bm}
\usepackage{diagbox}
\usepackage{makecell}
\usepackage{multirow}
\usepackage{caption}
\usepackage{subcaption}
\usepackage{graphicx}

\usepackage{mathtools}
\usepackage{enumitem}

\usepackage{pifont}
\newcommand{\cmark}{\text{\ding{51}}}
\newcommand{\xmark}{\text{\ding{55}}}

\usepackage[capitalise]{cleveref}

\newcommand{\R}{\mathbb{R}}

\newcommand{\x}{\mathbf{x}}

\def \ba {\mathbf{a}}
\def \bw {\mathbf{w}}

\def \si {\sigma}
\newcommand{\eps}{\epsilon}

\newcommand{\dd}{\cdot}

\def \ep {\varepsilon}

\def \mc {\mathcal}
\newcommand{\norm}[1]{{\left\lVert #1 \right\lVert}}
\def \rd {{\rm d}}

\def \q {\quad}
\newcommand{\na}{\nabla}

\IfFileExists{mathabx.sty}%
  {\DeclareFontFamily{U}{mathx}{\hyphenchar\font45}%
   \DeclareFontShape{U}{mathx}{m}{n}{<->mathx10}{}%
   \DeclareSymbolFont{mathx}{U}{mathx}{m}{n}%
   \DeclareFontSubstitution{U}{mathx}{m}{n}%
   \DeclareMathAccent{\widebar}{0}{mathx}{"73}%
}{%
  \PackageWarning{mathabx}{%
    Package mathabx not available, therefore\MessageBreak substituting
    widebar with overline\MessageBreak }%
  \newcommand{\widebar}[1]{\overline{#1}}%
}
\newcommand{\wb}[1]{\widebar{#1}}
\DeclareMathOperator{\sgn}{sgn}

\def \Vec {{\rm vec}}

\newtheorem{theorem}{Theorem}
\newtheorem{definition}{Definition}
\newtheorem{corollary}{Corollary}
\newtheorem{lemma}{Lemma}
\newtheorem{proposition}{Proposition}
\newtheorem{assumption}{Assumption}

\newtheorem{remark}{Remark}

\numberwithin{equation}{section}

\title{Gradient Alignment in Physics-informed Neural Networks: A Second-Order Optimization Perspective}

\author{
Sifan Wang$^{1}$\thanks{These authors contributed equally to this work.
}, Ananyae Kumar Bhartari$^{2*}$, Bowen Li$^{3*}$, Paris Perdikaris$^{4}$ \\
$^{1}$Institution for Foundation of Data Science, Yale University \\
$^{2}$Penn Institute for Computational Science, University of Pennsylvania \\
$^{3}$Department of Mathematics, City University of Hong Kong \\
$^{4}$Department of Mechanical Engineering and Applied Mechanics, University of Pennsylvania  \\
\texttt{sifan.wang@yale.edu},  \texttt{bowen.li@cityu.edu.hk},  
\texttt{\{ananyaeb, pgp\}@seas.upenn.edu}
}

\begin{document}

\maketitle

\begin{abstract}
Physics-informed neural networks (PINNs) have shown significant promise in computational science and engineering, yet they often face optimization challenges and limited accuracy. In this work, we identify directional gradient conflicts during PINN training as a critical bottleneck. We introduce a novel gradient alignment score to systematically diagnose this issue through both theoretical analysis and empirical experiments.
Building on these insights, we show that (quasi) second-order optimization methods inherently mitigate gradient conflicts, thereby consistently outperforming the widely used Adam optimizer. Among them, we highlight the effectiveness of SOAP \cite{vyas2024soap} by establishing its connection to Newton’s method.
Empirically, SOAP achieves state-of-the-art results on 10 challenging PDE benchmarks, including the first successful application of PINNs to turbulent flows at Reynolds numbers up to 10,000. It yields 2–10x accuracy improvements over existing methods while maintaining computational scalability, advancing the frontier of neural PDE solvers for real-world, multi-scale physical systems. All code and datasets used in this work are publicly available at:  
\url{https://github.com/PredictiveIntelligenceLab/jaxpi/tree/pirate}
\end{abstract}

\section{Introduction}
\vspace{-1mm} 

Physics-informed neural networks (PINNs) have emerged as a powerful paradigm in scientific machine learning by incorporating physical principles through carefully designed loss functions. These loss functions act as soft constraints, guiding neural networks to learn solutions that respect underlying physical laws while simultaneously fitting experimental data. The elegance and versatility of PINNs have led to their widespread adoption in solving both forward and inverse problems involving partial differential equations (PDEs). Their success spans numerous domains in computational science, from fluid mechanics \cite{raissi2020hidden,almajid2022prediction,eivazi2022physics,cao2024surrogate}, heat transfer \cite{xu2023physics,bararnia2022application,gokhale2022physics} to bio-engineering \cite{kissas2020machine,zhang2023physics,caforio2024physics} and materials science \cite{zhang2022analyses,jeong2023physics, hu2024physics}. The impact of PINNs extends even further, with significant applications in electromagnetics \cite{kovacs2022conditional,khan2022physics,baldan2023physics}, geosciences \cite{smith2022hyposvi, song2023simulating,ren2024seismicnet}, etc.

Despite their broad applications, PINNs currently face limitations in convergence speed and accuracy that affect their reliability as forward PDE solvers. This has motivated extensive research efforts to enhance their performance through various methodological innovations. Significant advances have emerged in neural architecture design, including novel network backbones \cite{wang2021understanding,sitzmann2020implicit,fathony2021multiplicative,moseley2021finite,kang2022pixel,cho2024separable,wang2024piratenets}, improved activation functions \cite{jagtap2020adaptive,abbasi2024physical}, and effective coordinate embeddings \cite{wang2021eigenvector,costabal2024delta,zeng2024rbf,huang2024efficient}.
Other improvements have focused on optimizing the training process through enhanced collocation point sampling strategies \cite{nabian2021efficient,daw2022rethinking,wu2023comprehensive}, more efficient optimizers \cite{muller2023achieving,jnini2024gauss,song2024admm,urban2025unveiling}, and advanced training strategies such as sequential training \cite{wight2020solving,krishnapriyan2021characterizing,cao2023tsonn} and transfer learning \cite{desai2021one,goswami2020transfer,chakraborty2021transfer}.
Researchers have also explored alternative formulations of the learning objective, incorporating numerical differentiation \cite{chiu2022can}, variational principles inspired by Finite Element Methods \cite{kharazmi2021hp,patel2022thermodynamically}, and specialized regularization terms \cite{yu2022gradient,son2021sobolev}.

A particularly active area of research has centered on addressing gradient pathologies during training \cite{wang2021understanding,wang2022and}. One prominent issue involves imbalanced backpropagated gradients across different loss terms, leading to significant discrepancies in convergence rates and reduced solution accuracy, especially in complex physical systems. This has led to the development of various adaptive weighting strategies \cite{wang2021understanding,wang2022and,li2022revisiting,chen2024self,anagnostopoulos2024residual,liu2024discontinuity,song2025vw}. However, the equally critical issue of directional gradient conflicts -- where gradients from different losses point in conflicting directions -- remains largely underexplored \cite{liu2024config,hwang2024dual}.
Our work aims to bridge this gap by systematically investigating, analyzing, and resolving these directional conflicts in PINNs training. The key contributions of this work are summarized as follows:

\begin{itemize}[itemsep=1mm, topsep=1mm, parsep=1mm, leftmargin=*]
    \item We introduce a novel gradient alignment metric that extends cosine similarity to quantify directional conflicts between multiple loss terms.
    \item We demonstrate that gradient conflicts hinder PINN training, with higher conflict scores linked to slower convergence in various PDE systems.
    \item We show that second-order optimizers enhance gradient alignment by implicitly preconditioning the loss landscape.
    \item We reveal that  SOAP \cite{vyas2024soap} can be viewed as an efficient approximation of the Newton preconditioner, revolving gradient conflicts.
    \item We provide comprehensive experimental results across 10 PDE benchmarks, including the first successful PINN application to turbulent flows with Reynolds numbers up to 10,000, achieving 2-10x accuracy improvements.
\end{itemize}
Taken together, this work advances our understanding of optimization dynamics in PINNs while demonstrating how quasi second-order methods can enable more reliable neural PDE solvers for solving complex physical systems. These insights pave the way for developing next-generation optimizers for physics-informed machine learning, and beyond.

\section{Overview of PINNs}
\vspace{-1mm} 
\label{sec: background}
\label{subsec: pinns}
Multi-task learning in deep neural networks requires simultaneously minimizing multiple competing objectives -- a challenge that manifests acutely in physics-informed neural networks (PINNs). Building upon the work of \cite{raissi2019physics}, PINNs approximate solutions to partial differential equations by minimizing a composite loss function that enforces both physical constraints and data-fitting objectives. Consider a general PDE system: 
\begin{align}
\label{eq: PDE}
     &\mathbf{u}_t +  \mathcal{N}[\mathbf{u}] = 0, \ \  t \in [0, T],  \ \mathbf{x} \in \Omega, 
\end{align}
with inital and boundary conditions 
\begin{align}
 \label{eq: IC}
     &\mathbf{u}( 0, \mathbf{x})=\mathbf{g}(\mathbf{x}), \ \ \mathbf{x} \in \Omega, \\
  \label{eq: BC}
 &\mathcal{B}[\mathbf{u}] = 0,  \ \   t\in [0, T], \  \mathbf{x} \in  \partial \Omega,
\end{align}
where $\mathcal{N}[\cdot]$ represents a differential operator and $\mathcal{B}[\cdot]$ denotes boundary conditions. The core idea of PINNs is to approximate the solution $\mathbf{u}(t, \mathbf{x})$ using a neural network $\mathbf{u}_{\mathbf{\theta}}(t, \mathbf{x})$. Through automatic differentiation \cite{griewank2008evaluating}, we can compute the PDE residual: 
\begin{align}
    \label{eq: pde_residual}
    \mathcal{R}[u_{\mathbf{\theta}}](t, \mathbf{x}) = \frac{\partial \mathbf{u}_{\mathbf{\theta}}}{\partial t}(t_r, \mathbf{x}_r) + \mathcal{N}[\mathbf{u}_{\mathbf{\theta}}](t_r, \mathbf{x}_r).
\end{align} 
This leads to a composite loss function that encapsulates multiple competing objectives:
\begin{align}
\label{eq: PINN_loss}
\mathcal{L}(\theta)=\underbrace{\frac{1}{N_{i c}} \sum_{i=1}^{N_{i c}}\left|\mathbf{u}_\theta\left(0, \mathbf{x}_{i c}^i\right)-\mathbf{g}\left(\mathbf{x}_{i c}^i\right)\right|^2}_{\mathcal{L}_{i c}(\theta)} 
+\underbrace{\frac{1}{N_{b c}} \sum_{i=1}^{N_{b c}}\left|\mathcal{B}\left[\mathbf{u}_\theta\right]\left(t_{b c}^i, \mathbf{x}_{b c}^i\right)\right|^2}_{\mathcal{L}_{b c}(\theta)}
+\underbrace{\frac{1}{N_r} \sum_{i=1}^{N_r}\left|\mathcal{R}\left[\mathbf{u}_\theta\right]\left(t_r^i, \mathbf{x}_r^i\right)\right|^2}_{\mathcal{L}_r(\theta)}.
\end{align}
These loss functions aim to fit the initial, boundary conditions and the PDE residuals, respectively. And 
$\{\mathbf{x}_{ic}^i\}_{i=1}^{N_{ic}}$, $\{t_{bc}^i, \mathbf{x}_{bc}^i\}_{i=1}^{N_{bc}}$ and $\{t_{r}^i, \mathbf{x}_{r}^i\}_{i=1}^{N_{r}}$ may be selected either as fixed mesh vertices or through random sampling during each training iteration.

\section{Gradient Alignment in PINNs} \label{subsec:gradconflict}
\vspace{-1mm} 



\begin{wrapfigure}{r}{0.4\textwidth}
\vspace{-7mm}
  \begin{center}
    \includegraphics[width=0.4\textwidth]{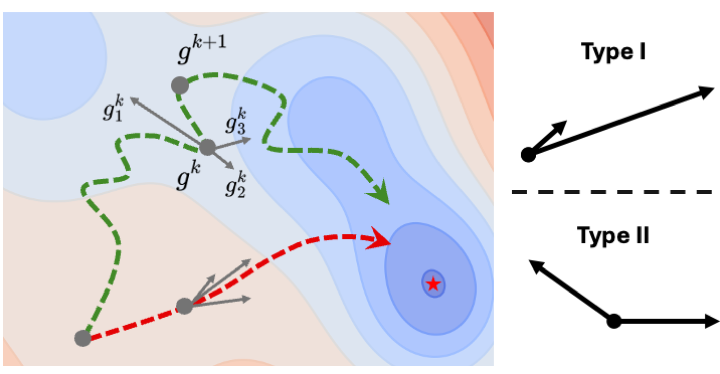}
  \end{center}
  \vspace{-1mm}
\caption{{\small Gradient conflicts and their impact on PINNs optimization. The irregular green trajectory illustrates how the optimization struggles when facing two types of gradient conflicts: Type I, where gradients have similar directions but vastly different magnitudes, and Type II, where gradients have similar magnitudes but opposing directions. The red trajectory shows how appropriate preconditioning through second-order information could mitigate these conflicts by aligning gradients both within and between optimization steps, enabling efficient convergence.}}
  \label{fig:loss_landscape}
\vspace{-5mm}
\end{wrapfigure}

A fundamental challenge in training PINNs is that different loss terms often conflict during optimization. As illustrated in Figure~\ref{fig:loss_landscape}, PINNs encounter two modes of gradient conflict during training. The first, identified by \cite{wang2021understanding,wang2022and}, involves significant imbalances in gradient magnitudes. In such cases, dominant loss terms overwhelm others, often leading to model failure. Various self-adaptive weighting strategies have been proposed to address this issue \cite{li2022revisiting, chen2024self, anagnostopoulos2024residual, liu2024discontinuity, song2025vw}.

This second mode arises when gradients from different loss terms point in conflicting directions, forcing optimization into inefficient, compromised trajectories. 
For example, in the Navier–Stokes equations, enforcing no-slip boundary conditions demands precise control of velocity gradients near walls, which can conflict with preserving mass and momentum conservation in the bulk flow. 
First-order optimizers like gradient descent or Adam must follow the average gradient direction, resulting in inefficient zigzagging between competing objectives. The severity of these conflicts increases with problem complexity, becoming particularly acute for turbulent flows where maintaining physical constraints across multiple scales is crucial. 
To better understand and address these directional gradient conflicts, we introduce the alignment score, defined as follows.

\begin{definition} \label{def:gradalign}
Suppose that $v_1, v_2, \dots, v_n$ are vectors, then the  alignment score is defined as
    \begin{align}
       \mathcal{A}(v_1, v_2, \dots, v_n)=2 \left\|\frac{\sum_{i=1}^n \frac{v_i}{\left\|v_i\right\|}}{n}\right\|^2 - 1.
    \end{align}
\end{definition}
This score ranges from $[-1,1]$ and naturally extends the concept of cosine similarity to multiple vectors. As illustrated in Proposition \ref{prop1}, for the special case of $n=2$, our score exactly recovers the standard cosine similarity $\text{cos}(v_1, v_2) = \frac{v_1 \cdot v_2}{\|v_1\| \|v_2\|}$, where 1 indicates perfect alignment, 0 suggests orthogonal directions, and -1 represents complete opposition.

\begin{proposition}
    \label{prop1}
     For n=2, the alignment score $\mathcal{A}(v_1, v_2)$ equals the cosine similarity between $v_1$ and $v_2$:
\begin{align}
\mathcal{A}(v_1, v_2) = \cos(v_1, v_2) = \frac{v_1 \dd v_2}{\|v_1\|\|v_2\|}\,.
\end{align}
\end{proposition} 
The proof is provided in Appendix \ref{proof: prop1}. The alignment score enables us to quantify both the local conflicts between individual loss terms within each gradient descent step and the global conflicts across consecutive steps. Formally:
\begin{definition}
Let $\mathcal{L} = \sum_{i=1}^n \mathcal{L}_i$ be a composite loss function. At the $k$-th step of gradient descent, let $g^k$ denote the full gradient and $g_1^k, g_2^k, \dots, g_n^k$ denote the gradients of individual loss terms. We define:
\begin{enumerate}[label=(\alph*)]
    \item The intra-step gradient alignment score:     
\begin{align}
\label{eq: intra_align}
\mathcal{A}_{intra}^k = \mathcal{A}(g_1^k, g_2^k, \dots, g_n^k).
\end{align}
\item The inter-step gradient alignment score: 
\begin{align}
\label{eq: inter_align}
\mathcal{A}_{inter}^k = \mathcal{A}(g^{k-1}, g^k).
\end{align} 
\end{enumerate}
\end{definition}
The intra-step gradient alignment score becomes especially useful for PINNs applications where the total loss  typically consists of multiple terms. For example, in the case of solving the 2D Navier-Stokes equations, the loss function includes separate components corresponding to momentum equations in the $x$ and $y$ directions, the continuity equation, and boundary conditions for the velocity fields $u$ and $v$. Since these loss terms can have different scales and properties, they should be treated separately rather than grouped together. Our intra-step score effectively quantifies gradient conflicts across all these terms simultaneously.

The impact of alignment can be formalized through the following result, which shows that the rate of loss decay under preconditioned gradient descent depends on the alignment scores.
\begin{proposition}
\label{prop_alignment}
Let $\mathcal{L}=\sum_{i=1}^n \mathcal{L}_i:\mathbb{R}^d\to\mathbb{R}$ be $L$-smooth (w.r.t. the Euclidean norm). Consider preconditioned gradient descent
\[
\theta_{t+1}=\theta_t-\eta h_t,\qquad h_t=P_t g_t,\qquad g_t=\sum_{i=1}^n g_t^i,
\]
where each $P_t\succ 0$ satisfies $\mu\le \lambda_{\min}(P_t)\le \lambda_{\max}(P_t)\le M$.
Let $\Delta_t:=\mathcal L(\theta_t)-\mathcal L(\theta_{t+1})$ and assume $\eta\le 1/(LM)$. Then:

\begin{enumerate}[label=(\roman*)]
\item \textbf{Single-step drop.} For all $t$, if $\|h_t^i\|=\lambda$ for all $i$,
\[
\Delta_t \;\ge\; \Big(\tfrac{\eta}{M}-\tfrac{L\eta^2}{2}\Big)\,
\frac{n}{2}\big(A_t^{\mathrm{intra}}+1\big)\sum_{i=1}^n \|h_t^i\|^2.
\]
\item \textbf{Two-step cumulative drop.} Let $a:=\|h_{t-1}\|$, $b:=\|h_t\|$. Then
\[
\Delta_{t-1}+\Delta_t \;\ge\;
\tfrac{\eta}{M}\Big(1-\tfrac{L\eta M}{2}\Big)a^2
+\tfrac{\eta}{M}\big(1-L\eta M\big)\,ab\,A_t^{\mathrm{inter}}
-\tfrac{L\eta^2}{2}\,b^2.
\]
\end{enumerate}
\end{proposition}
This result shows that higher intra- and inter-step alignment scores directly accelerate loss reduction. The assumption $\left|h_t^k\right|=\lambda$ for all $k$ is not restrictive in the PINNs setting, where balancing the scales of different losses is common practice. In fact, we adopt the weighting scheme proposed by Wang et al. \cite{wang2021understanding} (Appendix \ref{appendix:training}), which ensures that all weighted gradients have the same norm.

In the following, we will demonstrate that gradient direction conflicts widely exist in training PINNs, especially in the early stages of training.  To this end, we conduct experiments on five representative PDEs spanning from linear wave propagation to reaction-diffusion systems like the Ginzburg-Landau equation and fluid dynamics governed by the Navier-Stokes equations.  The detailed results  are presented in Figure \ref{fig:grad_align_score} and corresponding experimental setup is provided in Appendix \ref{appendix: experiments}.


\begin{figure}[t]
    \centering
    \vspace{-1mm}
    \includegraphics[width=1.0\linewidth]{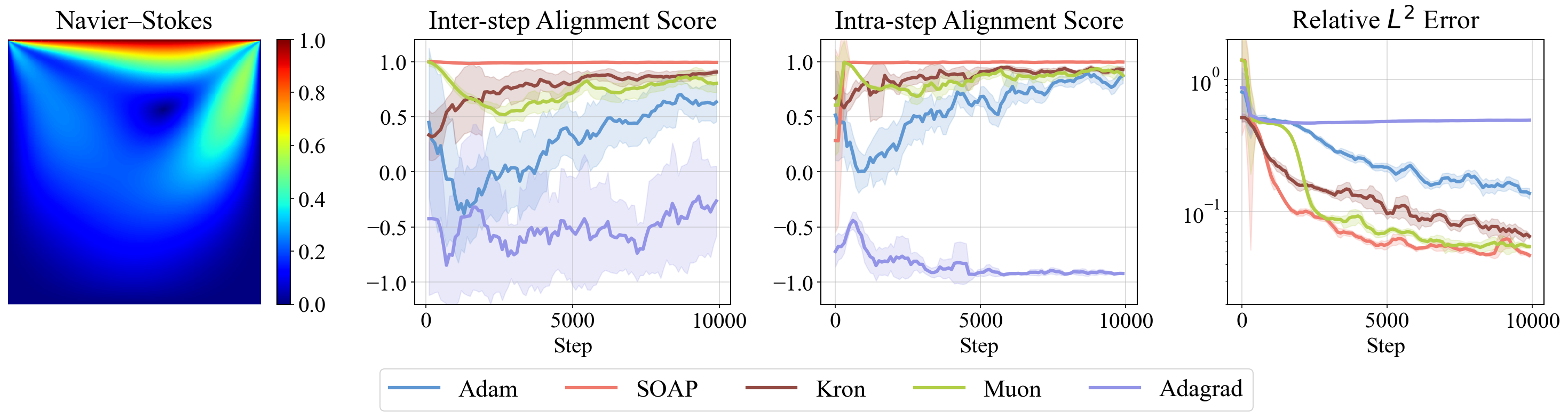}
    \caption{Gradient alignment scores and test errors during PINN training for solving the Navier-Stokes equations with different optimizers.   Additional benchmarks are provided in Figure \ref{fig:grad_align_score}, where we observe the consistent phenomenon that first-order optimizers exhibit poor gradient alignment and slow convergence of test errors.}
    \label{fig:grad_align_score_last_row}
    \vspace{-2mm}
\end{figure}

Figure \ref{fig:grad_align_score_last_row} presents gradient alignment scores and test errors during PINN training for solving the Navier-Stokes equations with different optimizers. 
Importantly, we compute these scores from the gradients after applying each optimizer’s gradient transformations, which captures the actual update directions and their degree of alignment or conflict.
We observe that the scores from first order optimizer oscillate significantly near or below zero at early stages. It provides strong evidence for persistent directional conflicts between gradients throughout the training process. Intuitively, these conflicting gradients force the network parameters to follow an inefficient zigzag trajectory in the loss landscape, significantly impeding convergence speed.

In contrast, the quasi second-order optimizers (e.g., Muon \cite{jordan2024muon}, SOAP \cite{vyas2024soap}, Kron \cite{li2017preconditioned}) consistently maintains the higher positive values for both inter-step and intra-step gradient alignment scores throughout training. This effective resolution of gradient direction conflicts directly corresponds to significantly faster convergence in test error. It is worth nothing that SOAP achieves the highest alignment scores among all optimizers, demonstrating its superior effectiveness.
In the following sections, we provide a theoretical explanation for why PINNs inherently suffer from directional gradient conflicts and establish a formal connection between SOAP and Newton’s method, offering deeper insight into its empirical success.

\section{Intra-step Gradient Alignment of PINNs at Initialization}

In this section, we show that gradient conflicts are intrinsic to PINNs near initialization, regardless of the optimizer used. For simplicity, we analyze the intra-step gradient alignment for a two-layer neural network solving the one-dimensional Laplace equation under small weight initialization. This setup allows for a tractable analysis while capturing key phenomena, and the results naturally extend to more general PDEs.
Following the setup in \cref{subsec: pinns}, and without loss of generality, we consider the 1D Laplace equation:
\begin{align}
    \left\{
\begin{array}{ll}
\Delta u = u'' = 0 & \text{on } [-1,1], \\
u(\pm 1) = g_{\pm 1}. & 
\end{array}
\right.
\end{align}
We approximate the solution $u(x)$ by a two-layer network with width $N$: 
\begin{equation} 
\label{eq:twolayersol}
    u(x, \theta) = \sum_{i = 1}^N a_i \si ( w_i x) = \ba \dd \si(\bw x)\,,
\end{equation}
where $\ba = (a_1, \ldots, a_N), \bw = (w_1, \ldots, w_N)  \in \R^N$, and $\theta = (\ba,\bw) \in \R^{2N}$. Moreover, we limit ourselves to the activation function $\si(x) = \tanh(x)$. In this case, the loss \eqref{eq: PINN_loss} reduces to 
\begin{equation} \label{eq:losseg}
    \min_{\theta = (\ba, \mathbf{w})} \mc{L}(\theta) = \underbrace{\frac{1}{N_r} \sum_{p = 1}^{N_r} |u''(x_p, \theta)|^2}_{\mc{L}_{r}(\theta)} + \underbrace{\frac{1}{2} \sum_{s = \pm 1} |u(s, \theta) - g_{s}|^2}_{\mc{L}_{bc}(\theta)}\,.
\end{equation}
\begin{proposition} 
\label{prop: grad_align}
At initialization, we assume that the weights $a_i, w_i$ are initialized by i.i.d. Gaussian $\mc{N}(0,\ep^2)$ with small $\ep = o(1)$. Then the alignment score converges to a binary random variable in the infinite width limit:
\begin{align}
    \lim_{N \rightarrow \infty }\mathcal{A}(\square(\nabla \mathcal{L}_b),\square(\nabla \mathcal{L}_r)) = O(\ep^2) + C_\square \begin{cases}\operatorname{sgn}\left(g_{-1}-g_1\right) & \text { with prob. } \frac{1}{2}, \\ -\operatorname{sgn}\left(g_{-1}-g_1\right) & \text { with prob. } \frac{1}{2} .\end{cases}
\end{align}
where $\square = \operatorname{GD}, \operatorname{Adam}, \operatorname{Shampoo}, \text{ or } \operatorname{Soap}$  
 denotes the corresponding optimizer update rule, and
$C_\square$ is a constant depending on the optimizer. 
\end{proposition}

The proof is provided in Appendix \ref{app:gradientconflict}.
As a direct implication, the intra-step alignment score exhibits no preferred direction.  Consequently, all optimizers fail to induce consistent intra-step alignment between $\nabla \mathcal{L}_b$ and $\nabla \mathcal{L}_r$ near initialization, as empirically validated in Figure \ref{fig:grad_align_score}. In contrast, SOAP maintains a perfect inter-step alignment score of 1 throughout the entire training process. In the following section, we provide a theoretical explanation for this distinctive behavior.



\vspace{-1mm} 
\section{Gradient Alignment in Quasi Second-Order Optimization}
\vspace{-1mm} 

In this section, we aim to reveal why quasi Second-order optimizers can naturally promote gradient alignment via  the following lemma.


\begin{proposition}
\label{prop: newton_alignment_score}
Let $\mathcal{L}(\theta)$ be a twice differentiable loss function with Hessian $H(\theta)$, and let $P(\theta)$ be a  preconditioner with uniformly bounded inverse $P^{-1}(\theta)$. Consider the preconditioned gradient descent update with exponent $0 \leq s \leq 1$ and learning rate $\eta > 0$:
\begin{align}
\theta_{t+1} = \theta_t - \eta P^{-s}(\theta_t) \nabla \mathcal{L}(\theta_t).
\end{align}
Let  $g_t = \nabla \mathcal{L}(\theta_t)$, then  the alignment score $\mathcal{A}(g_t, g_{t+1})$ satisfies:
\begin{align}
\mathcal{A}(g_t, g_{t+1}) = 1 - \frac{\eta^2}{2}\frac{\|HP^{-s}g_t\|^2}{\|g_t\|^2} + O(\eta^3).
\end{align}
To maintain alignment $\mathcal{A}\left(g_t, g_{t+1}\right) \geq 1-\epsilon$ for $\epsilon>0$, the learning rate $\eta$ must satisfy:
\begin{align}
\eta \leq \sqrt{\frac{2\epsilon\|g_t\|^2}{\|HP^{-s}g_t\|^2}}.
\end{align}

\end{proposition}
The proof is presented in Appendix \ref{app:preconditioned_gd}. This bound specializes to the following cases: For vanilla gradient descent $(P = I)$, $\eta_{\max} = \sqrt{\frac{2\epsilon}{\lambda_{\max}^2(H)}}$, and for Newton's method $(P = H, s = 1)$, the maximum learning rate can be relaxed to $\eta_{\max} = \sqrt{2\epsilon}$.
These results imply that preconditioners approximating the Hessian effectively relax learning rate constraints, enabling the use of larger learning rates while maintaining optimization trajectory consistency. Comparing these cases reveals that Newton's method eliminates the dependency on the condition number of the Hessian, allowing for a constant maximum learning rate regardless of problem conditioning. This finding aligns with our theoretical understanding that accurate second-order information mitigates the effects of ill-conditioning, allowing for stable optimization even with aggressive learning rates.

Next, we establish that under some assumptions, the SOAP optimizer \cite{vyas2024soap} can be interpreted as approximating to using Hessian as the preconditioner:
\begin{align}
    w_{t+1} = w_{t} - \eta \operatorname{Soap}(g_t) \approx w_t - \eta H^{-1} g_t.
\end{align}
Formal assumptions and proof are provided in Appendix \ref{app:soap_newton}.

\begin{table}[h]
\centering
\caption{Comparison of optimization methods showing preconditioner types, storage and computational complexity for a $n \times n$ weight matrix, and practical compatibility with mini-batches and scalability with large neural networks. The theoretical connections between Shampoo, Muon, and quasi-second-order methods are detailed in Appendix \ref{app:soap_newton} and \ref{app:connection_shampoo_muon}.
}
\renewcommand{\arraystretch}{1.2}
\resizebox{\textwidth}{!}{%
\begin{tabular}{l|c|c|c|c|c}
\hline
\textbf{Method} & \textbf{(Approx.) Precond. } & \textbf{Storage} & \textbf{Computation} & \textbf{Mini-batch} & \textbf{DNN} \\
\hline\hline
Natural Gradient \cite{amari1998natural, muller2023achieving} & $F^{-1}$ & $O(n^4)$ & $O(n^6)$ & \xmark & \xmark \\
BFGS/L-BFGS \cite{liu1989limited} & $H^{-1}$ & $O(n^2)$ & $O(n^2)$ & \xmark & \cmark \\
SOAP \cite{vyas2024soap} & $H^{-1}$ & $O(n^2)$ & $O(n^3)$ & \cmark & \cmark \\
Kron \cite{li2017preconditioned} & $F^{-1/2}$ & $O(n^2)$ & $O(n^3)$ & \cmark & \cmark \\
Shampoo \cite{gupta2018shampoo} & $H_{\text{ada}}^{-1/2}$  & $O(n^2)$ & $O(n^3)$ & \cmark & \cmark \\
Muon \cite{jordan2024muon} & $H_{\text{ada}}^{-1/2}$ & $O(n^2)$ & $O(n^3)$ & \cmark & \cmark \\
ConFig \cite{liu2024config} & N/A & $O(n^2)$ & $O(n^2)$ & \cmark & \cmark \\
DCGD \cite{hwang2024dual} & N/A & $O(n^2)$ & $O(n^2)$ & \cmark & \cmark \\
\hline
\end{tabular}%
}
\label{tab:optimizers}
\end{table}

In Table \ref{tab:optimizers}, we compare  various popular methods to tackle the identified directional gradient conflicts in the context of PINNs. 
While all these methods are theoretically promising, some face practical limitations in the context of complex PDE systems.
L-BFGS is unsuitable for large-scale or stochastic training, as gradient noise disrupts its Hessian updates and line search procedures.  First-order variants such as ConFiG \cite{liu2024config} and DCGD \cite{hwang2024dual} attempt to alleviate conflicts via gradient surgery or projection. These methods yield incremental improvements but remain constrained by the inherent limitations of first-order updates.

Natural gradient descent (NGD) requires computing and inverting the Fisher information matrix at every iteration and is restricted to \texttt{float64} precision, which is inefficient on GPUs and consumes 2× more memory while being 2-4x slower than \texttt{float32}. As a result, NGD has only been demonstrated on relatively simple benchmarks with smooth solutions, where very small networks suffice and convergence issues rarely appear. On more challenging PDEs with sharp transitions or complex dynamics, NGD fails to scale: it is highly sensitive to hyperparameters, lacks mini-batching support, and often diverges.

To illustrate the limitations of NGD, we revisited the 2D Poisson and heat benchmarks of \cite{muller2023achieving}, using their official implementation and comparing against SOAP under both \texttt{float32} and \texttt{float64} precision. We tested MLPs of varying depth and width, reporting the best results across ten random seeds for each method. While SOAP reliably converged, NGD exhibited some failures for some random seeds, which is also acknowledged in \cite{muller2023achieving}. Even under successful runs, NGD underperforms SOAP in both accuracy and stability, as summarized in Table \ref{tab:ngd_vs_soap}.

\begin{table}[htbp]
\vspace{-1mm}
\centering
\caption{Comparison of relative $L^2$ errors for different MLP architectures and optimization methods (NGD vs.\ SOAP) on Poisson and Heat 2D PDE benchmarks, evaluated in both \texttt{float32} and \texttt{float64} precision.}

\label{tab:ngd_vs_soap}
\resizebox{\textwidth}{!}{
\begin{tabular}{@{}llcc@{}}
\toprule
\textbf{PDE} & \textbf{Architecture (Method)} & \textbf{Float32} & \textbf{Float64} \\
\midrule
\multirow{7}{*}{\textbf{Poisson}} 
  & [2, 32, 1] (NGD)           & $4.87 \times 10^{-2} \pm 3.61 \times 10^{-2}$ & $3.84 \times 10^{-7} \pm 2.92 \times 10^{-7}$ \\
  & [2, 32, 32, 1] (NGD)       & $1.65 \times 10^{-1} \pm 3.93 \times 10^{-2}$ & $1.27 \times 10^{-6} \pm 2.86 \times 10^{-7}$ \\
  & [2, 32, 32, 32, 1] (NGD)   & $2.48 \times 10^{-1} \pm 1.75 \times 10^{-2}$ & $3.14 \times 10^{-6} \pm 5.10 \times 10^{-7}$ \\
  & [2, 256, 1] (NGD)          & $1.56 \times 10^{-1} \pm 6.34 \times 10^{-2}$ & $6.10 \times 10^{-7} \pm 1.68 \times 10^{-7}$ \\
  \cmidrule{2-4}
  & [2, 256, 1] (SOAP)         & $3.06 \times 10^{-6} \pm 7.12 \times 10^{-7}$ & $6.08 \times 10^{-7} \pm 2.13 \times 10^{-7}$ \\
  & [2, 256, 256, 1] (SOAP)    & $1.87 \times 10^{-6} \pm 5.19 \times 10^{-7}$ & $4.06 \times 10^{-7} \pm 1.81 \times 10^{-7}$ \\
  & [2, 256, 256, 256, 1] (SOAP) & $\mathbf{1.35 \times 10^{-6} \pm 3.45 \times 10^{-7}}$ & $\mathbf{2.99 \times 10^{-7} \pm 1.05 \times 10^{-7}}$ \\
\midrule
\multirow{7}{*}{\textbf{Heat 2D}} 
  & [2, 32, 1] (NGD)           & $5.98 \times 10^{-2} \pm 5.46 \times 10^{-2}$ & $7.68 \times 10^{-6} \pm 1.85 \times 10^{-6}$ \\
  & [2, 32, 32, 1] (NGD)       & $5.95 \times 10^{-1} \pm 2.29 \times 10^{-4}$ & $2.32 \times 10^{-6} \pm 1.17 \times 10^{-6}$ \\
  & [2, 32, 32, 32, 1] (NGD)   & $5.95 \times 10^{-1} \pm 4.58 \times 10^{-4}$ & $5.13 \times 10^{-6} \pm 5.25 \times 10^{-7}$ \\
  & [2, 256, 1] (NGD)          & $5.95 \times 10^{-1} \pm 1.05 \times 10^{-3}$ & $8.69 \times 10^{-6} \pm 6.49 \times 10^{-6}$ \\
  \cmidrule{2-4}
  & [2, 256, 1] (SOAP)         & $4.61 \times 10^{-6} \pm 8.12 \times 10^{-7}$ & $3.03 \times 10^{-6} \pm 6.06 \times 10^{-7}$ \\
  & [2, 256, 256, 1] (SOAP)    & $2.74 \times 10^{-6} \pm 9.52 \times 10^{-7}$ & $2.04 \times 10^{-6} \pm 4.10 \times 10^{-7}$ \\
  & [2, 256, 256, 256, 1] (SOAP) & $\mathbf{2.65 \times 10^{-6} \pm 5.00 \times 10^{-7}}$ & $\mathbf{1.33 \times 10^{-6} \pm 2.61 \times 10^{-7}}$ \\
\bottomrule
\end{tabular}
}
\end{table}

In contrast, quasi-second-order methods (e.g., SOAP \cite{vyas2024soap}, Kron \cite{li2017preconditioned}, Muon \cite{jordan2024muon}) naturally promote directional gradient conflicts via preconditioning and  mitigate ill-conditioning in the loss landscape \cite{rathore2024challenges} while maintaining computational tractability. 
As previously illustrated in Figure\ref{fig:grad_align_score}, SOAP consistently achieves the highest gradient alignment scores, likely due to its closer approximation to Newton’s method -- an observation further supported by our theoretical analysis in Appendix \ref{app:soap_newton}.

Building on these insights, we now present comprehensive numerical experiments that demonstrate the superior accuracy and convergence behavior of quasi-second-order methods across a diverse set of PDE benchmarks.

\section{Experiments}

\label{sec:results}

\begin{figure}[t]
\vspace{-1mm}
    \centering
    \includegraphics[width=1.0\linewidth]{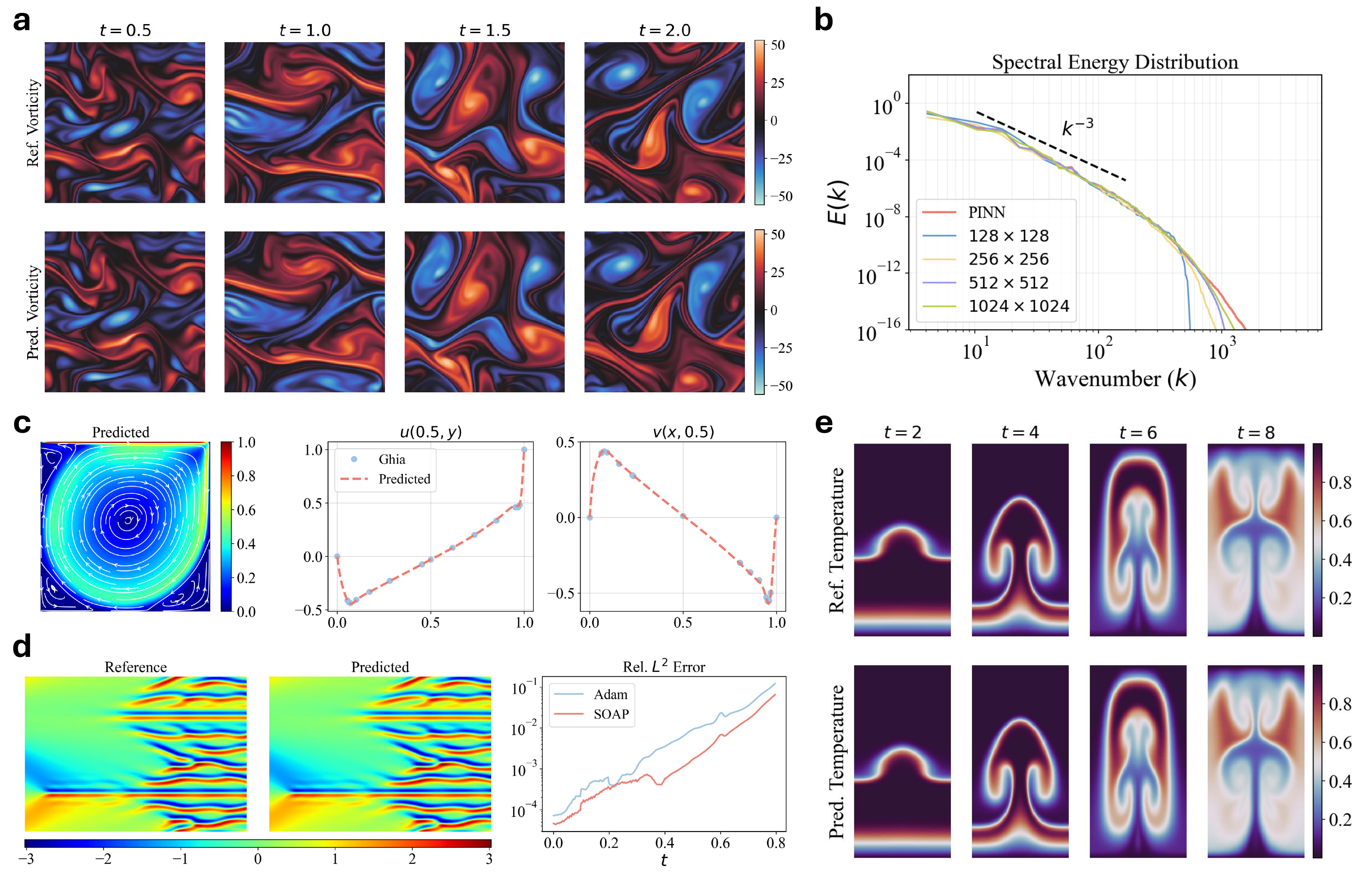}
    \caption{Simulating complex fluid dynamics using PINNs with SOAP optimization. (a) Kolmogorov flow at Re=10,000: comparison between reference solution and PINN predictions demonstrates accurate capture of turbulent structures across multiple time steps. (b) Spectral energy distribution showing PINN's superior resolution of fine-scale dynamics compared to traditional numerical solutions at various grid resolutions. (c) Lid-driven cavity flow at Re=5,000: streamlines and centerline velocity profiles show excellent agreement with benchmark data from \cite{GHIA1982387}. (d) Kuramoto-Sivashinsky equation: PINNs accurately predicts complex spatiotemporal patterns and chaotic dynamics. (e) Rayleigh-Taylor instability (Pr=0.71, Ra=$10^6$): evolution of temperature field shows precise capture of interface dynamics and mushroom-shaped structures characteristic of this flow.}
    \label{fig:master_figure}
    \vspace{-2mm}
\end{figure}

To rigorously evaluate the performance of the aforementioned quasi second-order methods, we examine a diverse set of 10 representative and challenging PDEs that govern fundamental physical phenomena. These equations span wave propagation, shock formulation, chaotic systems, reaction-diffusion processes,  fluid dynamics, and heat transfer. The detailed description of the problem setup, including the PDE parameters, initial and boundary conditions, numerical implementations, and supplementary visualizations, are presented in Appendix \ref{appendix: experiments}.

\paragraph{Baselines.} 
We focus our comparisons on PINN approaches that are scalable to large neural networks, as scalability is essential for solving realistic, large-scale physical problems. Consequently, we exclude methods based on natural gradients \cite{muller2023achieving, jnini2024gauss} and L-BFGS variants \cite{urban2025unveiling}, as these typically rely on full-batch training, require high-precision (e.g., float64) computation, and are limited to small network sizes—making them impractical for the complex PDE benchmarks considered in this work.

Our baseline setup is based on the current state-of-the-art training pipeline proposed by \cite{wang2024piratenets}. 
Specifically, we adopt PirateNet \cite{wang2024piratenets} as the backbone architecture, which is known for its stability and scalability to deeper networks. 
All weight matrices are initialized using random weight factorization (RWF) \cite{wang2022random},
Exact periodic boundary conditions are strictly enforced when applicable \cite{dong2021method}.

For model training, we use mini-batch gradient descent with the Adam optimizer \cite{kingma2014adam}, which has become the de facto standard for training PINNs due to its robust performance and computational efficiency.
To improve training efficiency and robustness, we use
learning rate annealing \cite{wang2021understanding,wang2023expert} for loss balancing. 
In addition, we employ causal training \cite{wang2022respecting,wang2023expert} to address causality violations when solving time-dependent PDEs.  
For challenging benchmarks, we implement time-marching and curriculum learning strategies \cite{krishnapriyan2021characterizing}.

Importantly, this baseline represents one of the most accurate and scalable PINN pipelines currently available, achieving state-of-the-art performance across a diverse set of PDE benchmarks \cite{wang2023expert, hao2023pinnacle}. For instance, on the standard Allen-Cahn benchmark, which has been extensively evaluated by various PINN methods, our proposed baseline outperforms most existing approaches, as demonstrated in Table \ref{tab: AC}. A comprehensive description of the techniques employed and the hyperparameter configurations are provided in Appendices~\ref{appendix:arch} and \ref{appendix:training}, respectively.




\paragraph{State-of-the-art results.} 
Table~\ref{tab: sota} highlights the consistent performance improvements of quasi-second-order optimizers across a wide range of PDE benchmarks. Among them, SOAP achieves the best overall results, likely due to its closer alignment with Newton’s method (Table \ref{tab:optimizers}), making it particularly well-suited for the multi-objective nature of PINN optimization.

Compared to our baselines, SOAP reduces relative error by 6.4x on the wave equation. For nonlinear 1D problems, it achieves a 6.9x improvement on the Allen-Cahn equation, and approximately 2x improvements on both the Korteweg-de Vries and Kuramoto-Sivashinsky equations.
The performance gains become particularly pronounced for coupled diffusion-reaction systems. The Grey-Scott and Ginzburg-Landau equations exhibit an order of magnitude reduction in error. On challenging Navier-Stokes benchmarks, including the lid-driven cavity and Rayleigh-Taylor instability problems, SOAP demonstrates a more than 10x error reduction. We highlight and discuss these substantial improvements in detail below.

\paragraph{Complex fluid dynamics.}
Our most significant achievement is successfully applying PINNs to complex fluid dynamics problems that were previously considered beyond their capabilities. In particular, we demonstrate breakthrough results in three challenging cases that combine multiple physical constraints and have historically proven difficult for PINNs, see Figure \ref{fig:master_figure}.

For the lid-driven cavity flow at Reynolds number 5,000, SOAP enables a dramatic improvement in accuracy, reducing the relative $L^2$ error from 32.4\% to 3.99\%. As shown in Figure \ref{fig:master_figure}c, our model successfully captures intricate flow features including secondary and tertiary corner vortices, showing excellent agreement with the benchmark results of \cite{GHIA1982387}. 

The Rayleigh-Taylor instability presents an even more challenging test, requiring simultaneous handling of interface dynamics and coupled velocity-density evolution. SOAP enables accurate prediction of the characteristic mushroom-shaped structures that develop as heavier fluid penetrates into lighter fluid, achieving a relative $L^2$ error of 0.52\% -- nearly an order of magnitude improvement over the best baseline's 7.32\%. Figure \ref{fig:master_figure}e demonstrates excellent agreement with reference solutions across multiple time steps, capturing both the initial linear growth phase and subsequent nonlinear development.

Our most impressive result comes from the turbulent Kolmogorov flow at Reynolds number 10,000 -- marking the first time PINNs have successfully captured turbulent dynamics at such high Reynolds numbers. Our model achieves a relative $L^2$ error of 3.20\%, compared to 20.4\% with our baseline. Figure \ref{fig:master_figure}a shows that our predictions accurately reproduce both the large-scale flow structures and the complex cascade of smaller eddies characteristic of turbulent flows. Moreover, spectral analysis reveals that our PINN solution maintains higher spectral energy at high wavenumbers compared to traditional numerical solvers, even those using a $1024 \times 1024$ grid resolution. This demonstrates PINNs' potential advantage in resolving fine-scale dynamics without requiring explicit grid discretization -- a key capacity for turbulence modeling.

\begin{table}[t]
    \vspace{-1mm}
    \centering
    \renewcommand{\arraystretch}{1.4}
    \caption{Comparison of optimizer performance obtained by training PINNs with Adam, Adam+L-BFGS, and SOAP, respectively, across various PDEs, following the training pipeline described in Section \ref{sec:results}. The evaluation metric is the relative $L^2$ error over the entire spatial-temporal domain.}
    \label{tab: sota}
    \resizebox{\textwidth}{!}{
    \begin{tabular}{l ccccc} 
        \toprule
        \textbf{Benchmark} & \textbf{Adam} & \textbf{Adam+L-BFGS} & \textbf{Kron} & \textbf{Muon} & \textbf{SOAP} \\
        \midrule
        Wave & ${5.15 \times 10^{-5}}$ & ${5.08 \times 10^{-5}}$  & ${8.62 \times 10^{-6}}$ & ${9.34 \times 10^{-6}}$ & $\mathbf{8.05 \times 10^{-6}}$ \\ 
        Burgers & ${8.20 \times 10^{-5}}$ & ${8.20 \times 10^{-5}}$ & ${4.85\times 10^{-5}}$ & ${4.52\times 10^{-5}}$ & $\mathbf{4.03\times 10^{-5}}$ \\   
        Allen-Cahn & ${2.24 \times 10^{-5}}$ &  ${2.25 \times 10^{-5}}$ & ${3.63 \times 10^{-6}}$ & ${4.95 \times 10^{-6}}$ & $\mathbf{3.48 \times 10^{-6}}$ \\ 
        Korteweg–De Vries & ${7.04 \times 10^{-4}}$ & ${7.33 \times 10^{-4}}$ & ${5.48 \times 10^{-4}}$ & ${4.19 \times 10^{-4}}$ & $\mathbf{3.40 \times 10^{-4}}$ \\
        Kuramoto-Sivashinsky & ${7.48 \times 10^{-2}}$ & {--} & ${5.49 \times 10^{-2}}$ & $\mathbf{3.51 \times 10^{-2}}$ & ${3.86 \times 10^{-2}}$ \\
        Grey-Scott & ${3.61 \times 10^{-3}}$ & {--} & $1.89 \times 10^{-4}$ & ${1.95 \times 10^{-4}}$ & $\mathbf{1.88 \times 10^{-4}}$ \\
        Ginzburg-Landau & ${1.49 \times 10^{-2}}$ & {--} & ${7.53 \times 10^{-3}}$ & $\mathbf{4.58 \times 10^{-3}}$ & ${4.78 \times 10^{-3}}$ \\
        Lid-driven cavity ($\text{Re}=5 \times 10^3$) & $3.24 \times 10^{-1}$ & {--} & ${7.05 \times 10^{-2}}$ & ${6.70 \times 10^{-2}}$ & $\mathbf{3.99 \times 10^{-2}}$ \\
        Kolmogorov flow ($\text{Re}=10^4$) & $2.04 \times 10^{-1}$ & {--} & $8.62 \times 10^{-2}$ & ${6.89 \times 10^{-2}}$ & $\mathbf{3.20 \times 10^{-2}}$ \\
        Rayleigh-Taylor instability ($\text{Ra}=10^6$) & $7.32 \times 10^{-2}$ & {--} & ${5.74 \times 10^{-3}}$ & ${1.80 \times 10^{-2}}$ & $\mathbf{5.22 \times 10^{-3}}$ \\
        \bottomrule 
    \end{tabular}
    }
    \vspace{-2mm}
\end{table}

\begin{figure}
    \vspace{-1mm}
    \centering
    \includegraphics[width=1.0\linewidth]{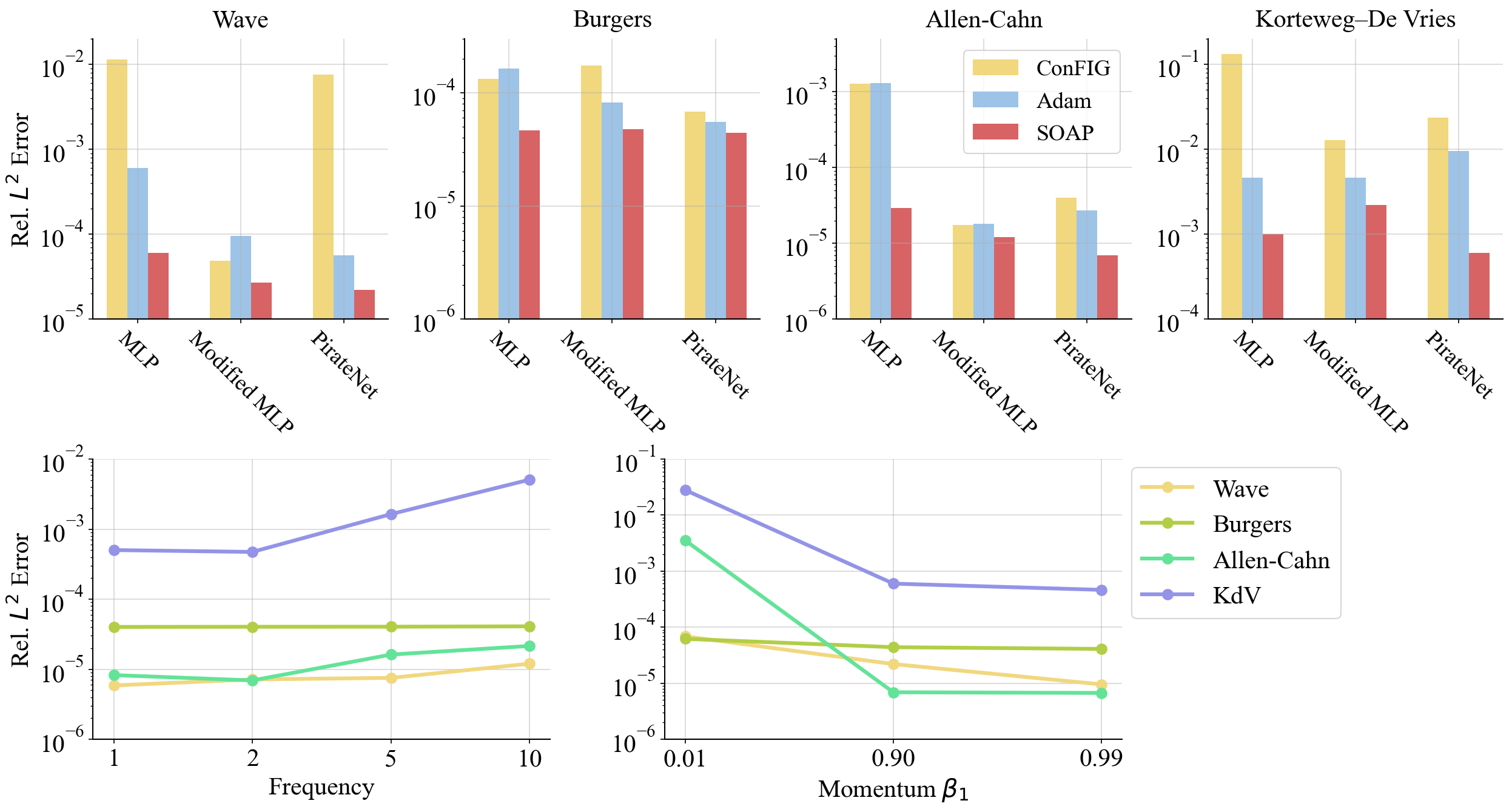}
\caption{Optimizer performance comparison and ablation studies.  Top: Relative $L^2$ error across PDE benchmarks using different optimizers. Bottom left: Relative $L^2$ error for varying preconditioner update frequencies in SOAP optimizer. Bottom right: Relative $L^2$ error with different momentum values in SOAP optimizer.}
    \label{fig:ablation}
    \vspace{-2mm}
\end{figure}

\paragraph{Ablation studies.} We conduct systematic experiments to evaluate SOAP's performance across different architectures and hyperparameter settings, establishing the robustness of our approach. Our first investigation examines SOAP's effectiveness across three representative architectures: standard MLP, modified MLP \cite{wang2021understanding}, and PirateNet \cite{wang2024piratenets}. As shown in the top panel of Figure \ref{fig:ablation}, testing each architecture on four benchmark PDEs (Wave, Burgers, Allen-Cahn, and KdV equations), we find that SOAP consistently improves accuracy compared to Adam regardless of the underlying network backbones. In particular, PirateNet seems to be the most effective architecture across all test cases, leading to its selection for our main experiments.

As illustrated in the bottom panel of Figure \ref{fig:ablation}, our results of SOAP's hyperparameters reveal two critical factors affecting performance. The preconditioner update frequency presents a clear trade-off between accuracy and computational cost. While more frequent updates yield better results, the improvements diminish beyond an update frequency of 2, which we selected as the optimal balance for our experiments. The momentum parameter $\beta_1$ proved especially crucial: high momentum ($\beta_1=0.99$) consistently achieves the best results, while low momentum ($\beta_1$=0.01) significantly degrades performance across all test cases.

For completeness, we also compared SOAP against ConFiG \cite{liu2024config}, a recently proposed method for addressing gradient conflicts in PINNs that has demonstrated relative good performance compared to established baselines such as PCGrad \cite{yu2020gradient} and  IMTL-G  \cite{liu2021towards} in multi-task learning. Despite its promising theoretical foundations, ConFiG showed some sensitivity to hyperparameter settings in our experiments, resulting in inconsistent performance. These results highlight the practical advantages of SOAP's more robust optimization approach. 



\paragraph{Computational costs.} While SOAP requires approximately 2x longer training time compared to baselines (Table \ref{tab: cost}), our focus is exploring the performance frontier of PINNs through extended training to full convergence. Impressively, error and loss convergence curves (Appendix \ref{appendix:benchmarks}) indicate that SOAP typically achieves rapid initial convergence, reaching a reasonable accuracy (approximately $10^{-4}$)  within the first 10,000 iterations, followed by gradual error reduction in subsequent iterations. This suggests the potential for reducing training time by up to 10x while maintaining competitive performance.
These findings motivate future research into designing computationally efficient optimization algorithms and training strategies for PINNs, paving the way for practical and scalable applications in complex physics simulations.

\section{Conclusion}
\label{sec: conclusion}

This work advances our understanding of gradient conflicts in training of PINNs. We proposed gradient alignment scores as a quantitative measure of such conflicts, and provide both theoretical analysis and empirical evidence demonstrating their prevalence during PINN optimization.  Furthermore, we show that second-order optimization methods implicitly promote gradient alignment, offering a principled approach to mitigating these conflicts. Particularly, we uncover a connection between SOAP and Newton's method. Extensive experiments across 10 PDE benchmarks confirm the effectiveness of quasi second-order optimizers, 
 with SOAP achieving state-of-the-art performance.

Building on these insights, several promising research directions emerge.  While SOAP demonstrates the power of gradient alignment in handling coupled physical constraints, opportunities exist for more efficient preconditioned algorithms that maintain their effectiveness with reduced computational cost. More broadly, our work suggests that the principles of gradient alignment and second-order preconditioning could benefit many deep learning applications involving competing objectives, though challenges remain in scaling these approaches to larger systems. Success in these directions could transform both scientific computing and multi-task optimization.

\section*{Acknowledgment}
B.L. would like to acknowledge support from National Key R$\&$D Program of China Grant No. 2024YFA1016000. P.P. and S.W. acknowledge support from the US Department of Energy under the Advanced Scientific Computing Research program (grant DE-SC0024563), the Nvidia Academic Grant Program, and the Institute for Foundations of Data Science at Yale University. 
We also thank the developers of the software that enabled our research, including JAX \cite{jax2018github},  Matplotlib \cite{hunter2007matplotlib}, and NumPy \cite{harris2020array}.

\bibliographystyle{unsrt}  
\bibliography{references}


\appendix

\section{Nomenclature}

\begin{table}[H]
    \renewcommand{\arraystretch}{1.4}
\centering
\caption{Notation used throughout the paper. We use uppercase letters for matrices and lowercase letters for their vectorized forms. All gradients and Hessians are with respect to the loss function $\mathcal{L}$.}
\label{tab:notation}
\begin{tabular}{c|l}
\toprule
\textbf{Symbol} & \textbf{Description} \\
\midrule
$\mathcal{L}$ & Loss function \\
$\mathcal{L}_{ics}$ & Initial condition losses \\
$\mathcal{L}_{res}$ & PDE residual losses \\
$\theta$ & Neural network parameters \\
$W$ & Weight matrix for a given layer \\
$w$ & Vectorized weight matrix, $w = \text{Vec}(W)$ \\
$G$ & Gradient matrix, $G = \nabla_W \mathcal{L}$ \\
$g$ & Vectorized gradient, $g = \text{Vec}(G)$ \\
$F$ & Fisher information matrix\\
$H$ & Hessian matrix, $H = \nabla^2_\theta \mathcal{L}$ \\
$H_{\text{Ada}}$ & Full Adagrad preconditioner matrix \\
$H_{\text{GN}}$ & Gauss-Newton approximation of the Hessian \\
$\mathcal{A}$ & Gradient alignment score between loss components \\
\bottomrule
\end{tabular}

\end{table}

\section{Analysis of Intra-step Gradient Alignment}

\begin{figure}[ht]
    \centering
    \includegraphics[width=1.0\linewidth]{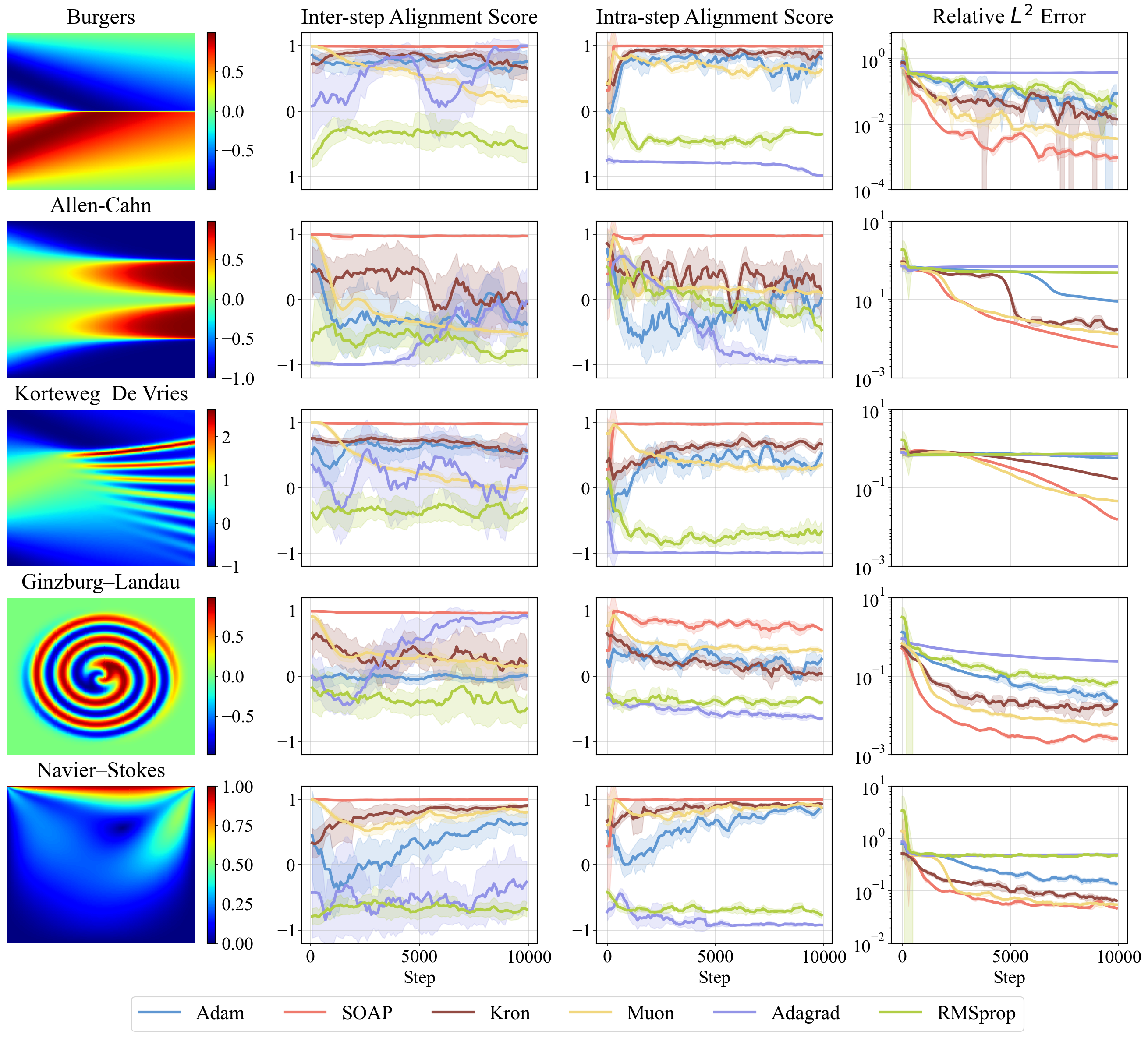}
    \caption{\small{Gradient alignment scores and test errors obtained by training PINNs with different optimizers across different PDEs. From left to right: ground truth PDE solution, intra-step gradient alignment scores (Eq. \eqref{eq: intra_align}), inter-step gradient alignment scores (Eq. \eqref{eq: inter_align}), and test error convergence during training.}}
    \label{fig:grad_align_score}
\end{figure}
\label{app:gradientconflict}
We present some preliminary analysis to understand intra-step gradient conflicts in training PINNs via standard gradient descent, Adam \cite{kingma2014adam}, and Shampoo algorithms \cite{gupta2018shampoo}, and how SOAP can effectively resolve them during training. For simplicity, we consider the simplest case of using PINNs with the two-layer NN to solve the one-dimensional Laplace equation and 
focus on the analysis of the intra-step gradient alignment \eqref{eq: intra_align} 
with small initialization. The analysis can be easily extended to other types of PDEs. 
Following the general setup in \cref{subsec: pinns}, without loss of generality, we consider 1D Laplace equation as follows
\begin{align}
    \left\{
\begin{array}{ll}
\Delta u = u'' = 0 & \text{on } [-1,1], \\
u(\pm 1) = g_{\pm 1}. & 
\end{array}
\right.
\end{align}
We approximate the solution $u(x)$ by a two-layer network with width $N$: 
\begin{equation} 
    u(x, \theta) = \sum_{i = 1}^N a_i \si ( w_i x) = \ba \dd \si(\bw x)\,,
\end{equation}
where $\ba = (a_1, \ldots, a_N), \bw = (w_1, \ldots, w_N)  \in \R^N$, and $\theta = (\ba,\bw) \in \R^{2N}$. Moreover, we limit ourselves to the activation function $\si(x) = \tanh(x)$. In this case, the loss \eqref{eq: PINN_loss} reduces to 
\begin{equation} 
    \min_{\theta = (\ba, \mathbf{w})} \mc{L}(\theta) = \underbrace{\frac{1}{N_r} \sum_{p = 1}^{N_r} |u''(x_p, \theta)|^2}_{\mc{L}_{r}(\theta)} + \underbrace{\frac{1}{2} \sum_{s = \pm 1} |u(s, \theta) - g_{s}|^2}_{\mc{L}_{bc}(\theta)}\,.
\end{equation}
To analyze the gradient conflict phenomenon in training PINNs, we consider the \emph{small initialization} regime. 
\begin{assumption} 
\label{assp1}
    The weights $a_i, w_i$ are initialized by i.i.d. Gaussian $\mc{N}(0,\ep^2)$ with small $\ep = o(1)$. 
\end{assumption}
This allows us to introduce the normalized parameters:  
\begin{equation*}
    \bar{\ba} = \ep^{-1} \ba\,, \q \bar{\bw} = \ep^{-1} \bw\,, 
\end{equation*}
initialized as standard Gaussian.

\begin{lemma}
\label{lemma51}
Under small initialization, the gradients of the residual and boundary loss terms can be approximated as:
\begin{align}
\nabla_\theta \mathcal{L}_r(\theta) &= \ep^7 G^r(\bar{\mathbf{a}},\bar{\mathbf{w}}) + O(\ep^9), \\
\nabla\theta \mathcal{L}_{bc}(\theta) &= \ep G^{bc}(\bar{\mathbf{a}},\bar{\mathbf{w}}) + O(\ep^3),
\end{align}
where
\begin{align} \label{eq:effgrad1}
    G^r(\wb{\ba},\wb{\bw}) &=  c_r  \wb{\ba} \dd \wb{\bw}^{\odot 3} \left(  \wb{\bw}^{\odot 3},   3  \wb{\ba} \odot \wb{\bw}^{\odot 2}  \right), \\
  \label{eq:effgrad2}
    G^{bc}(\wb{\ba},\wb{\bw}) &= (g_{-1} - g_1) \left(\wb{\bw}, \wb{\ba} \right).
\end{align}
Here $\bar{\mathbf{a}} = \ep^{-1}\mathbf{a}$, $\bar{\mathbf{w}} = \ep^{-1}\mathbf{w}$ are the normalized parameters, and $G^r$, $G^{bc}$ are the effective gradient terms.
\end{lemma}
We remark that these elementary computations also provide insights into the gradient magnitude imbalance discussed in \cref{subsec:gradconflict}, noting $\|\na_\theta \mc{L}_r(\theta) \| = O(\ep^7)$ while $\| \na_{\theta} \mc{L}_{bc}(\theta) \| = O(\ep)$. 

\begin{proof}
    We recall the Taylor expansions of the activation function $\si(x) = \tanh(x)$ and its derivatives for later use:
\begin{equation} \label{eq:taylorexp}
    \begin{aligned}
         &\si(x) = x - \frac{x^3}{3} + O(x^5)\,, \quad \si'(x) = 1 - x^2 + O(x^4)\,, \\ & \si''(x) = - 2 x + \frac{8 x^3}{3} + O(x^5)\,, \quad \si'''(x) = -2 + 8 x^2 + O(x^4)\,.
    \end{aligned}
\end{equation}
The gradient of loss function $\mc{L}(\theta)$ consists of two parts computed as follows:
\begin{equation*}
    \na_{\theta = (\ba, \bw)} \mc{L}_r(\theta) = \frac{2}{N_r} \sum_{p = 1}^{N_r} u_{xx}(x_p, \theta) \na_{\theta = (\ba, \bw)} u_{xx}(x_p\,, \theta)\,,
\end{equation*}
and 
\begin{equation*}
    \na_{\theta = (\ba, \bw)} \mc{L}_{bc}(\theta) = \sum_{s = \pm 1} (u(s, \theta) - g(s)) \na_{\theta = (\ba, \bw)} u(s, \theta)\,,
\end{equation*}
with, thanks to \eqref{eq:twolayersol} and \eqref{eq:taylorexp}, 
\begin{align*}
     \na_{\ba}u(x,\theta) &= \si(\mathbf{w} \x) = \ep \wb{\bw} x + O(\ep^3)\,,\\
     \na_{w_i} u(x,\theta) &= a_i \si'(w_i x) x = \ep \wb{a_i} x - \ep^3 \wb{a_i} \wb{w_i}^2 x^3 + O(\ep^4)\,,
\end{align*}
and 
\begin{align*}
     \na_{a_i} u_{xx}(x, \theta) &= \si''(w_i x) |w_i|^2 = - 2 \ep^3 \wb{w_i}^3 x + O(\ep^5)\,, \\ 
     \na_{w_i} u_{xx}(\x, \theta) & = a_i \si'''(w_i x) |w_i|^2 x + 2 a_i \si''(w_i x) w_i \\
     & = - 6 \ep^3 \wb{a_i} \wb{w_i}^2 x + O(\ep^5) \,.
\end{align*}
We also compute 
\begin{align*}
    u(x,\theta) = \ep^2 \wb{\ba} \dd \wb{\bw} x + O(\ep^4)\,,
\end{align*}
and 
\begin{equation*}
    u_{xx}(x, \theta) = \sum_i a_i \si''(w_i x) |w_i|^2 = - 2 \ep^4\sum_i \wb{a_i} \wb{w_i}^3 x + O(\ep^6)\,.
\end{equation*}
For convenience, we define the componentwise power $\x^{\odot k} = (x_1^k,\ldots, x_N^k)$ and product $ \x \odot \mathbf{y} = (x_1y_1, \ldots, x_N y_N)$ for $\x, \mathbf{y} \in \R^N$.
By the above computation, it follows that at the initialization, there hold
\begin{align*}
      \na_{\theta = (\ba, \bw)} \mc{L}_r(\theta) & = \frac{2}{N_r} \sum_{p = 1}^{N_r} \left(- 2 \ep^4  \wb{\ba} \dd \wb{\bw}^{\odot 3} x_p \right) \left(- 2 \ep^3 \wb{\bw}^{\odot 3} x_p,   - 6 \ep^3 \wb{\ba} \odot \wb{\bw}^{\odot 2} x_p   \right) + O(\ep^9)  \\ 
      & = \ep^7 \frac{8}{N_r} \sum_{p = 1}^{N_r} \wb{\ba} \dd \wb{\bw}^{\odot 3} \left(  \wb{\bw}^{\odot 3},   3  \wb{\ba} \odot \wb{\bw}^{\odot 2}  \right) x_p^2 + O(\ep^9)\,,
\end{align*}
and 
\begin{align*}
        \na_{\theta = (\ba, \bw)} \mc{L}_{bc}(\theta) & = \sum_{s = \pm 1} \left(\ep^2 \wb{\ba} \dd \wb{\bw} s + O(\ep^4) - g(s)\right) \left(\ep \wb{\bw} s + O(\ep^3), \ep \wb{\ba} s - \ep^3 \wb{\ba} \odot \wb{\bw}^{\odot 2} s^3 + O(\ep^4) \right) \\
        & = - \ep \sum_{s = \pm 1} s g(s) \left(\wb{\bw}, \wb{\ba} \right) + O(\ep^3) =  \ep (g_{-1} - g_1) \left(\wb{\bw}, \wb{\ba} \right) + O(\ep^3)\,.
\end{align*}
We then define constant $c_r = 8 N_r^{-1} \sum_{p = 1}^{N_r} x_p^2 > 0$ and the effective gradients 
\begin{align} 
    G^r(\wb{\ba},\wb{\bw}) = \left(G^r_a(\wb{\ba},\wb{\bw}), G^r_w(\wb{\ba},\wb{\bw})\right) =  c_r  \wb{\ba} \dd \wb{\bw}^{\odot 3} \left(  \wb{\bw}^{\odot 3},   3  \wb{\ba} \odot \wb{\bw}^{\odot 2}  \right)\,,
\end{align}
and 
\begin{align}  
    G^{bc}(\wb{\ba},\wb{\bw}) = \left(G^{bc}_a(\wb{\ba},\wb{\bw}), G^{bc}_w(\wb{\ba},\wb{\bw})\right) = (g_{-1} - g_1) \left(\wb{\bw}, \wb{\ba} \right)\,,
\end{align}
enabling us to write 
\begin{equation} \label{eq:effecgrad}
    \na_\theta \mc{L}_r(\theta) = \ep^7 G^r(\wb{\ba},\wb{\bw}) + O(\ep^9)\,, \q \na_{\theta} \mc{L}_{bc}(\theta) = \ep G^{bc}(\wb{\ba},\wb{\bw}) + O(\ep^3)\,.
\end{equation}
\end{proof}

We are now ready to understand the gradient conflict for various optimizers applied to the residual and boundary loss terms separately. 

\textbf{Proposition 3}
{\it At initialization, the alignment score converges to a binary random variable in the infinite width limit:
\begin{align}
    \lim_{N \rightarrow \infty }\mathcal{A}(\square(\nabla \mathcal{L}_b),\square(\nabla \mathcal{L}_r)) = O(\ep^2) + C_\square \begin{cases}\operatorname{sgn}\left(g_{-1}-g_1\right) & \text { with prob. } \frac{1}{2}, \\ -\operatorname{sgn}\left(g_{-1}-g_1\right) & \text { with prob. } \frac{1}{2} .\end{cases}
\end{align}
where $\square = \operatorname{GD}, \operatorname{Adam}, \operatorname{Shampoo}, \text{ or } \operatorname{Soap}$  
 denotes the optimizer update rule, and
$C_\square$ is a constant depending on the optimizer. }

 We can see that these optimizers fail to resolve intra-step gradient conflicts in the initialization, aligning with the near-zero initial intra-step gradient scores shown in Figure \ref{fig:grad_align_score}.

\begin{proof}
    \noindent 
{\bf Gradient descent.} We start with the standard continuous-time gradient descent: 
\begin{align} \label{eq:gradientflow}
    \frac{\rd \theta}{\rd t} = - \na_{\theta} \mc{L}(\theta)\,.
\end{align}
Motivated by \cite{zhou2022towards,chen2024dynamics}, under small initialization \cref{assp1}, in the \emph{initial stage} of training dynamics where the leading-order expansion \eqref{eq:effecgrad} holds for the weights $\ba, \bw$, the gradients $ \na_\theta \mc{L}_r(\theta)$ and $\na_{\theta} \mc{L}_{bc}(\theta)$ can be effectively described by  $G^r(\wb{\ba},\wb{\bw})$ and $G^{bc}(\wb{\ba},\wb{\bw})$, respectively, up to some scaling factors, then the gradient flow \eqref{eq:gradientflow} can be approximated by the effective dynamics for the normalized parameter $\wb{\theta} = (\wb{\ba},\wb{\bw})$: 
\begin{align} \label{eq:effdym}
    \ep \frac{\rd \wb{\theta}}{\rd t} = - \left(\ep^7 G^r(\wb{\theta}) + \ep G^{bc}(\wb{\theta}) \right)\,.    
\end{align}
Recalling \cref{def:gradalign}, under our assumptions, we have the intra alignment score:
\begin{align*}
    \mc{A}\left(\na_\theta \mc{L}_r(\theta), \na_\theta \mc{L}_{bc}(\theta) \right) = \mc{A}\left( G^r(\wb{\theta}) , G^{bc}(\wb{\theta})  \right) + O(\ep^2)\,,
\end{align*}
where 
\begin{align*}
    \mc{A}\left( G^r(\wb{\theta}) , G^{bc}(\wb{\theta})  \right) = \sgn (\wb{\ba} \dd \wb{\bw}^{\odot 3}) \sgn (g_{-1} - g_1) \frac{\sum_i \wb{w_i}^4 + 3 \sum_i \wb{a_i}^2 \wb{w_i}^2}{\sqrt{\sum_i \wb{w_i}^6 + 9 \wb{a_i}^2 \wb{w_i}^4} \sqrt{\sum_i \wb{w_i}^2 + \wb{a_i}^2}}\,.
\end{align*}
Then we find that at the initialization, by
$\wb{a_i}, \wb{w_i} \sim \mc{N}(0,1)$ from \cref{assp1} and the law of large numbers, 
\begin{equation*}
    \frac{\frac{1}{N}\sum_i \wb{w_i}^4 + 3 \frac{1}{N} \sum_i \wb{a_i}^2 \wb{w_i}^2}{\sqrt{\frac{1}{N}\sum_i \wb{w_i}^6 + 9 \wb{a_i}^2 \wb{w_i}^4} \sqrt{\frac{1}{N}\sum_i \wb{w_i}^2 + \wb{a_i}^2}} \longrightarrow \frac{6}{\sqrt{15 + 27} \sqrt{2}} = \frac{3}{\sqrt{21}}\,,\q \text{almost surely}.
\end{equation*}
Also, by the symmetry of Gaussian, there holds 
$\mathbb{P}(\sum_i \wb{a_i} \wb{w_i}^3 > 0) = \frac{1}{2}$. It follows that the alignment score $\mc{A}( G^r(\wb{\theta}) , G^{bc}(\wb{\theta}))_{t = 0}$ converges to a binary random variable with expectation zero in the infinite width limit: 
\begin{equation*}
    \mc{A}\left( G^r(\wb{\theta}) , G^{bc}(\wb{\theta})  \right)_{t = 0} \longrightarrow  A =  \begin{cases}
        \sgn (g_{-1} - g_1)  \frac{3}{\sqrt{21}}  & \text{with prob. $\frac{1}{2}$}\,, \\
       - \sgn (g_{-1} - g_1)  \frac{3}{\sqrt{21}} & \text{with prob. $\frac{1}{2}$}\,.
      \end{cases}
\end{equation*}

\noindent 
{\bf Adam.} We now consider the deterministic version of the Adam optimizer \cite{kingma2014adam}, recalled below for completeness. Let $f(x)$ be a differentiable objective function on $\R^d$. The Adam iteration is defined by $z_n = T_{\gamma ,\alpha ,\beta}(n,z_{n - 1})$ 
for $z_n = (x_n, m_n,v_n) \in \R^d \times \R^d \times \R^d$ with $z_0 = (x_0, 0, 0)$, where 
\begin{equation*}
    T_{\gamma, \alpha, \beta}(n,z) = \begin{pmatrix}
        x - \frac{\gamma (1 - \alpha^n)^{- 1} ( \alpha m+ (1 - \alpha )\nabla f(x))}{\epsilon +(1 - \beta^n)^{- 1/2} \sqrt{\beta v+(1 - \beta )\nabla f(x)^{\odot 2}}} \\
        \alpha m+ (1 - \alpha )\nabla f(x) \\
        \beta v+ (1 - \beta )\nabla f(x)^{\odot 2}
        \end{pmatrix}\,.
\end{equation*}
We still consider the gradient conflict at the initialization, since the Adam dynamics is more complicated than the gradient flow one. From \cite{barakat2021convergence}, we have that starting from $(x_0,0,0) \in \R^{3d}$, the Adam dynamics at $t = 0$ satisfies $\dot{m}(0) \propto \na f(x_0)$, $\dot{v}(0) \propto \na f(x_0)^{\odot 2}$, and 
\begin{align*}
    \dot{x}(0) = - \frac{\na f(x_0)}{\eps + \sqrt{\na f(x_0)^{\odot 2}}} \underset{\eps = o(1)}{\approx}  - \frac{\na f(x_0)}{\sqrt{\na f(x_0)^{\odot 2}}}\,,
\end{align*}
indicating that at early iterations of Adam, 
the algorithm performance would be similar to the \emph{sign gradient
descent} \cite{balles2018dissecting}. Back to our problem \eqref{eq:losseg}, by the above discussion, if we apply Adam to the loss functions $\mc{L}_r(\theta)$ and $\mc{L}_{bc}(\theta)$, respectively, at the initialization, 
the normalized weights $\wb{\theta} = \ep^{-1} \theta$ will be updated along the directions:
\begin{align*}
    \frac{\na \mc{L}_r(\theta)}{\sqrt{\na \mc{L}_r(\theta)^{\odot 2}}} = \frac{G^r(\wb{\theta})}{\sqrt{G^r(\wb{\theta})^{\odot 2}}} + O(\ep^2)\,, \q  \frac{\na \mc{L}_{bc}(\theta)}{\sqrt{\na \mc{L}_{bc}(\theta)^{\odot 2}}} = \frac{G^{bc}(\wb{\theta})}{\sqrt{G^{bc}(\wb{\theta})^{\odot 2}}} + O(\ep^2)\,.
\end{align*}
Then, by \eqref{eq:effgrad1} and \eqref{eq:effgrad2}, one can compute 
\begin{align*}
    G^r_{\rm Adam}(\wb{\theta}) = \frac{G^r(\wb{\theta})}{\sqrt{G^r(\wb{\theta})^{\odot 2}}} = \sgn(\wb{\ba} \dd \wb{\bw}^{\odot 3})\left(\sgn(\wb{\bw}),\sgn(\wb{\ba}) \right),
\end{align*}
and 
\begin{align*}
    G^{bc}_{\rm Adam}(\wb{\theta}) = \frac{G^{bc}(\wb{\theta})}{\sqrt{G^{bc}(\wb{\theta})^{\odot 2}}} = \sgn(g_{-1} - g_1)\left(\sgn(\wb{\bw}),\sgn(\wb{\ba}) \right).
\end{align*}
It follows that the  alignment score is 
\begin{align*}
    \mc{A}\left( G^r_{\rm Adam}(\wb{\theta}) , G^{bc}_{\rm Adam}(\wb{\theta})  \right)  = \sgn(\wb{\ba} \dd \wb{\bw}^{\odot 3}) \sgn(g_{-1} - g_1) =  \begin{cases}
        \sgn (g_{-1} - g_1)  & \text{with prob. $\frac{1}{2}$}\,, \\
       - \sgn (g_{-1} - g_1) & \text{with prob. $\frac{1}{2}$}\,,
      \end{cases}
\end{align*}
which holds for any two-layer NN with width $N$. 


\noindent 
{\bf Shampoo and SOAP.} We proceed to consider Shampoo \cite{gupta2018shampoo}, which is a second-order optimizer with 
Kronecker product preconditioners. 
For the reader's convenience, we recall the Shampoo iterations for training neural networks. Following the notation in Section \ref{app:soap_newton}, let $W_t, G_t \in \R^{m \times n}$ be the weight matrix and gradient matrix for a layer at time step $t$, respectively. Shampoo generates left and right preconditioners: 
\begin{align*}
    L_t = L_{t - 1} + G_t G_t^T\,, \quad R_t = R_{t - 1} + G_t^T G_t\,,
\end{align*}
and then updates the weight matrix by 
\begin{align*}
    W_{t + 1} = W_t - \eta L_t^{-1/4} G_t R_t^{-1/4}\,,
\end{align*}
with step size $\eta > 0$. If we disable the accumulation in the preconditioners and set $ L_t =  G_t G_t^T$ and $R_t = G_t^T G_t$, then the Shampoo optimizer is simplified to
\begin{align*}
    W_{t + 1} = W_t - \eta\, {\rm Shampoo}(G_t)\,, \q {\rm Shampoo}(G_t): = (G_t G_t^T)^{-1/4} G_t (G_t^T G_t)^{-1/4}\,,
\end{align*}
with ${\rm Shampoo}(G_t) = U_t V_t^T$, where $U_t$ and $V_t$ are from the reduced singular value decomposition of $G_t = U_t \Sigma_t V_t^T$. It is clear that if we apply Shampoo to $\mc{L}_r(\theta)$ or $\mc{L}_{bc}(\theta)$ with the two-layer NN \eqref{eq:twolayersol}, then under small initialization \cref{assp1}, at the initialization, the updates of the normalized weights $\wb{\theta} = (\wb{\ba},\wb{\bw}) = \ep^{-1} \theta$ would be 
\begin{align*}
   \wb{\ba} \leftarrow \wb{\ba} - \eta \, {\rm Shampoo}\left(G_a^{\textit{r or bc}}(\wb{\ba},\wb{\bw})\right),\q   \wb{\bw} \leftarrow \wb{\bw} - \eta \, {\rm Shampoo}\left(G_w^{\textit{r or bc}}(\wb{\ba},\wb{\bw})\right).
\end{align*}
Here $G_a^{\textit{r or bc}}, G_w^{\textit{r or bc}} \in \R^N$ are given in \eqref{eq:effgrad1} and \eqref{eq:effgrad2}. Moreover, note that for any vector $x \in \R^d$, ${\rm Shampoo}(x)$ is simply $x/\|x\|$. Therefore, we can compute the (effective) initial Shampoo gradient directions for $\mc{L}_r(\theta)$ and $\mc{L}_{bc}(\theta)$:
\begin{align*}
    G_{\rm Shampoo}^r(\wb{\theta}): = \left({\rm Shampoo}\left(G_a^{r}(\wb{\theta})\right),{\rm Shampoo}\left(G_a^r(\wb{\theta})\right)\right) = \sgn(\wb{\ba} \dd \wb{\bw}^{\odot 3}) \left(\frac{\wb{\bw}^{\odot 3}}{\|\wb{\bw}^{\odot 3}\|},\frac{ \wb{\ba} \odot \wb{\bw}^{\odot 2}}{\| \wb{\ba} \odot \wb{\bw}^{\odot 2}\|} \right)\,.
\end{align*}
and 
\begin{align*}
    G_{\rm Shampoo}^{bc}(\wb{\theta}): = \left({\rm Shampoo}\left(G_a^{bc}(\wb{\theta})\right),{\rm Shampoo}\left(G_a^{bc}(\wb{\theta})\right)\right) = \sgn(g_{-1} - g_1)\left(\frac{\wb{\bw}}{\|\wb{\bw}\|},\frac{\wb{\ba}}{\|\wb{\ba}\|}\right)\,.
\end{align*}
It follows that the  alignment score is, as $N \to \infty$, 
\begin{align*}
    \mc{A}\left( G^r_{\rm Shampoo}(\wb{\theta}) , G^{bc}_{\rm Shampoo}(\wb{\theta})  \right) \longrightarrow C_{{\rm Shampoo}}   \begin{cases}
        \sgn (g_{-1} - g_1)  & \text{with prob. $\frac{1}{2}$}\,, \\
       - \sgn (g_{-1} - g_1) & \text{with prob. $\frac{1}{2}$}\,.
      \end{cases}
\end{align*}

We finally consider SOAP. Following the notations in \cref{app:soap_newton}, if $G_t$ is a vector, then $\widetilde{G}_t = [1,0,\cdots, 0]^\top$ and $\operatorname{Adam}(\widetilde{G}_t) = \widetilde{G}_t$. We transform this gradient back and obtain $G_t/\norm{G_t}$. It means that at initialization, 
\begin{equation*}
    {\rm Shampoo}\left(G_a^{\textit{r or bc}}(\wb{\ba},\wb{\bw})\right) =  {\rm SOAP}\left(G_a^{\textit{r or bc}}(\wb{\ba},\wb{\bw})\right)\,.
\end{equation*}
Therefore, the initial gradient conflict of SOAP follows from the case of Shampoo. 

\end{proof}

\section{Inter-step gradient alignment of preconditioned gradient descent}
\label{app:preconditioned_gd}

\textbf{Proposition 4} 
{\it Let $\mathcal{L}(\theta)$ be a twice differentiable loss function with Hessian $H(\theta)$, and let $P(\theta)$ be a positive definite preconditioner with uniformly bounded inverse $P^{-1}(\theta)$. Consider the preconditioned gradient descent update with exponent $0 \leq s \leq 1$ and learning rate $\eta > 0$:
\begin{align}
\theta_{t+1} = \theta_t - \eta P^{-s}(\theta_t) \nabla \mathcal{L}(\theta_t)
\end{align}

For consecutive gradient vectors $g_t = \nabla \mathcal{L}(\theta_t)$ and $g_{t+1} = \nabla \mathcal{L}(\theta_{t+1})$, the alignment score $\mathcal{A}(g_t, g_{t+1}) = \frac{g_t^T g_{t+1}}{\|g_t\|\|g_{t+1}\|}$ satisfies:
\begin{align}
\mathcal{A}(g_t, g_{t+1}) = 1 - \frac{\eta^2}{2}\frac{\|HP^{-s}g_t\|^2}{\|g_t\|^2} + O(\eta^3)
\end{align}

Furthermore, to ensure gradient alignment $\mathcal{A}(g_t, g_{t+1}) \geq 1-\epsilon$ for some $\epsilon > 0$, the learning rate must satisfy:
\begin{align}
\eta \leq \sqrt{\frac{2\epsilon\|g_t\|^2}{\|HP^{-s}g_t\|^2}}
\end{align}

This bound specializes to the following cases:
\begin{enumerate}
\item Vanilla Gradient Descent $(P = I)$: $\eta_{\max} = \sqrt{\frac{2\epsilon}{\lambda_{\max}^2(H)}}$
\item Newton's Method $(P = H, s = 1)$: $\eta_{\max} = \sqrt{2\epsilon}$
\end{enumerate}}

\begin{proof}
    Let $P(\theta)$ be a positive definite preconditioner with uniformly bounded inverse $P^{-1}(\theta)$. For learning rate $\eta>0$ and exponent $0<s \leq 1$, the parameter update is:

\begin{align}
    \theta_{t+1}=\theta_t-\eta P^{-s}\left(\theta_t\right) g_t
\end{align}
where $g_t=\nabla \mathcal{L}\left(\theta_t\right)$.

We are interested in the behavior of the alignment between successive gradients under this update. To study this, we expand $g_{t+1} = \nabla \mathcal{L}(\theta_{t+1})$ via a second-order Taylor series around $\theta_t$, using the Hessian $H(\theta_t) = \nabla^2 \mathcal{L}(\theta_t)$:
\begin{align}
g_{t+1}=g_t+H\left(\theta_t\right) \Delta \theta_t+\frac{1}{2} D^3 \mathcal{L}\left(\theta_t\right)\left[\Delta \theta_t, \Delta \theta_t\right]+O\left(\left\|\Delta \theta_t\right\|^3\right)
\end{align}
Substituting $\Delta \theta_t$:
\begin{align}
    g_{t+1}=g_t-\eta H P^{-s} g_t+\frac{\eta^2}{2} D^3 \mathcal{L}\left(\theta_t\right)\left[P^{-s} g_t, P^{-s} g_t\right]+O\left(\eta^3\left\|P^{-s} g_t\right\|^3\right) .
\end{align}

Assuming $\eta$ is small, we retain only the first-order term:
\begin{align}
    g_{t+1} = g_t-\eta H P^{-s} g_t+O\left(\eta^2\left\|P^{-s} g_t\right\|^2\right)
\end{align}
We compute the inner product between the current and next gradient:
\begin{align}
&g_t^{\top} g_{t+1} = \left\|g_t\right\|^2-\eta g_t^{\top} H P^{-s} g_t+O\left(\eta^2\left\|P^{-s} g_t\right\|^2\left\|g_t\right\|\right).
\end{align}
We also expand the norm of the new gradient:
\begin{align}
   \left\|g_{t+1}\right\|^2=\left\|g_t\right\|^2-2 \eta g_t^{\top} H P^{-s} g_t+\eta^2\left\|H P^{-s} g_t\right\|^2+O\left(\eta^3\right) .
\end{align}
Taking the square root and using the approximation $\sqrt{1 - x} \approx 1 - \frac{x}{2}$ for small $x$, we get:
\begin{align}
\left\|g_{t+1}\right\|=\left\|g_t\right\|\left(1-\eta \frac{g_t^{\top} H P^{-s} g_t}{\left\|g_t\right\|^2}+\frac{\eta^2}{2} \frac{\left\|H P^{-s} g_t\right\|^2}{\left\|g_t\right\|^2}\right)+O\left(\eta^3\right) .
\end{align}
The cosine similarity (alignment score) between $g_t$ and $g_{t+1}$ is defined as:
\begin{align}
    \mathcal{A}\left(g_t, g_{t+1}\right)=\frac{g_t^{\top} g_{t+1}}{\left\|g_t\right\|\left\|g_{t+1}\right\|}
\end{align}
Substituting the approximations above:
\begin{align}
   \mathcal{A}\left(g_t, g_{t+1}\right)=\frac{\left\|g_t\right\|^2-\eta g_t^{\top} H P^{-s} g_t+O\left(\eta^2\right)}{\left\|g_t\right\|^2\left(1-\eta \frac{g_t^{\top} H P^{-s} g_t}{\left\|g_t\right\|^2}+\frac{\eta^2}{2} \frac{\left\|H P^{-s} g_t\right\|^2}{\left\|g_t\right\|^2}+O\left(\eta^3\right)\right)} .
\end{align}
Factor $\|g_t\|_2^2$ in denominator:
\begin{align}
\mathcal{A}\left(g_t, g_{t+1}\right)=\frac{1-\eta \frac{g_t^{\top} H P^{-s} g_t}{\left\|g_t\right\|^2}+O\left(\eta^2\right)}{1-\eta \frac{g_t^{\top} H P^{-s} g_t}{\left\|g_t\right\|^2}+\frac{\eta^2}{2} \frac{\left\|H P P^{-s} g_t\right\|^2}{\left\|g_t\right\|^2}+O\left(\eta^3\right)} .
\end{align}
Using $(1-a+b)^{-1}=1+a-b+O\left(a^2+b^2\right)$ for small $a, b$ :
\begin{align}
    \mathcal{A}\left(g_t, g_{t+1}\right)=\left(1-\eta \frac{g_t^{\top} H P^{-s} g_t}{\left\|g_t\right\|^2}\right)\left(1+\eta \frac{g_t^{\top} H P^{-s} g_t}{\left\|g_t\right\|^2}-\frac{\eta^2}{2} \frac{\left\|H P^{-s} g_t\right\|^2}{\left\|g_t\right\|^2}\right)+O\left(\eta^3\right) .
\end{align}
Multiplying and simplifying to $O\left(\eta^2\right)$ :
\begin{align}
    \mathcal{A}\left(g_t, g_{t+1}\right)=1-\frac{\eta^2}{2} \frac{\left\|H P^{-s} g_t\right\|^2}{\left\|g_t\right\|^2}+O\left(\eta^3\right)
\end{align}

Thus, the alignment between gradients remains close to 1 if the second-order term is small. To ensure the gradients stay well-aligned, i.e., $\mathcal{A}(g_t, g_{t+1}) \geq 1 - \epsilon$ for a given tolerance $\epsilon > 0$, we require:
\begin{align}
    \frac{\eta^2}{2} \frac{\left\|H P^{-s} g_t\right\|^2}{\left\|g_t\right\|^2} \leq \epsilon
\end{align}
Solving for the maximum permissible learning rate $\eta$ yields:
\begin{align}
    \eta \leq \sqrt{\frac{2 \epsilon\left\|g_t\right\|^2}{\left\|H P^{-s} g_t\right\|^2}}
\end{align}
We now examine this upper bound in a few special cases of interest:

1. Vanilla GD $(P=I)$ :
\begin{align}
    \eta_{\max }=\sqrt{\frac{2 \epsilon}{\lambda_{\max }^2(H)}}
\end{align}
where $\lambda_{\max }(H)$ is the largest eigenvalue of $H$.

2. Newton's Method ( $P=H, s=1$ ):
\begin{align}
    H P^{-1}=I \Longrightarrow \sigma_{\max }\left(H P^{-1}\right)=1 \Longrightarrow \eta_{\max }=\sqrt{2 \epsilon}
\end{align}


\end{proof}

\section{Connection between SOAP and Newton's method}  \label{app:soap_newton}

SOAP \cite{vyas2024soap} enhances Shampoo's efficiency by performing optimization in a transformed space aligned with the preconditioner's principal directions. For each layer's weight matrix and gradient $G_t \in \R^{m \times n}$, SOAP maintains two covariance matrices using exponential moving averages:
\begin{align}
    L_{t} = \beta_2 L_{t-1}+\left(1-\beta_2\right) G_{t} G_t^T, \\
    R_{t} =  \beta_2 R_{t-1} +\left(1-\beta_2\right) G_t^T G_t\,.
\end{align}
These matrices are then eigendecomposed as $L_t = Q_L \Lambda_L Q_L^T$ and $R_t = Q_R \Lambda_R Q_R^T$, where $\Lambda_L$ and $\Lambda_R$ contain the eigenvalues that capture the principal curvature directions of the loss landscape.

At each iteration $t$, SOAP updates each layer's weight matrix $W_t$ using its corresponding gradient $G_t$ as follows:
\begin{enumerate}
    \item Project the weight and gradient into the eigenspace: 
    \vspace{-0.5em}
    \[\widetilde{W}_t = Q_L^T W_t Q_R, \widetilde{G}_t = Q_L^T G_t Q_R.\]
    \vspace{-2em}
    \item Apply the Adam update in the \emph{rotated} space:
    \vspace{-0.5em}
    \[\widetilde{W}_{t+1} = \widetilde{W}_{t} - \eta \, \operatorname{Adam}(\widetilde{G}_t).\]
    \vspace{-2em}
    \item Transform back to the original parameter space:
    \vspace{-0.5em}
    \[W_{t+1} = Q_L \widetilde{W}_{t+1} Q_R^T.\]
    \vspace{-2em}
\end{enumerate}
To reduce computational overhead, the preconditioners $L_t$ and $R_t$ are updated with frequency $f$ in practice. We will analyze the impact of update frequency and momentum parameters through ablation studies in Section \ref{sec:results}.

Before diving into the formal analysis, let us  build some intuition for why SOAP is particularly effective at resolving gradient conflicts. The key insight comes from understanding how second-order information captures interactions between different loss terms. When gradients conflict, it typically indicates that improving one objective requires coordinated changes across multiple parameters -- information that is encoded in the Hessian matrix's off-diagonal elements.

SOAP approximates this second-order information in two complementary ways:
(i) Its block-diagonal structure naturally captures parameter interactions within each network layer; (ii) Its adaptive preconditioner accumulates information about gradient correlations across training steps.
This allows SOAP to implicitly identify and exploit parameter update directions that simultaneously improve multiple objectives. Rather than simply following the average gradient, SOAP can utilize the local loss landscape geometry to find more direct paths to good solutions. The following sections make this intuition precise through formal analysis of SOAP's convergence properties and gradient alignment characteristics.

To establish SOAP's connection to Newton's method, we begin by examining how the Hessian matrix can be approximated in neural networks. We limit our analysis with networks trained with cross-entropy loss.
Here the Gauss-Newton approximation takes the form:
\begin{align}
H_{\text{GN}} = \mathbb{E}\left[\frac{\partial f}{\partial W} \frac{\partial^2 \mathcal{L}}{\partial f^2} \frac{\partial f^T}{\partial W}\right] = \mathbb{E}\left[g g^T\right],
\end{align}
where $\mathcal{L}$ denotes the loss function, $f$ represents network outputs, and $G = \frac{\partial \mathcal{L}}{\partial W}$ is the gradient matrix with vectorization $g = \Vec(G)$. Empirical evidence from \cite{sankar2021deeper} supports a key simplifying assumption:
\begin{assumption}
\label{assumption}
The Gauss-Newton component provides a good approximation to the true Hessian: $H_{\text{GN}} \approx H$. 
\end{assumption}

For our purpose, we begin by noting that there exists a one-to-one correspondence between the original parameter space and the rotated space that preserves the matrix-vector multiplication. 
\begin{lemma}
\label{lemma: equivalence}
Let $Q_L \in \R^{m \times m}$ and $Q_R \in \R^{n \times n}$ be two orthogonal matrices. For any matrix $A \in \R^{mn \times mn}$ and vector $v \in \R^{m n}$, define $\widetilde{v} := (Q_L \otimes Q_R)v$ and $\widetilde{A} := (Q_L\otimes Q_R)A(Q_L^T \otimes Q_R^T)$. Then there holds 
\begin{equation*}
    \widetilde{A v} = (Q_L \otimes Q_R) A v = \widetilde{A} \widetilde{v}\,.
\end{equation*}
\end{lemma}
The proof follows directly by applying the transformation $Q_L \otimes Q_R$ to $A v$ and the definitions of $\widetilde{A}$ and $\widetilde{v}$. Building on the above lemma, one can easily transform the preconditioned gradient descent in the original space to the rotated one and vice versa. 

we can now establish the equivalence between the preconditioned gradient descent in the original and rotated spaces.


\begin{corollary} 
\label{corollary: rotated}
Let $W_t, G_t \in \R^{m \times n}$ be the weight matrix and gradient matrix for a layer at iteration $t$, respectively, with vectorizations $w_t = \Vec(W_t)$ and $g_t = \Vec(G_t)$. The preconditioned gradient descent update:
\begin{align}
w_{t+1} = w_t - \eta P^{-1}g_t\,,
\end{align}
is equivalent to performing preconditioning in the rotated space:
\begin{align}
\widetilde{w}_{t+1} = \widetilde{w}_t - \eta \widetilde{P}^{-1}\widetilde{g}_t\,,
\end{align}
where $P \in \R^{mn \times mn}$ is the preconditioner, and $\widetilde{w}$, $\widetilde{g}$, and $\widetilde{P}$ are the rotated weight, gradient, and preconditioner defined by the transformations in \cref{lemma: equivalence}. 
\end{corollary}

\begin{proposition}
\label{prop2}
Let $L_t = \mathbb{E}\left[G_tG_t^T\right]$ and $R_t = \mathbb{E}\left[G_t^T G_t\right]$ have eigendecompositions
$L_t = Q_L \Lambda_L Q_L^T$ and $R_t = Q_R \Lambda_R Q_R^T$. Under the assumption of \cref{lemma: exact_h_gn},
 the equivalent preconditioner in the rotated space is diagonal, i.e., $\widetilde{H}_{\text{GN}} = \operatorname{diag}(\widetilde{H}_{\text{GN}})$.
\end{proposition}

\begin{proof}
    The proof follows from the combination of Lemma \ref{lemma: equivalence} and Lemma \ref{lemma: exact_h_gn}. First, we express $H_{\text{GN}}$ using the Kronecker approximation:
\begin{align}
H_{\text{GN}} = L_t^{1/2} \otimes R_t^{1/2} / \operatorname{Tr}\left(\mathbb{E}\left[GG^T\right]\right).
\end{align}
Then, we derive the rotated preconditioner:
\begin{align*}
\widetilde{H}_{\text{GN}} &= \left(Q_L \otimes Q_R\right) H_{\text{GN}} \left(Q_L^T \otimes Q_R^T\right) \\
&= \left(Q_L \otimes Q_R\right) (L_t^{1/2} \otimes R_t^{1/2}) \left(Q_L^T \otimes Q_R^T\right) / \operatorname{Tr}\left(\mathbb{E}\left[GG^T\right]\right) \\
&= (Q_L L_t^{1/2} Q_L^T) \otimes (Q_R R_t^{1/2} Q_R^T) / \operatorname{Tr}\left(\mathbb{E}\left[\Lambda_L\right]\right) \\
&= \Lambda_L^{1/2} \otimes \Lambda_R^{1/2} / \operatorname{Tr}\left(\mathbb{E}\left[\Lambda_L\right]\right).
\end{align*}
The final expression shows that $\widetilde{H}_{\text{GN}}$ is diagonal, as it is the Kronecker product of diagonal matrices scaled by a scalar factor.
\end{proof}

Finally, we connect our analysis to Adam's update rule by adapting the following result from Molybog et al \cite{molybog2023theory}: 
\begin{proposition}[Adapt from \cite{molybog2023theory}]
\label{prop:adam_hessian}
Suppose that $\theta^*$ is a local minimum and assume that $\theta - \theta^* \sim \mathcal{N}(0, \sigma^2 I)$. For Adam update rule denoted by $\theta_{t+1} = \theta_t - \eta \operatorname{Adam}(g_t)$, 
we have
\begin{align}
    \operatorname{Adam}(g_t) \approx \operatorname{diag}\left( H \right)^{-1} g_t.
\end{align}
\end{proposition}

\begin{proof}
The Adam optimizer follows the update rule:
\begin{align*}
&m_t = \beta_1 m_{t-1} + (1-\beta_1) g_t, \\
&v_t = \beta_2 v_{t-1} + (1-\beta_2) g_t \odot g_t ,\\
&\hat{m}_t = m_t/(1-\beta_1^t), \\
&\hat{v}t = v_t/(1-\beta_2^t), \\
&\theta_t = \theta{t-1} - \eta \hat{m}_t/(\sqrt{\hat{v}_t}+\epsilon).
\end{align*}
Taking a first-order Taylor expansion of the gradient around a local minimum $\theta^*$:
\begin{align*}
    g_\theta \approx g_{\theta^*}+   H_{\theta^*}\left(\theta-\theta^*\right) \approx    H_{\theta^*}\left( {\theta}- {\theta}^*\right).
\end{align*}
This yields
\begin{align*}
      {g}_\theta   {g}_{  {\theta}}^{\top} \approx    H_{\theta^*}\left(  {\theta}-  {\theta}^*\right)\left(  {\theta}-  {\theta}^*\right)^{\top}    H_{  {\theta}^*}^{\top}.
\end{align*}

Under our assumption that $\theta - \theta^* \sim \mathcal{N}(0, \sigma^2 I)$,
\begin{align*}
    \mathbb{E}\left[   {g}_\theta    {g}_{   {\theta}}^{\top}\right] \approx    H_{   {\theta}} \mathbb{E}\left[\left(   {\theta}-   {\theta}^*\right)\left(   {\theta}-   {\theta}^*\right)^{\top}\right]    H_{   {\theta}^*}^{\top}=\sigma^2    H_{\theta^*}    H_{   {\theta}^*}^{\top}.
\end{align*}

By construction, $v_t$ approximates the diagonal of $\mathbb{E}_{\theta \sim \theta\tau}[g_\theta g_\theta^T]$, where $\theta_\tau$ represents the distribution of model weights over the past $O(1/(1-\beta_2))$ steps:
\begin{align*}
    v_t \approx \operatorname{diag}\left(\mathbb{E}_{\theta \sim \theta_\tau}  \left[ g_\theta g_\theta^T \right] \right) \approx \sigma^2  \text{diag}(   H_{\theta^*}^2).
\end{align*}
Finally, assuming $m_t \approx g_t$:
\begin{align*}
    \operatorname{Adam}\left(g_t\right) \approx \frac{m_t}{\sqrt{v_t}+\varepsilon} \approx \frac{m_t}{\sqrt{v_t}}
    \approx \operatorname{diag}(H)^{-1} g_t.
\end{align*}
\end{proof}

\begin{theorem} 
\label{thm: soap}
Under assumption of Proposition \ref{prop:adam_hessian}, SOAP's update approximates Newton's method:
\begin{align}
    w_{t+1} = w_{t} - \eta \operatorname{Soap}(g_t) \approx w_t - \eta H^{-1} g_t.
\end{align}
\end{theorem}
\begin{proof}
    Combining Propositions \ref{prop2} and \ref{prop:adam_hessian}, we obtain
\begin{align}
\operatorname{Adam}(\widetilde{G}_t) \approx \operatorname{diag}(\widetilde{H})^{-1} \widetilde{g}_t \approx \operatorname{diag}(\widetilde{H}_{\text{GN}})^{-1} \widetilde{g}_t = \widetilde{H}_{\text{GN}}^{-1} \widetilde{g}_t \approx \widetilde{H}^{-1} \widetilde{g}_t.
\end{align}
By  Corollary \ref{corollary: rotated}, this is equivalent to  the Newton update in the original space:
\begin{align}
w_{t+1} = w_t - H^{-1}g_t.
\end{align}
As a direction implication, the Hessian matrix is approximately diagonal in the rotated space.
\end{proof}

The key insight is that SOAP effectively approximates the block-diagonal Gaussian Newton component of  Hessian in a rotated space, with each block corresponding to a layer-wise Kronecker factorization. This structure naturally promotes gradient alignment across optimization steps, as we demonstrated theoretically in Proposition \ref{prop: newton_alignment_score} and observed empirically in Figure \ref{fig:grad_align_score}.

\begin{remark} \label{rem1}
While SOAP effectively approximates Newton's method through its block-diagonal structure, other optimizers make different compromises in their approximations. Adam can approximate Newton's method, but requires the highly restrictive assumption that the Hessian matrix is diagonal. Similarly, Shampoo takes a different approach by using the square root of the Gauss-Newton component as its preconditioner \cite{morwani2024new}:
\begin{align}
    \operatorname{Shampoo}(g_t) \approx H_{\text{GN}}^{-1/2} g_t \approx H^{-1/2} g_t.
\end{align}
\end{remark}
These structural differences help explain why SOAP achieves better gradient alignment than both Adam and Shampoo.

\section{Connection of Shampoo and Muon to Quasi-second-order Methods}
\label{app:connection_shampoo_muon}

We review the connections between Shampoo, Muon, and quasi-second-order optimization methods, building upon results from \cite{anil2020scalable} and \cite{morwani2024new}. This section aims to provide a self-contained exposition of these relationships.

\paragraph{Adagrad.} Adagrad is a preconditioned online learning algorithm that leverages the accumulated covariance of gradients as a preconditioner. Let $\theta_t \in \mathbb{R}^p$ denote the parameters at time $t$ and $g_t \in \mathbb{R}^p$ denote the corresponding gradient. Adagrad maintains a preconditioner $H_{\text{Ada}} = \sum_{t=1}^T g_t g_t^{\top}$. The parameter update with learning rate $\eta$ is given by:
\begin{align}
    \theta_{T+1} = \theta_T - \eta H_{\mathrm{Ada}}^{-1/2} g_T
\end{align}

\paragraph{Shampoo.} Shampoo tracks two statistical matrices throughout training, $L_t \in \mathbb{R}^{m \times m}$ and $R_t \in \mathbb{R}^{n \times n}$, defined as:
\begin{align}
    L_t = \epsilon I_m + \sum_{s=1}^t G_s G_s^{\top}; \quad R_t = \epsilon I_n + \sum_{s=1}^t G_s^{\top} G_s
\end{align}
where $G_s \in \mathbb{R}^{m \times n}$ is the gradient matrix at step $s$, and $\epsilon > 0$ is a small constant for numerical stability.

The full matrix Adagrad preconditioner $H_t$ can be approximated as $\left(L_t \otimes R_t\right)^{1/2}$. This approximation transforms the Adagrad update rule $w_{t+1} = w_t - \eta H_t^{-1/2} g_t$ into the Shampoo update rule for parameter matrix $W$:
\begin{align}
    W_{t+1} = W_t - \eta L_t^{-1/4} G_t R_t^{-1/4}
\end{align}

The theoretical foundation for this approximation is provided by the following lemma:

\begin{lemma}[\cite{anil2020scalable}, Lemma 1]
    \label{lemma:hessian}
    Let $G_1, \ldots, G_t \in \mathbb{R}^{m \times n}$ be matrices of rank at most $r$. Let $g_s = \operatorname{vec}\left(G_s\right)$ and define $\widehat{H}_t = \epsilon I_{mn} + \sum_{s=1}^t g_s g_s^{\top}$. Define $L_t, R_t$ as above: $L_t = \epsilon I_m + \sum_{s=1}^t G_s G_s^{\top}$, $R_t = \epsilon I_n + \sum_{s=1}^t G_s^{\top} G_s$. Then for any $p, q > 0$ such that $1/p + 1/q = 1$, we have $\widehat{H}_t \leq r L_t^{1/p} \otimes R_t^{1/q}$.
\end{lemma}
It follows from this lemma that for any $p, q > 0$ with $1/p + 1/q = 1$, the full Adagrad preconditioned gradient $H_{\text{Ada}}^{-1/2}g_t$ can be approximated by $(L_t^{1/p} \otimes R_t^{1/q})^{-1/2} g_t = \text{vec}(L_t^{-1/2p} G_t R_t^{-1/2q})$. The case where $p = q = 2$ yields the standard Shampoo update.

Moreover, \cite{morwani2024new} explored the Hessian approximation perspective of Shampoo, demonstrating that the preconditioner in Shampoo is a Kronecker product approximation of the Gauss-Newton component of the layerwise Hessian (see \cref{rem1}).

\paragraph{Muon.} The Muon optimizer \cite{jordan2024muon} was recently proposed for optimizing neural network weights representable as matrices. At iteration $t$, given current weight $\mathbf{W}_{t-1}$, momentum $\mu$, learning rate $\eta_t$, and objective $\mathcal{L}_t$, the update rule of the Muon optimizer is:
\begin{align}
\mathbf{M}_t &= \mu \mathbf{M}_{t-1} + \nabla \mathcal{L}_t\left(\mathbf{W}_{t-1}\right), \\
\mathbf{O}_t &= \text{Newton-Schulz}\left(\mathbf{M}_t\right), \\
\mathbf{W}_t &= \mathbf{W}_{t-1} - \eta_t \mathbf{O}_t.
\end{align}
Here, $\mathbf{M}_t$ is the momentum of gradient at iteration $t$, initialized as a zero matrix when $t = 0$. The Newton-Schulz iteration process is adopted to approximately compute $(\mathbf{M}_t \mathbf{M}_t^{\mathrm{T}})^{-1/2} \mathbf{M}_t$.

When preconditioner accumulation is removed, we can observe that the update simplifies to \cite{jordan2024muon}:
\begin{align}
W_{t+1} &= W_t - \eta(G_t G_t^{\top})^{-1/4} G_t (G_t^{\top} G_t)^{-1/4} \\
&= W_t - \eta(U S^2 U^{\top})^{-1/4}(U S V^{\top})(V S^2 V^{\top})^{-1/4} \\
&= W_t - \eta(U S^{-1/2} U^{\top})(U S V^{\top})(V S^{-1/2} V^{\top}) \\
&= W_t - \eta U S^{-1/2} S S^{-1/2} V^{\top} \\
&= W_t - \eta U V^{\top}
\end{align}
From this derivation, Muon can be viewed as approximating the use of $H_{\text{Ada}}^{-1/2}$ as a preconditioner, with additional orthogonalization benefits.

\section{Additional Lemma and Proof}

\begin{lemma} [\cite{morwani2024new}, Corollary 2]
\label{lemma: exact_h_gn}
Under the assumption that the reshaping of the Hessian tensor $H_{GN}$ is rank-1,
$$
\hat{H}_{GN} =\left(\mathbb{E}\left[G G^{\top}\right] \otimes \mathbb{E}\left[G^{\top} G\right]\right) / \operatorname{Tr}\left(\mathbb{E}\left[G G^{\top}\right]\right).
$$
\end{lemma}

\subsection{Proof of \cref{prop1}}
\label{proof: prop1}
\begin{proof}
    When $n=2$, we note 
\begin{align*}
\mathcal{A}(v_1, v_2) &= 2\left\|\frac{\frac{v_1}{\|v_1\|} + \frac{v_2}{\|v_2\|}}{2}\right\|^2 - 1 \notag \\
&= \frac{1}{2}\left(\left\|\frac{v_1}{\|v_1\|}\right\|^2 + 2\frac{v_1 \dd v_2}{\|v_1\|\|v_2\|} + \left|\frac{v_2}{\|v_2\|}\right|^2\right) - 1  \notag \\
&= \frac{1}{2}(1 + 2\cos(v_1, v_2) + 1) - 1  = \cos(v_1, v_2)\,. \qedhere
\end{align*} 
\end{proof}

\subsection{Proof of \cref{prop_alignment}}

\begin{proof}
\textbf{(i) Single-step drop.}
By $L$-smoothness (descent lemma),
\[
\mathcal L(\theta_{t+1})
\le \mathcal L(\theta_t) + \langle g_t,\theta_{t+1}-\theta_t\rangle + \tfrac{L}{2}\|\theta_{t+1}-\theta_t\|^2.
\]
With $\theta_{t+1}-\theta_t=-\eta h_t$,
\[
\Delta_t \;=\; \mathcal L(\theta_t)-\mathcal L(\theta_{t+1})
\;\ge\; \eta\langle g_t,h_t\rangle - \tfrac{L\eta^2}{2}\|h_t\|^2.
\]
Since $h_t=P_t g_t$ and $P_t^{-1}\succeq \tfrac{1}{M}I$, we have
\[
\langle g_t,h_t\rangle = h_t^\top P_t^{-1} h_t \;\ge\; \tfrac{1}{M}\|h_t\|^2,
\]
hence
\[
\Delta_t \;\ge\; \Big(\tfrac{\eta}{M}-\tfrac{L\eta^2}{2}\Big)\|h_t\|^2.
\]
Under the equal-norm assumption $\|h_t^i\|=\lambda$, let $u_t^i:=h_t^i/\lambda$. By the chosen definition of $A_t^{\mathrm{intra}}$,
\[
\Big\|\sum_{i=1}^n u_t^i\Big\|^2=\tfrac{n^2}{2}\big(A_t^{\mathrm{intra}}+1\big).
\]
Since $h_t=\sum_i h_t^i=\lambda\sum_i u_t^i$ and $\sum_i\|h_t^i\|^2=n\lambda^2$, we obtain the displayed equal-norm formulas and the refined bound.

\medskip
\textbf{(ii) Two-step cumulative drop.}
Apply $L$-smoothness with $\theta=\theta_{t-1}$, $\phi=\theta_{t+1}$:
\[
\mathcal L(\theta_{t+1}) \le \mathcal L(\theta_{t-1})
+ \langle g_{t-1},\theta_{t+1}-\theta_{t-1}\rangle
+ \tfrac{L}{2}\|\theta_{t+1}-\theta_{t-1}\|^2.
\]
Since $\theta_{t+1}-\theta_{t-1}=-\eta(h_{t-1}+h_t)$, we get
\[
\Delta_{t-1}+\Delta_t
\;\ge\; \eta\langle g_{t-1},h_{t-1}+h_t\rangle
- \tfrac{L\eta^2}{2}\|h_{t-1}+h_t\|^2.
\]
For the linear term, using $P_{t-1}^{-1}\succeq \tfrac{1}{M}I$,
\[
\langle g_{t-1},h_{t-1}\rangle
= h_{t-1}^\top P_{t-1}^{-1} h_{t-1} \;\ge\; \tfrac{1}{M}\|h_{t-1}\|^2,
\]
and
\[
\langle g_{t-1},h_t\rangle
= h_{t-1}^\top P_{t-1}^{-1} h_t \;\ge\; \tfrac{1}{M}\langle h_{t-1},h_t\rangle.
\]
Let $a:=\|h_{t-1}\|$, $b:=\|h_t\|$, and
$A_t^{\mathrm{inter}}:=\frac{\langle h_{t-1},h_t\rangle}{ab}$. Then
\[
\eta\langle g_{t-1},h_{t-1}+h_t\rangle
\;\ge\; \tfrac{\eta}{M}\big(a^2+ab\,A_t^{\mathrm{inter}}\big).
\]
For the quadratic term,
\(
\|h_{t-1}+h_t\|^2=a^2+2ab\,A_t^{\mathrm{inter}}+b^2.
\)
Putting the bounds together,
\[
\Delta_{t-1}+\Delta_t \;\ge\;
\tfrac{\eta}{M}\big(a^2+ab\,A_t^{\mathrm{inter}}\big)
-\tfrac{L\eta^2}{2}\big(a^2+2ab\,A_t^{\mathrm{inter}}+b^2\big),
\]
which rearranges to the stated inequality. The coefficient of $A_t^{\mathrm{inter}}$ equals
$\tfrac{\eta}{M}(1-L\eta M)\ge 0$ for $\eta\le 1/(LM)$, hence monotonicity.
\end{proof}


\section{Experimental Details}
\label{appendix: experiments}

\subsection{Architectures} 
\label{appendix:arch}

This section outlines the network architectures employed in our work, along with the enhancements introduced to improve their performance.

\paragraph{Modified MLP.} The modified MLP architecture is proposed by \cite{wang2021understanding}, which has been extensively used in the literature \cite{daw2022rethinking,wang2022respecting, anagnostopoulos2023residual,chen2024self,karakonstantis2024room,zhangphysics} due to its improved capability in learning complex PDE solutions. The network processes input coordinates through two parallel encoders:
\begin{align}
    \mathbf{U}=\sigma\left(\mathbf{W}_1 \mathbf{x}+\mathbf{b}_1\right), \quad \mathbf{V}=\sigma\left(\mathbf{W}_2 \mathbf{x}+\mathbf{b}_2\right).
\end{align}
Then, for $l=1,2, \ldots, L$, the forward pass is defined as:
\begin{align}
    & \mathbf{f}^{(l)}(\mathbf{x})=\mathbf{W}^{(l)} \cdot \mathbf{g}^{(l-1)}(\mathbf{x})+\mathbf{b}^{(l)}, \\
& \mathbf{g}^{(l)}(\mathbf{x})=\sigma\left(\mathbf{f}_\theta^{(l)}(\mathbf{x})\right) \odot \mathbf{U}+\left(1-\sigma\left(\mathbf{f}_\theta^{(l)}(\mathbf{x})\right)\right) \odot \mathbf{V}
\end{align}
The final network output is given by
\begin{align}
    \mathbf{f}_\theta(\mathbf{x})=\mathbf{W}^{(L+1)} \cdot \mathbf{g}^{(L)}(\mathbf{x})+\mathbf{b}^{(L+1)}.
\end{align}
where $\sigma$ is a nonlinear activation function, $\odot$ denotes element-wise multiplication, and the trainable parameters are:
\begin{align}
    \theta=\left\{\mathbf{W}_1, \mathbf{b}_1, \mathbf{W}_2, \mathbf{b}_2,\left(\mathbf{W}^{(l)}, \mathbf{b}^{(l)}\right)_{l=1}^{L+1}\right\}.
\end{align}
This architecture extends the standard MLP by incorporating dual input encoders and merging their features through point-wise multiplication at each hidden layer. While computationally more demanding, this modification demonstrates superior performance in minimizing PDE residuals compared to standard MLPs.

\paragraph{PirateNet.}  PirateNet is proposed by   \cite{wang2024piratenets}, which aims to enable stable and efficient training of deep PINN models.  The architecture first transforms input coordinates $\mathbf{x}$ into a high-dimensional feature space using random Fourier features \cite{tancik2020fourier}:
\begin{align*}
    \Phi(\mathbf{x})= \begin{bmatrix}
    \cos (\mathbf{B x} ) \\
    \sin (\mathbf{B x} )
    \end{bmatrix},
\end{align*}
where $\mathbf{B} \in \R^{m \times d}$ has entries sampled i.i.d. from $\mathcal{N}(0, s^2)$ with user-specified $s > 0$.  This embedding mitigates spectral bias in PINNs by improving the eigenfunction frequency of the Neural Tangent Kernel, enabling better learning of high-frequency components and multiscale features \cite{wang2021eigenvector}.

The embedded coordinates are processed through two dense layers that act as gates:
\begin{align*}
\mathbf{U} &= \sigma(\mathbf{W}_1 \Phi(\mathbf{x}) + \mathbf{b}_1  ), \quad
\mathbf{V} = \sigma(\mathbf{W}_2 \Phi(\mathbf{x}) + \mathbf{b}_2  ),
\end{align*}
where $\sigma$ is a point-wise activation function. This gating mechanism is essentially the same as in modified MLP.

Let $\mathbf{x}^{(1)} = \Phi(\mathbf{x})$ and $\mathbf{x}^{(l)}$ be the input to the $l$-th block ($1 \le l \le L$). Each block performs:
\begin{align}
    \mathbf{f}^{(l)}  &= \sigma\big(\mathbf{W}^{(l)}_1 \mathbf{x}^{(l)} + \mathbf{b}^{(l)}_1\big)\,, \label{eq: step1}  \\
    \mathbf{z}^{(l)}_1 &= \mathbf{f}^{(l)} \odot \mathbf{U} + (1 - \mathbf{f}^{(l)}) \odot \mathbf{V}\,,  \label{eq: gate1} \\
     \mathbf{g}^{(l)}  &= \sigma\big(\mathbf{W}^{(l)}_2 \mathbf{z}_1^{(l)} + \mathbf{b}^{(l)}_2\big)\,, \\
     \mathbf{z}^{(l)}_2 &= \mathbf{g}^{(l)} \odot \mathbf{U} + (1 - \mathbf{g}^{(l)}) \odot \mathbf{V}\,,  \label{eq: gate2} \\
      \mathbf{h}^{(l)}  &= \sigma\big(\mathbf{W}^{(l)}_3 \mathbf{z}_2^{(l)} + \mathbf{b}^{(l)}_3\big)\,, \\
    \mathbf{x}^{(l+1)} &= \alpha^{(l)}  \mathbf{h}^{(l)} + (1 - \alpha^{(l)})   \mathbf{x}^{(l)}\,,
    \label{eq: skip}
\end{align}
Each block comprises three dense layers with dual gating operations and an adaptive residual connection. The trainable $\alpha^{(l)}$ parameters control block nonlinearity: $\alpha^{(l)}=0$ yields an identity mapping, while $\alpha^{(l)}=1$ produces fully nonlinear transformation. 

The final output of a PirateNet of $L$ residual blocks is given by
\begin{align}
    \mathbf{u}_{\mathbf{\theta}} = \mathbf{W}^{(L+1)} \mathbf{x}^{(L)}\,.
\end{align}

Importantly, we initialize $\alpha^{(l)}=0$, making the initial output a linear combination of first-layer embeddings. This initialization strategy mitigates training difficulties in deep networks by starting with effectively shallow architecture and gradually increasing depth through learned $\alpha$ values. Additionally, the linear structure at initialization enables direct integration of prior solution data through least squares fitting:
\begin{align}
\label{eq: lq}
\min_{\mathbf{W}} \left\| \mathbf{W} \Phi  - \mathbf{Y} \right\|_2^2,
\end{align}
where $\mathbf{Y}$ represents available measurements. This approach provides an optimal initial guess based on various data sources, including experimental measurements, boundary conditions, or linearized PDE solutions.

\paragraph{Exact imposition of periodic boundary conditions.} We adopt the approach of \cite{dong2021method} to enforce periodic boundary conditions as hard constraints, improving both training convergence and accuracy. Consider a one-dimensional periodic function with period $P$ satisfying:
\begin{align}\label{eq:periodic_constraint}
    u^{(l)}(a) = u^{(l)}(a + P), \quad l=0, 1, 2, \dots.
\end{align}
We construct a Fourier feature embedding:
\begin{align}
    \label{eq: 1D_Fourier}
    \mathbf{v}(x) = \left(\cos (\omega x), \sin (\omega x) \right),
\end{align}
where $\omega = \frac{2 \pi}{L}$. Any network $u_{\mathbf{\theta}}(v(x))$ using this embedding inherently satisfies the periodic boundary condition.

The same idea can be directly extended to higher-dimensional domains. For two-dimensional domains, the periodicity constraints are:
\begin{align}
    &\frac{\partial^{l}}{\partial x^{l}} u\left(a, y\right)=\frac{\partial^{l}}{\partial x^{l}} u\left(a + P_x, y\right), \quad  y \in\left[b, b + P_y\right], \\
    &\frac{\partial^{l}}{\partial y^{l}} u\left(x, a\right)=\frac{\partial^{l}}{\partial y^{l}} u\left(x, b + P_y\right), \quad  x \in\left[a, a + P_x\right],
\end{align}
for $l=0, 1, 2, \dots$, where  $P_x$  and $P_y$ are the periods in the $x$ and $y$ directions, respectively. Similarly, these constraints are encoded using the embedding:
\begin{align}
    \mathbf{v}(x, y) = \begin{bmatrix}
    \cos \left(\omega_{x} x\right), \sin \left(\omega_{x} x\right), \cos \left(\omega_{y} y\right),  \sin \left(\omega_{y} y\right)
    \end{bmatrix}
\end{align}
with $w_x = \frac{2 \pi}{P_x}, w_y = \frac{2 \pi}{P_y}$.

For time-dependent problems, we concatenate time coordinates $t$ with spatial embeddings: $u_{\mathbf{\theta}}([t, \mathbf{v}(x)])$ or $u_{\mathbf{\theta}}([t, \mathbf{v}(x, y)])$. 



\paragraph{Random weight factorization.} We implement random weight factorization (RWF) \cite{wang2022random} to enhance PINN performance. RWF decomposes each neuron's weight vector as:
\begin{align}
    \mathbf{w}^{(k, l)}=s^{(k, l)} \cdot \mathbf{v}^{(k, l)},
\end{align}
where $k=1,\ldots,d_l$, $l=1,\ldots,L+1$, $\mathbf{w}^{(k, l)} \in \mathbb{R}^{d_{l-1}}$ is the $k$-th row of weight matrix $\mathbf{W}^{(l)}$, $s^{(k, l)} \in \mathbb{R}$ is a trainable scale factor, and $\mathbf{v}^{(k, l)} \in \mathbb{R}^{d_{i-1}}$. This factorization can be expressed in matrix form as:
\begin{align}
    \mathbf{W}^{(l)}=\operatorname{diag}\left(\mathbf{s}^{(l)}\right) \cdot \mathbf{V}^{(l)}, \quad l=1,2, \ldots, L+1
\end{align}
with $\mathbf{s}^{(l)} \in \mathbb{R}^{d_t}$.

Implementation involves: (1) initializing MLP parameters using the Glorot scheme \cite{glorot2010understanding}, (2) initializing scale vectors $\exp(s)$ where $s \sim \mathcal{N}(\mu, \sigma \mathrm{I})$, (3) factorizing each weight matrix as $\mathbf{W}=\operatorname{diag}(\exp (\mathbf{s})) \cdot \mathbf{V}$, and (4) optimizing parameters $\mathbf{s}, \mathbf{V}$ directly. We employ exponential parameterization following Weight Normalization \cite{salimans2016weight} to ensure non-zero scale factors across varied magnitudes. We recommend $\mu=0.5$ or 1 and $\sigma=0.1$, as these values consistently improve convergence and accuracy while avoiding the instability of larger values or the diminished effect of smaller ones.

\subsection{Training pipeline}
\label{appendix:training}

This section details the methodologies and strategies used to train PINN models.

\paragraph{Causal training.}
Recent work by \cite{wang2022respecting} shows that PINNs may violate temporal causality when solving time-dependent PDEs, as they tend to minimize residuals at later times before correctly solving earlier times. To address this, we introduce a causality-aware training approach.
We partition the temporal domain into $M$ equal segments and denote the PDE residual loss within the $i$-th segment as $L_r^i$. The modified residual loss becomes:
\begin{align}
       \mathcal{L}_r(\mathbf{\theta}) = \frac{1}{M} \sum_{i=1}^M w_i \mathcal{L}_r^i(\mathbf{\theta}).
\end{align}
We compute the temporal weights as
\begin{align}
      \label{eq: temporal_update}
    w_i = \exp\left(- \epsilon \sum_{k=1}^{i-1}  \mathcal{L}_r^k( \mathbf{\theta})\right)    , \text{ for } i = 2, 3, \dots, M.
  \end{align}
Then,
\begin{align}
     \mathcal{L}_r(\mathbf{\theta}) = \frac{1}{M} \sum_{i=1}^M     \exp\left(- \epsilon \sum_{k=1}^{i-1}  \mathcal{L}_r^k( \mathbf{\theta})\right) \mathcal L_r^i(\mathbf{\theta}).
\end{align}
The weight $w_i$ decreases exponentially with the cumulative residual loss from previous time steps. This ensures that $\mathcal{L}_r^i(\mathbf{\theta})$ is minimized only after previous residuals $\{\mathcal{L}_r^k(\mathbf{\theta})\}_{k=1}^{i-1}$ become sufficiently small, enforcing temporal causality in the optimization process.

The causality parameter $\epsilon$ requires careful tuning: small values may insufficiently enforce causality, while large values can create optimization difficulties by requiring extremely small early-time residuals before later times are considered. We recommend selecting a moderate $\epsilon$ that allows all temporal weights to converge to 1 by training completion, reducing it if necessary.

\paragraph{Learning rate annealing.} 

Another key challenge in training PINNs is balancing loss components, as they often exhibit multi-scale behaviors, 
 resulting in  unbalanced gradients and training failures.

We implement a self-adaptive learning rate annealing algorithm \cite{wang2021understanding} that automatically balances the weighted loss:
\begin{align}
    \mathcal{L}(\mathbf{\theta}) =  \lambda_{ic} \mathcal{L}_{ic}(\mathbf{\theta}) + \lambda_{bc} \mathcal{L}_{bc}(\mathbf{\theta}) +  \lambda_r \mathcal{L}_r(\mathbf{\theta}), 
\end{align}
The global weights are dynamically computed to equalize the gradient norms of each loss component:
  \begin{align}
  \label{eq: lambda_ic_update}
     \hat{\lambda}_{ic} &=  \frac{  \|\nabla_{\theta}  \mathcal{L}_{ic}(\theta)\| +  \|\nabla_{\theta}  \mathcal{L}_{bc}(\theta)\| +  \|\nabla_{\theta}  \mathcal{L}_{r}(\theta)\|   } {\|\nabla_{\theta}  \mathcal{L}_{ic}(\theta)\|}, \\
     \label{eq: lambda_bc_update}
      \hat{\lambda}_{bc} &= \frac{  \|\nabla_{\theta}  \mathcal{L}_{ic}(\theta)\| +  \|\nabla_{\theta}  \mathcal{L}_{bc}(\theta)\| +  \|\nabla_{\theta}  \mathcal{L}_{r}(\theta)\|   } {\|\nabla_{\theta}  \mathcal{L}_{bc}(\theta)\|},  \\
      \label{eq: lambda_r_update}
       \hat{\lambda}_{r} &=  \frac{  \|\nabla_{\theta}  \mathcal{L}_{ic}(\theta)\| +  \|\nabla_{\theta}  \mathcal{L}_{bc}(\theta)\| +  \|\nabla_{\theta}  \mathcal{L}_{r}(\theta)\|   } {\|\nabla_{\theta}  \mathcal{L}_{r}(\theta)\|}, 
  \end{align}
where $\|\cdot\|$ denotes the $L^2$ norm.
Then we obtain
\begin{align}
  \| \hat{\lambda}_{ic} \nabla_\theta \mathcal{L}_{ic} (\theta) \| =   \| \hat{\lambda}_{bc} \nabla_\theta \mathcal{L}_{ic} (\theta) \| = \| \hat{\lambda}_{r} \nabla_\theta \mathcal{L}_{ic} (\theta) \| = \| \nabla_\theta \mathcal{L}_{ic} (\theta) \| +  \| \nabla_\theta \mathcal{L}_{bc} (\theta) \|  +  \| \nabla_\theta \mathcal{L}_{r} (\theta) \|. 
\end{align}
This formulation equalizes the gradient norms of weighted losses, preventing bias toward any particular term during training. The weights are updated as running averages of their previous values, stabilizing stochastic gradient descent. These updates occur at user-specified intervals (typically every 100-1000 iterations), incurring minimal computational overhead.

\paragraph{Curriculum training and time-marching.} Despite the improvements described above, PINNs still face challenges in complex domains requiring high accuracy, such as chaotic systems like high Reynolds number Navier-Stokes equations where error accumulation can cause trajectory divergence. We address these challenges using curriculum training \cite{krishnapriyan2021characterizing}, which decomposes the optimization into more manageable sub-tasks.

An effective approach we employ is the curriculum training strategy introduced by \cite{krishnapriyan2021characterizing}. The core idea involves decomposing the entire optimization task for PINNs into a sequence of more manageable sub-tasks. In this work, we mainly focus on  integrating this strategy into our training pipeline for solving time-dependent PDEs and singular perturbation problems. 

For time-dependent PDEs, we implement temporal domain decomposition: the time domain is divided into smaller intervals. After the first window, initial conditions for subsequent windows are set using predictions from the final step of the  previous window.  This approach reduces the difficulty of the optimization task of learning full system dynamics, though at an increased computational cost due to per-window model retraining.

While we also partition the temporal domain to compute causal weights within each window, this differs from the time-marching strategy. Both techniques promote learning solutions sequentially along the time axis to respect causality, but causal weighting complements rather than replaces time-marching, as causality violations may still occur within individual time windows.

\subsection{Data Generation}
We generate our reference dataset using two numerical packages: Chebfun \cite{driscoll2014chebfun} in MATLAB and IncompressibleNavierStokes \cite{Agdestein_IncompressibleNavierStokes_jl_2024} in Julia. The data generation process employs a time step of $dt=10^{-4}$, followed by temporal downsampling to construct the final dataset. Table \ref{tab: data_gen} summarizes the PDE parameters and dataset details.

\begin{table}
\renewcommand{\arraystretch}{1.2}
\centering
\caption{Parameter settings and numerical configurations for generating the reference solution across PDE benchmarks.}
\label{tab: data_gen}
\vspace{1mm}
\resizebox{\textwidth}{!}{
\begin{tabular}{l|l c c}
\toprule
\textbf{PDE} & Parameter & Package & Resolution  \\
\midrule
\textbf{Wave} & $c=4$   & N/A & $200 \times 128$\\
\textbf{Burgers} & $\nu = 0.01 \ \pi$ & Chebfun & $200 \times 512$\\
\textbf{AC} & $\epsilon = 10^{-4}, a=5$ & Chebfun & $200 \times 512$ \\
\textbf{KdV} & $\eta=1, \mu=0.022$ & Chebfun & $200 \times 512$\\
\textbf{KS} & $\alpha = 100/ 16, \beta=100 / 16^2, \gamma=100 / 16^4$ & Chebfun & $250 \times 512$\\
\textbf{GS} & $\epsilon_1=0.2, \epsilon_2=0.1, b_1=40, b2=100, c_1=c_2=1,000$ & Chebfun & $100 \times 200 \times 200$\\
\textbf{GL} & $\epsilon=0.004, \mu=10, \gamma=10 + 15i$ & Chebfun & $100 \times 200 \times 200$\\
\textbf{LDC} & $\text{Re}{=}5{\times}10^3$ & IncompressibleNavierStokes  & $128 \times 128$\\
\textbf{KF} & $\text{Re}{=}10^4$ & IncompressibleNavierStokes & $50 \times 512 \times 512$\\
\textbf{RT} & $\text{Ra}{=}10^6, \text{Pr}=0.71$ & IncompressibleNavierStokes & $40 \times 100 \times 200$\\
\bottomrule
\end{tabular}
}
\end{table}

\subsection{Hyper-parameters}

Unless otherwise specified, we adopt the following hyperparameter configuration for our experiments.

\paragraph{Architecture.} We employ PirateNet \cite{wang2024piratenets} as our backbone architecture, configured with three residual blocks (9 layers in total), a hidden dimension of 256, and \texttt{Tanh} activation functions. All weight matrices are initialized using random weight factorization (RWF) \cite{wang2022random} with parameters $\mu=1.0$ and $\sigma=0.1$. When applicable, we strictly enforce exact periodic boundary conditions following \cite{dong2021method}.

\paragraph{Training Protocol.} We train our models using mini-batch gradient descent with 8,192 randomly sampled collocation points per iteration. The learning rate schedule comprises an initial linear warm-up phase over the first 5,000 steps (from 0 to $10^{-3}$), followed by exponential decay with a factor of 0.9. To enhance training stability and convergence, we implement learning rate annealing for loss balancing \citep{wang2021understanding,wang2023expert}, updating the loss weights every 1,000 iterations with a moving average. For time-dependent PDEs, we apply causal training \citep{wang2022respecting,wang2023expert} with a causal tolerance of 1.0. Additionally, we leverage time-marching and curriculum learning strategies \citep{krishnapriyan2021characterizing} for particularly challenging benchmarks.

\paragraph{Optimizers.} We evaluate several optimizers with the following configurations:
\begin{itemize}
    \item \textbf{Adam:} We use the Adam optimizer \cite{kingma2014adam} with standard hyperparameters $\beta_1 = 0.9$ and $\beta_2 = 0.999$, which has become the de facto standard for training PINNs due to its robust performance and computational efficiency.
    
    \item \textbf{SOAP \cite{vyas2024soap}:} Based on our ablation studies presented in Figure~\ref{fig:ablation}, we select $\beta_1 = 0.99$ and $\beta_2 = 0.999$ for SOAP, which yield optimal performance across our experimental tasks.
    
    \item \textbf{Muon \cite{jordan2024muon}:} For Muon, we adopt momentum hyperparameters matching those of Adam: $\beta_1 = 0.9$ and $\beta_2 = 0.999$. Notably, we employ more accurate Newton-Schulz coefficients $(2, -1.5, 0.5)$ and implement 10 Newton-Schulz matrix iterations to ensure convergence of the orthogonalization procedure.
    
    \item \textbf{Kron \cite{li2017preconditioned}:} For Kronecker-factored optimization, we use the same momentum hyperparameters as Adam: $\beta_1 = 0.9$ and $\beta_2 = 0.999$.
\end{itemize}

The complete set of hyperparameters is detailed in Table \ref{tab: hyper-parameters}, largely following the configurations established in \cite{wang2023expert} and \cite{wang2024piratenets}. The decay step is tailored for each benchmark to ensure the learning rate reaches a sufficiently small value (approximately $10^{-7}$) by the end of training. The number of time windows is determined empirically based on problem complexity, with fine-tuning guided by the loss convergence behavior observed during preliminary experiments.

\begin{table}
\renewcommand{\arraystretch}{1.2}
\centering
\caption{{\em Hyperparameter configurations for benchmark PDEs.} Hyperparameter settings used to reproduce our experimental results. The backbone architecture is PirateNet, where Depth indicates the number of adaptive residual blocks, and Width denotes the number of neurons per hidden layer. RFF and RWF represent Random Fourier Features and Random Weight Factorization, respectively.}
\label{tab: hyper-parameters}
\vspace{1mm}
\resizebox{\textwidth}{!}{
\begin{tabular}{l cccccccccc}
\toprule
\textbf{Parameter} & \textbf{Wave} & \textbf{Burgers} & \textbf{AC} & \textbf{KdV} & \textbf{KS} & \textbf{GS} & \textbf{GL} & \textbf{LDC} & \textbf{KF} & \textbf{RT} \\
\midrule
\textbf{Architecture} & & & & & & & & & & \\
\midrule
Depth & 3 & 3 & 3 & 3 & 3 & 3 & 3 & 4 & 3 & 3 \\
Width & 256 & 256 & 256 & 256 & 256 & 256 & 256 & 256 & 384 & 384 \\
Activation & Tanh & Tanh & Tanh & Tanh & Tanh & Swish & Swish & Tanh & Tanh & Tanh \\
RFF scale & 10.0 & 2.0 & 2.0 & 2.0 & 2.0 & 2.0 & 2.0 & 10.0 & 2.0 & 2.0 \\
RWF & \multicolumn{10}{c}{$\mu{=}1.0, \sigma{=}0.1$} \\
\midrule
\textbf{Learning rate schedule} & & & & & & & & & & \\
\midrule
Initial learning rate &  \multicolumn{10}{c}{ $10^{-3}$ } \\
Decay rate &  \multicolumn{10}{c}{ 0.9 } \\
Decay steps & $2 \times 10^3$ & $2 \times 10^3$ & $5 \times 10^3$ & $2 \times 10^3$ & $2 \times 10^3$ & $2 \times 10^3$ & $2 \times 10^3$ & $2 \times 10^3$ & $2 \times 10^3$ & $2 \times 10^3$ \\
Warmup steps &  \multicolumn{10}{c}{ $5 \times 10^{3}$ } \\
\midrule
\textbf{Training} & & & & & & & & & & \\
\midrule
Iters (per time window) & $10^5$ &  $10^5$ & $3{\times}10^5$ & $10^5$ & $10^5$ & $10^5$ & $10^5$ & $2{\times}10^5$ & $2{\times}10^4$ & $10^5$ \\
Batch size &  \multicolumn{10}{c}{ 8,192 } \\
\# Time windows & 1& 1 & 1& 1& 10& 10 & 5 & N / A & 25 & 4 \\
\midrule
\textbf{Weighting Scheme} &  \multicolumn{10}{c}{ Grad Norm } \\
\midrule
\textbf{Causal weighting} & & & & & & & & & & \\
\midrule
Tolerance & 1.0 & 1.0 & 1.0 & 1.0 & 1.0 & 1.0 & 1.0 & N / A & 1.0 & 1.0 \\
\# Chunks & 16 & 16 & 16  & 16  & 16  & 16  &16  & N / A  & 16  & 16  \\
\bottomrule
\end{tabular}
}
\end{table}

\subsection{Computational Cost}
Our implementation is based on JAX-PI \cite{wang2023expert} and we conducted all experiments on a single NVIDIA A6000 GPU, with detailed runtime benchmarks reported in Table \ref{tab: cost}. 

\begin{table}
\centering
\renewcommand{\arraystretch}{1.4}
\caption{Computational runtime (in hours) comparison of different methods across various PDEs. All experiments are performed on an Nvidia A6000 GPU, reporting the total training time needed to achieve convergence using PINNs with Adam and SOAP, respectively}
\label{tab: cost}
\begin{tabular}{l cc}
\toprule
\textbf{Benchmark} & \multicolumn{1}{c}{\textbf{Adam}  } & \multicolumn{1}{c}{\textbf{SOAP}  } \\
\midrule
Wave  & 2.80 &  4.35\\
Burgers  & 1.18  & 4.05 \\
Allen-Cahn  & 1.48 & 5.83 \\
Korteweg–De Vries  & 1.61 & 3.90 \\
Kuramoto-Sivashinsky & 19.51 & 34.16\\
Grey-Scott & 19.52 & 40.01\\
Ginzburg-Landau  & 15.98 & 23.75 \\
Lid-driven cavity ($\text{Re}=5 \times 10^3$) & 5.67 & 8.25 \\
Kolmogorov flow ($\text{Re}=10^4$) & 9.56 & 11.00 \\
Rayleigh-Taylor instability ($\text{Pr}=0.71, \text{Ra}=10^6$) & 20.23 & 21.73 \\
\bottomrule
\end{tabular}
\vspace{1mm}
\end{table}

\subsection{Benchmarks}
\label{appendix:benchmarks}

\paragraph{Wave equation.}  We consider a one-dimensional wave equation in the domain $\Omega=[0,1] \times[0,1]$ taking the form
\begin{align*}
    & u_{t t}(x, t)-4 u_{x x}(x, t)=0, \quad(x, t) \in(0,1) \times(0,1), \\
& u(0, t)=u(1, t)=0, \quad t \in[0,1], \\
& u(x, 0)=\sin (\pi x)+\frac{1}{2} \sin (4 \pi x), \quad x \in[0,1], \\
& u_t(x, 0)=0, \quad x \in[0,1].
\end{align*}
where  $u$ represents the wave amplitude, and $c$ is the wave propagation speed, determined by the medium's physical properties.

By d'Alembert's formula, the solution $u(x, t)$ is given by
\begin{align*}
    u(x, t)=\sin (\pi x) \cos (2 \pi t)+\frac{1}{2} \sin (4 \pi x) \cos (8 \pi t).
\end{align*}

\begin{figure}
    \centering
    \includegraphics[width=1.0\linewidth]{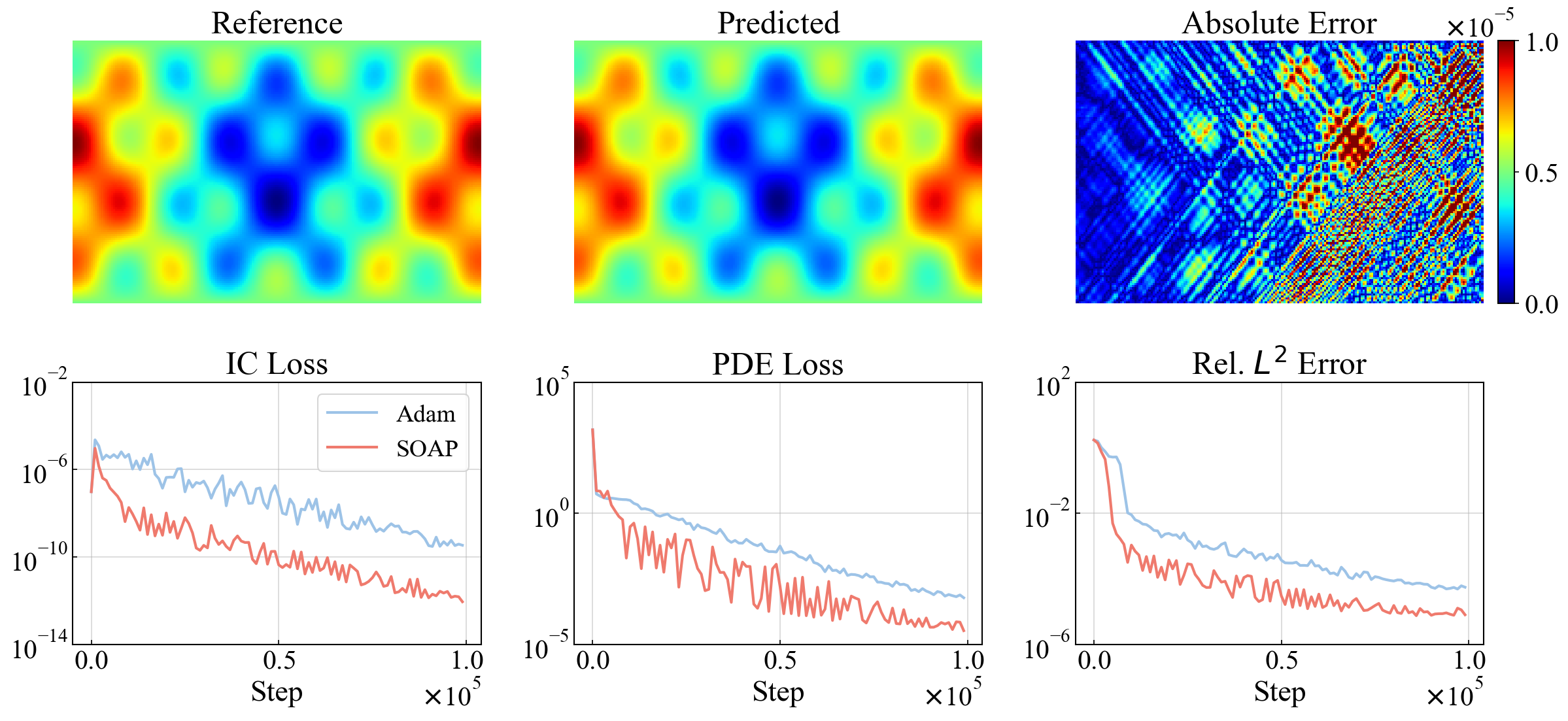}
\caption{{\em Wave equation.}  Top: Comparison between the reference solution and the model predictions. Bottom: Training loss and test error trajectories for the Adam and SOAP optimizers.}
    \label{fig:wave}
\end{figure}

\paragraph{Burgers equation.}  The 1D Burgers equation is defined as:
\begin{align*}
u_t + u u_x = \nu u_{xx},
\end{align*}
where $u$ represents the velocity field, and $\nu$ is the kinematic viscosity coefficient controlling the diffusion strength. Here we set  $(x, t) \in \Omega = [-1, 1] \times [0, 1]$, with initial and boundary conditions:
\begin{align*}
u(x, 0) &= -\sin(\pi x), \\
u(-1, t) &= u(1, t) = 0,
\end{align*}
and viscosity parameter $\nu = 0.01/\pi$.

\begin{figure}
    \centering
    \includegraphics[width=1.0\linewidth]{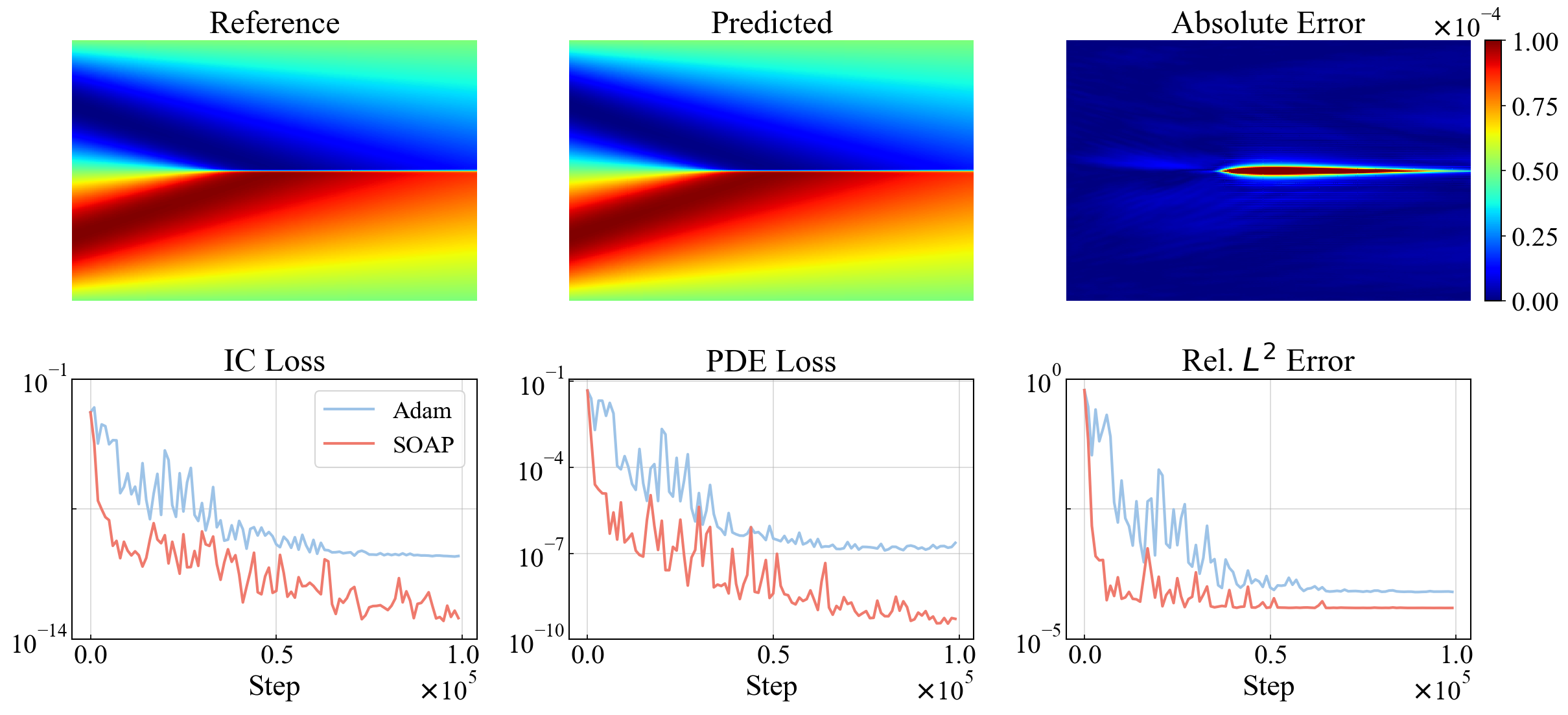}
\caption{{\em Burgers' equation.}  Top: Comparison between the reference solution and  model predictions. Bottom: Training loss and test error trajectories for the Adam and SOAP optimizers.}
    \label{fig:burgers}
\end{figure}

\paragraph{Allen-Cahn equation.} We investigate the one-dimensional Allen-Cahn equation with periodic boundary conditions:
\begin{align*}
    &u_{t}-0.0001 u_{x x}+5 u^{3}-5 u=0\,, \quad t \in[0,1]\,,\  x \in[-1,1]\,, \\
    &u(0, x)=x^{2} \cos (\pi x)\,, \\
    &u(t, -1)=u(t, 1)\,, \quad u_{x}(t, -1)=u_{x}(t, 1)\,.
\end{align*}
where $u$ represents the order parameter (e.g., concentration difference between two phases),  $\epsilon$ controls the interfacial width, $a$ is the reaction rate coefficient, and the term $({u}-{u}^3)$ drives the phase separation.

It is worth noting that this benchmark has been extensively used to validate the effectiveness of PINNs methodologies. In Table \ref{tab: AC}, we compare the test errors across different PINNs advancements, demonstrating that our approach achieves state-of-the-art performance with an improvement of up to one order of magnitude in accuracy.

\begin{figure}
    \centering
    \includegraphics[width=1.0\linewidth]{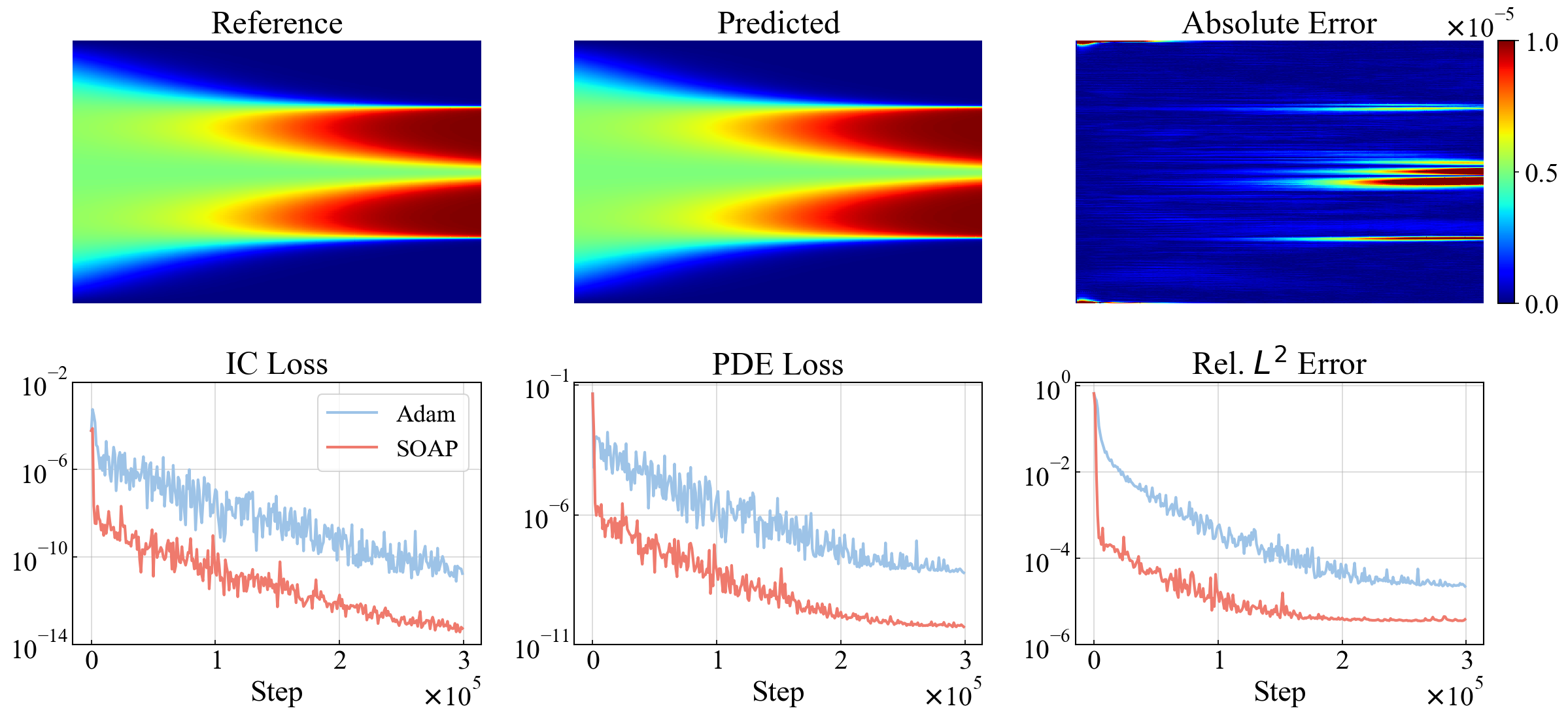}
\caption{{\em Allen-Cahn equation.}  Top: Comparison between the reference solution and model predictions. Bottom: Training loss and test error trajectories for the Adam and SOAP optimizers.}
    \label{fig:ac}
\end{figure}

\begin{table}
    
    \renewcommand{\arraystretch}{1.4}
    \centering
    \caption{{\em Allen-Cahn equation:} Relative $L^2$ test errors obtained by different PINNs variants.}
    \label{tab: AC}
    \begin{tabular}{l|c}
    \hline
    \textbf{Method}   & \textbf{Relative $L^2$ error}  \\
     \hline
      Original formulation of Raissi {\it et al.} \cite{raissi2019physics}    &  $4.98 \times 10^{-1}$ \\
      Adaptive time sampling \cite{wight2020solving} & $2.33 \times 10^{-2}$ \\
       Self-attention \cite{mcclenny2020self} & $2.10 \times 10^{-2}$  \\
       Time marching \cite{mattey2022novel}  & $1.68 \times 10^{-2}$ \\
       Causal training \cite{wang2022respecting} & $1.39 \times 10^{-4}$ \\
       Dirac delta function causal training \cite{es2023optimal}  & $6.29  \times 10^{-5}$ \\
       JAX-PI \cite{wang2023expert} & $5.37 \times 10^{-5}$ \\
        RBA-PINNs  \cite{anagnostopoulos2023residual}  & $4.55 \times 10^{-5}$ \\
    PirateNet \cite{wang2024piratenets} & ${2.24 \times 10^{-5}}$ \\
         BRDR-PINNs  \cite{chen2024self}  & $1.45 \times 10^{-5}$ \\
        \textbf{Ours} & $\mathbf{3.48 \times 10^{-6}}$ \\
    \hline
    \end{tabular}
\end{table}

\paragraph{Korteweg–De Vries equation.} The one-dimensional KdV equation is expressed as follows:
\begin{align*}
& u_t + \eta u u_x + \mu^2 u_{x x x} = 0\,, \quad t \in(0,1), \quad x \in(-1,1)\,, \\
& u(x, 0) = \cos (\pi x)\,, \\
& u(t,-1) = u(t, 1)\,,
\end{align*}
where $u$ represents the wave amplitude or water surface elevation, and $\eta$ governs the strength of the nonlinearity, while $\mu$ controls the dispersion level. Under the KdV dynamics,  this initial wave evolves into a series of solitary-type waves.

For our study, we adopt the classical parameters of the KdV equation, setting $\eta = 1$ and $\mu = 0.022$ \cite{zabusky1965interaction}.

\begin{figure}
    \centering
    \includegraphics[width=1.0\linewidth]{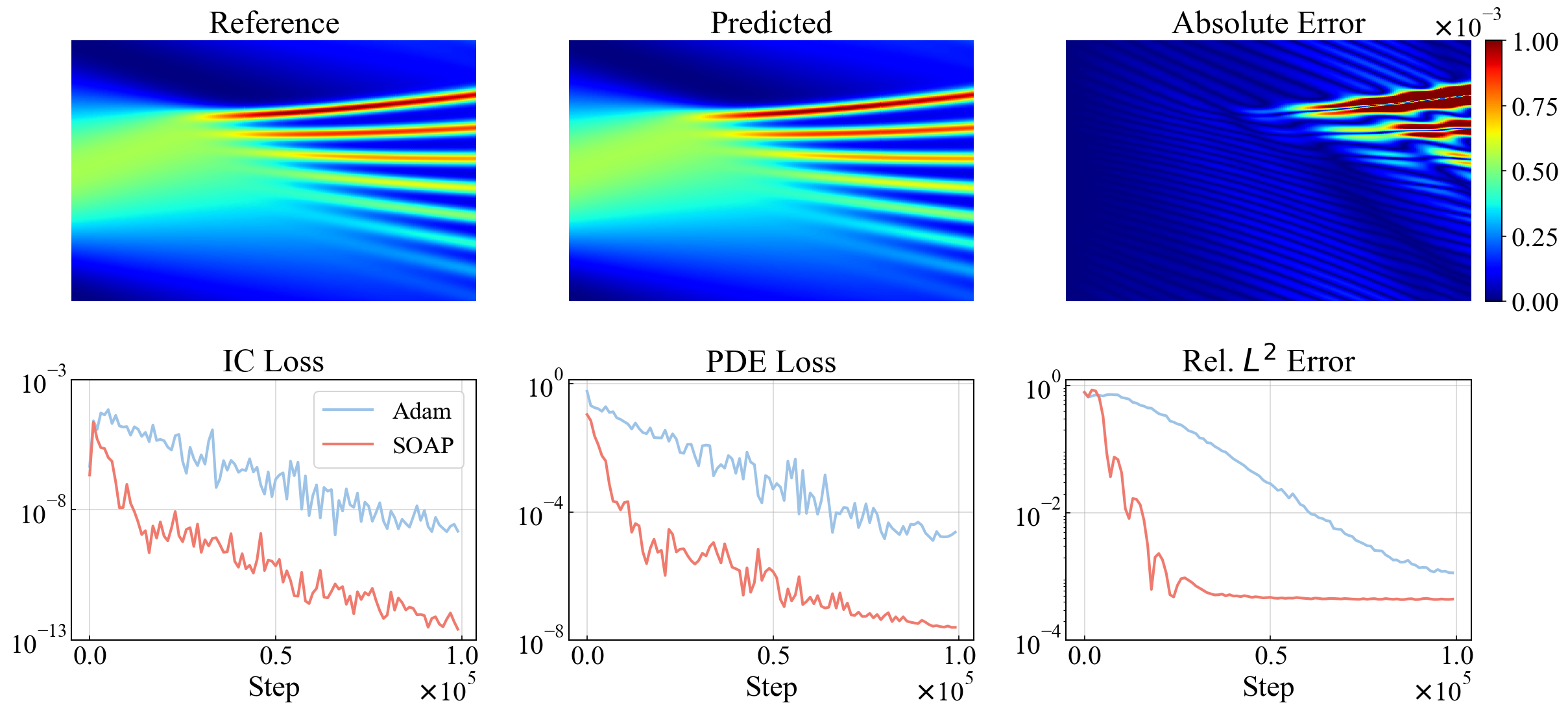}
\caption{{\em Korteweg–De Vries equation.}  Top: Comparison between the reference solution and model predictions. Bottom: Training loss and test error trajectories for the Adam and SOAP optimizers.}
    \label{fig:kdv}
\end{figure}

\paragraph{Kuramoto-Sivashinsky equation.} The one-dimensional equation takes the form:
\begin{align*}
    &u_t+\alpha u u_x+\beta u_{x x}+\gamma u_{x x x x}=0, \quad t \in[0,T], x \in[0,2 \pi], \\
    & u(0, x)=u_0(x),
\end{align*}
where $u$ represents the height of a thin film or flame front. This equation arises in various physical contexts, including flame front propagation, thin film flows, and plasma instabilities.

In this example, we take  $T=0.8$, $\alpha=100 / 16, \beta=100 / 16^2, \gamma=100 / 16^4$ and $u_0(x)=\cos (x)(1+\sin (x))$.

\begin{figure}
    \centering
    \includegraphics[width=0.7\linewidth]{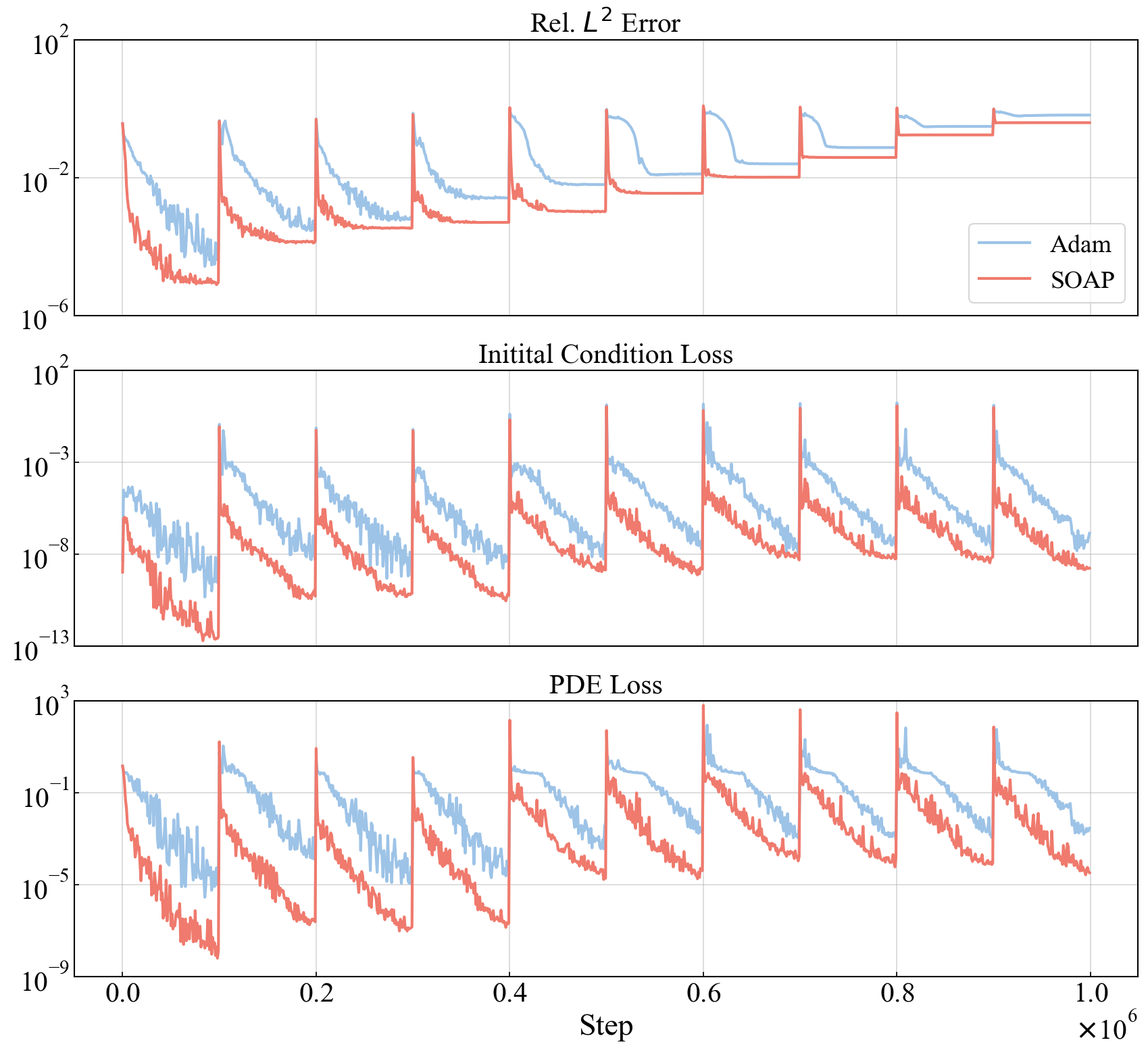}
\caption{{\em Kuramoto-Sivashinsky equation.} Training loss and test error trajectories for the Adam and SOAP optimizers.}
    \label{fig:ks_loss}
\end{figure}

\paragraph{Grey-Scott equation.} The system is described by the following coupled PDEs:
\begin{align*}
    u_t &=\epsilon_1 \Delta u + b_1(1-u) - c_1 u v^2,  \quad t \in (0, 2)\,, \ (x, y) \in (-1, 1)^2\,, \\
    v_t &=\epsilon_2 \Delta v - b_2 v + c_2 u v^2\,, \quad t \in (0, 2)\,, \ (x, y) \in (-1, 1)^2\,,
\end{align*}
With periodic boundary conditions, the initial conditions are:
\begin{align*}
    &u_0(x, y) = 1 - \exp(-10 ((x + 0.05)^2 + (y + 0.02)^2))\,, \\
    &v_0(x, y) = 1 - \exp(-10 ((x - 0.05)^2 + (y - 0.02)^2))\,.
\end{align*}
where $u$ and $v$ represent activator and inhibitor concentrations respectively, $\varepsilon_1$ and $\varepsilon_2$ are diffusion coefficients, and $(b_1, b_2, c_1, c_2)$ control reaction kinetics. This system generates diverse spatial patterns including spots and stripes.

We set parameters $\epsilon_1=0.2$, $\epsilon_2=0.1$, $b_1=40$, $b_2=100$, and $c_1=c_2=1,000$, which generates characteristic pattern formations. Due to the similar behavior of $u$ and $v$, we report only the relative $L^2$ error of $u$ in Table \ref{tab: sota}.

\begin{figure}
    \centering
    \includegraphics[width=0.7\linewidth]{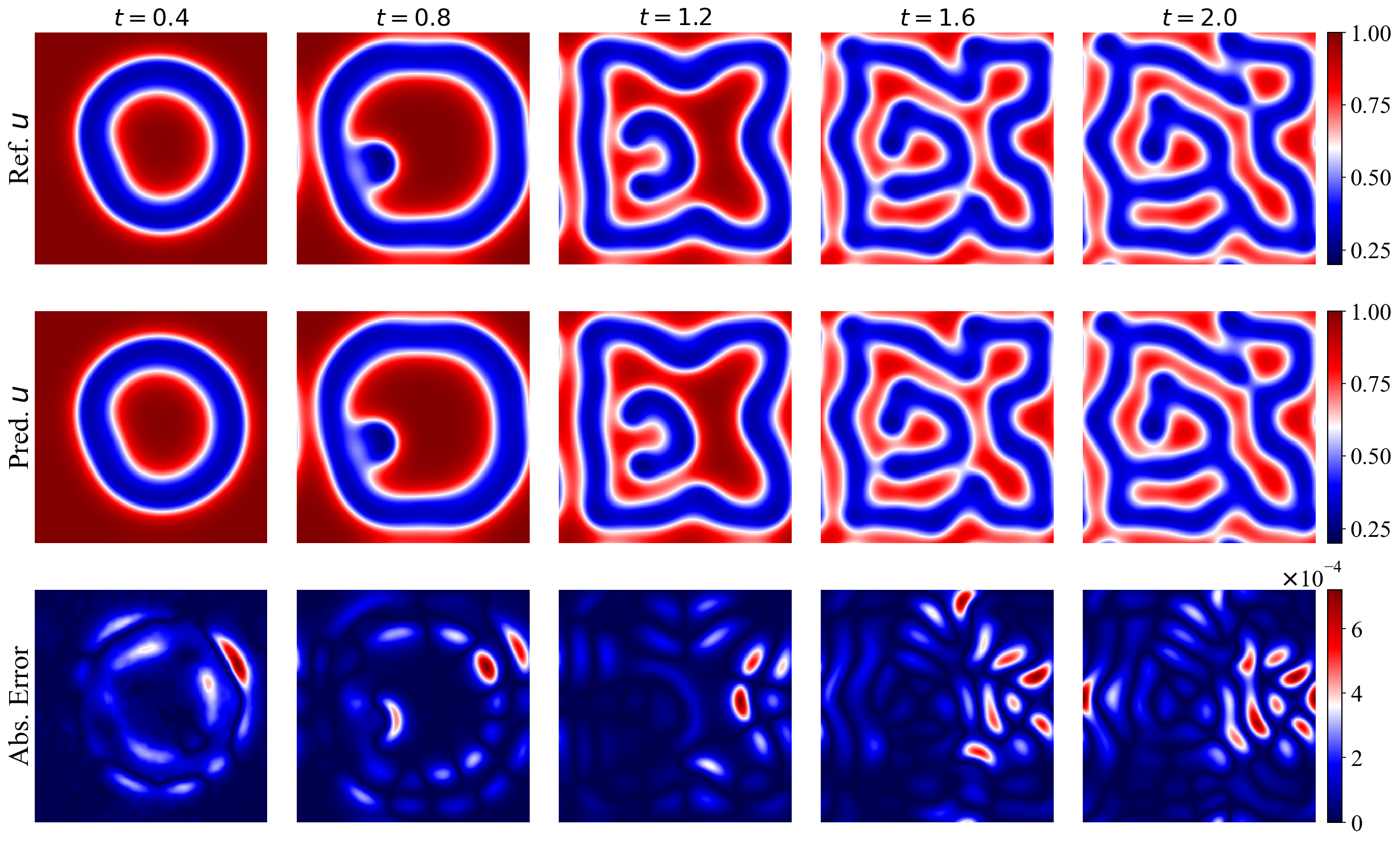}
\caption{{\em Grey-Scott equation.} Comparison between reference solution and model predictions.}
    \label{fig:gs_pred_u}
\end{figure}


\begin{figure}
    \centering
    \includegraphics[width=0.7\linewidth]{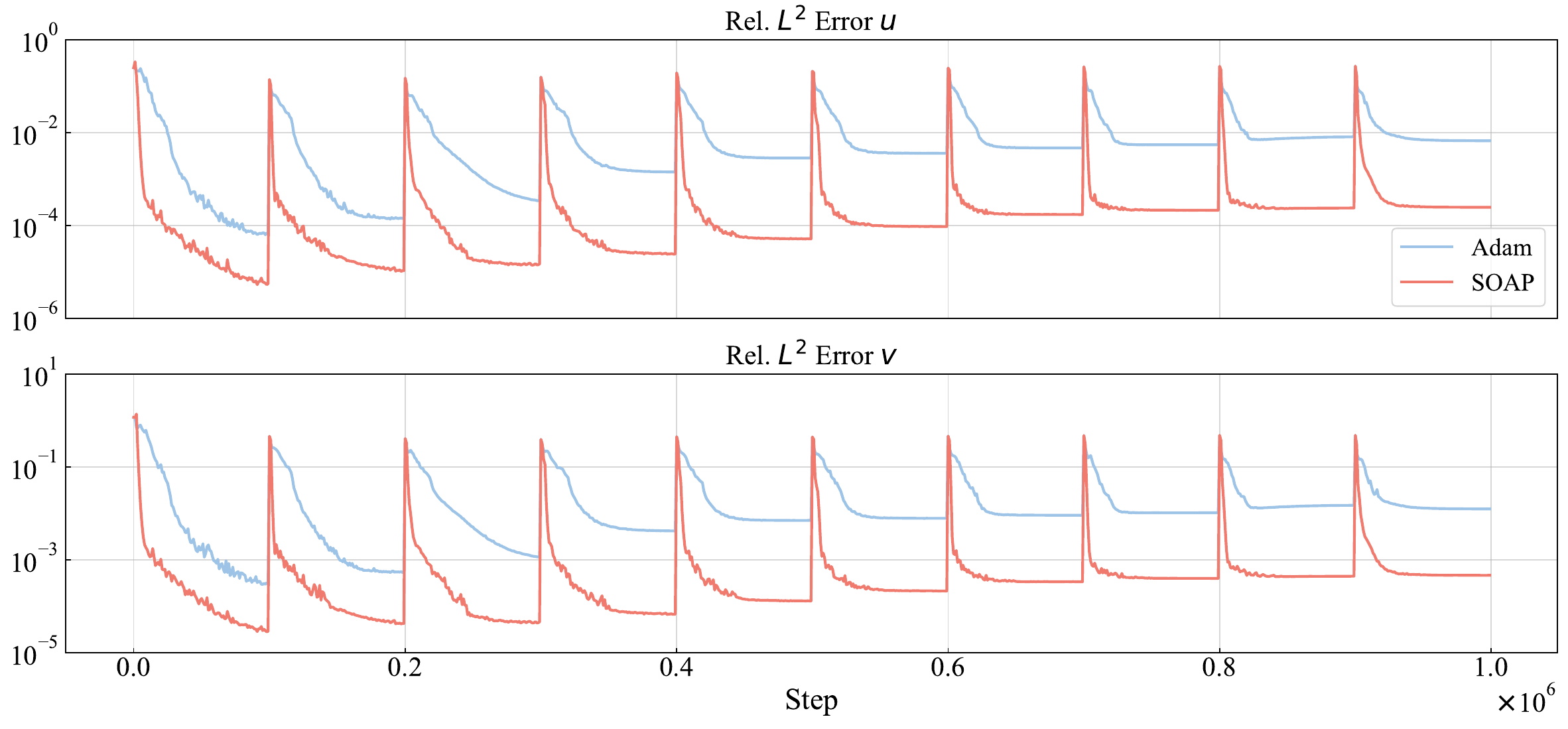}
\caption{{\em Grey-Scott equation.} Test error trajectories for the Adam and SOAP optimizers.}
    \label{fig:gs_error}
\end{figure}

\begin{figure}
    \centering
    \includegraphics[width=0.7\linewidth]{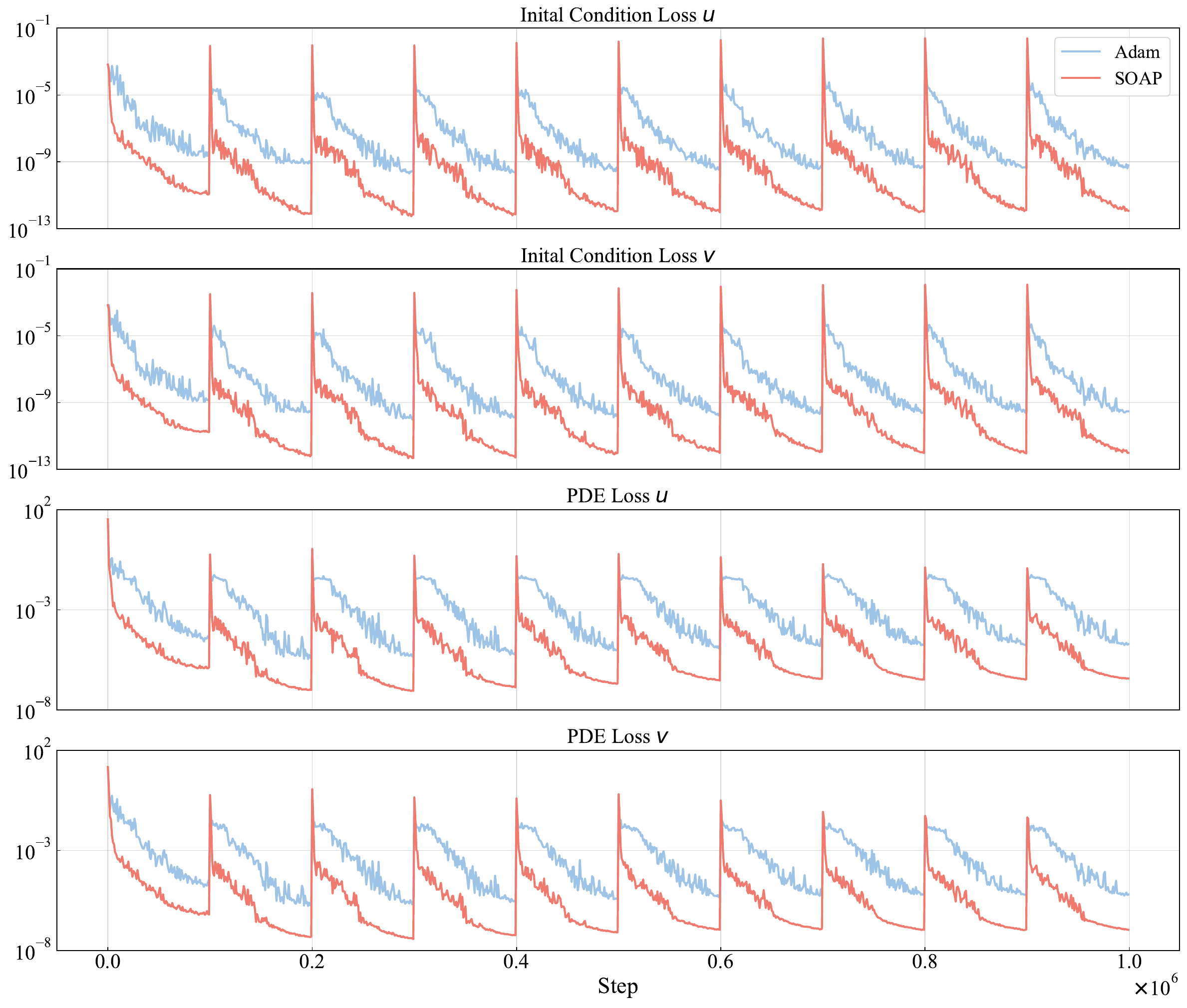}
\caption{{\em Grey-Scott equation.} Training loss and test error trajectories for the Adam and SOAP optimizers.}
    \label{fig:gs_loss}
\end{figure}

\paragraph{Ginzburg-Landau equation.} The complex Ginzburg-Landau equation in 2D takes the form
\begin{align*}
    \frac{\partial A}{\partial t}= \epsilon \Delta A + \mu A - \gamma  A|A|^2\,, \quad t \in (0, 1)\,,
    \ (x, y) \in (-1, 1)^2\,,
\end{align*}
with periodic boundary conditions, an initial condition
\begin{align*}
    A_0(x, y) = (10y + 10 i  x) \exp\left(-0.01 (2500 x^2 + 2500 y^2)\right)\,,
\end{align*}
where  $A$ is the complex amplitude representing the envelope of oscillations, $\epsilon$ represents the diffusion coefficient, $\mu$ is the linear growth rate, and $\gamma$ controls the nonlinear saturation. For this example, we set $\epsilon=0.004$, $\mu = 10$ and $\gamma= 10 + 15i$.

By denoting $A = u + i v$, we can decompose the equation into real and imaginary components, resulting in the following system of PDEs,
\begin{align*}
     \frac{\partial u}{\partial t} &= \epsilon \Delta u + \mu (u - (u-1.5 v) (u^2 + v^2))\,, \\
      \frac{\partial v}{\partial t} &= \epsilon \Delta v + \mu (v - (v + 1.5 u) (u^2 + v^2))\,.
\end{align*}

Given the coupled dynamics of $u$ and $v$, we present the relative $L^2$ error of $u$ in Table \ref{tab: sota}.

\begin{figure}
    \centering
    \includegraphics[width=0.7\linewidth]{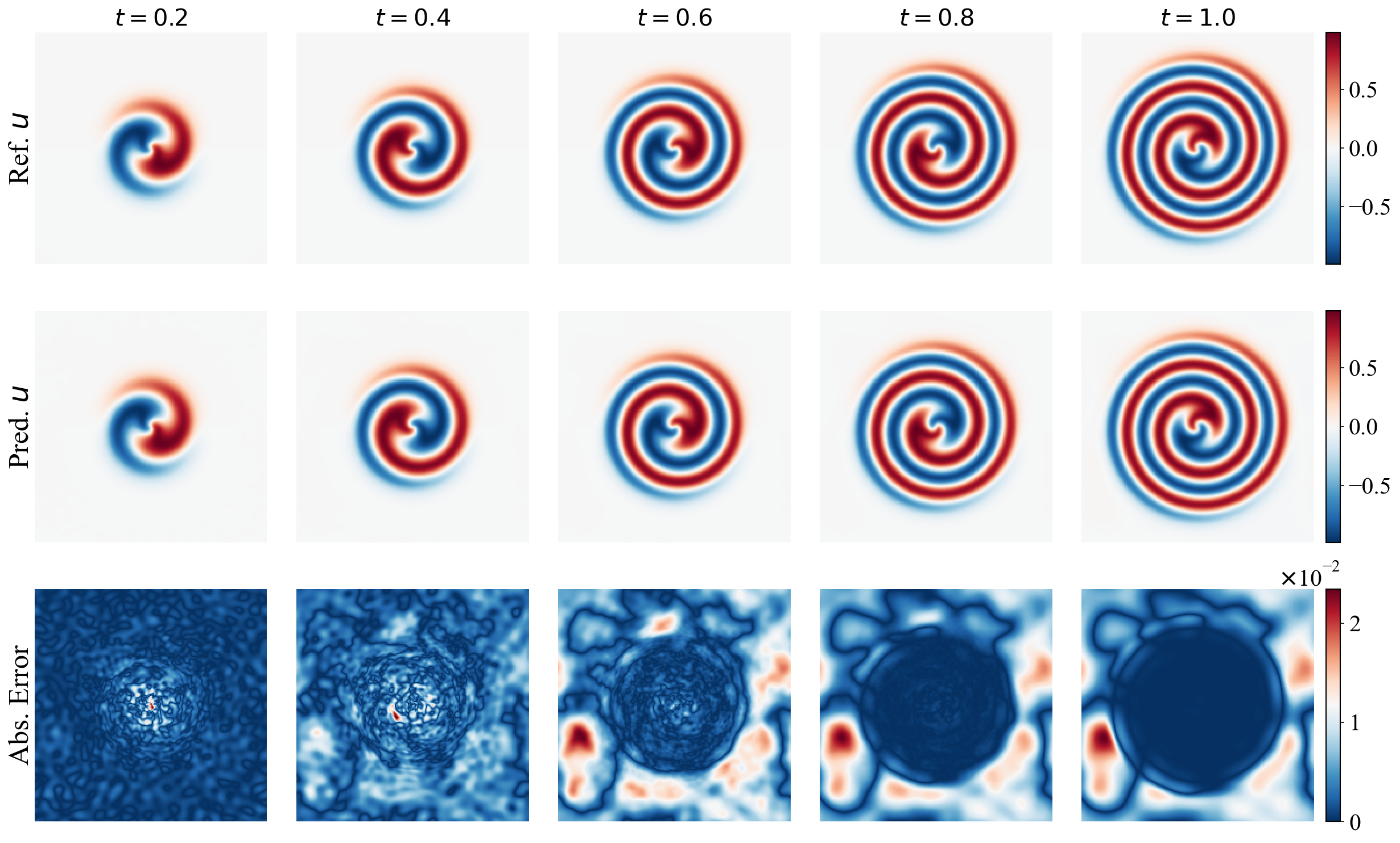}
\caption{{\em Ginzburg-Landau equation.} Comparison between the reference solution and model predictions.}
    \label{fig:gl_pred_u}
\end{figure}

\begin{figure}
    \centering
    \includegraphics[width=0.7\linewidth]{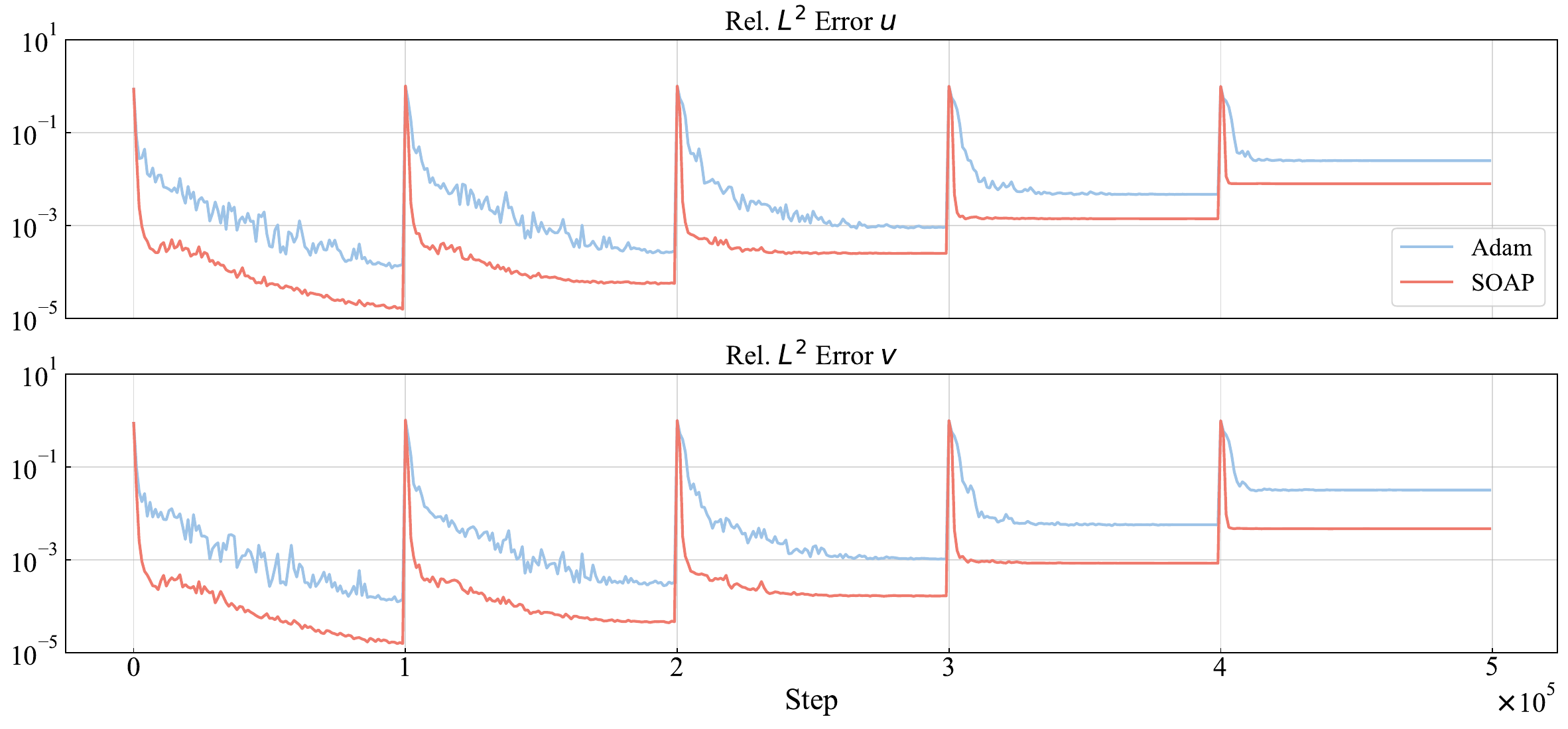}
\caption{{\em Ginzburg-Landau equation.} Test error trajectories for the Adam and SOAP optimizers.}
    \label{fig:gl_error}
\end{figure}

\begin{figure}
    \centering
    \includegraphics[width=0.7\linewidth]{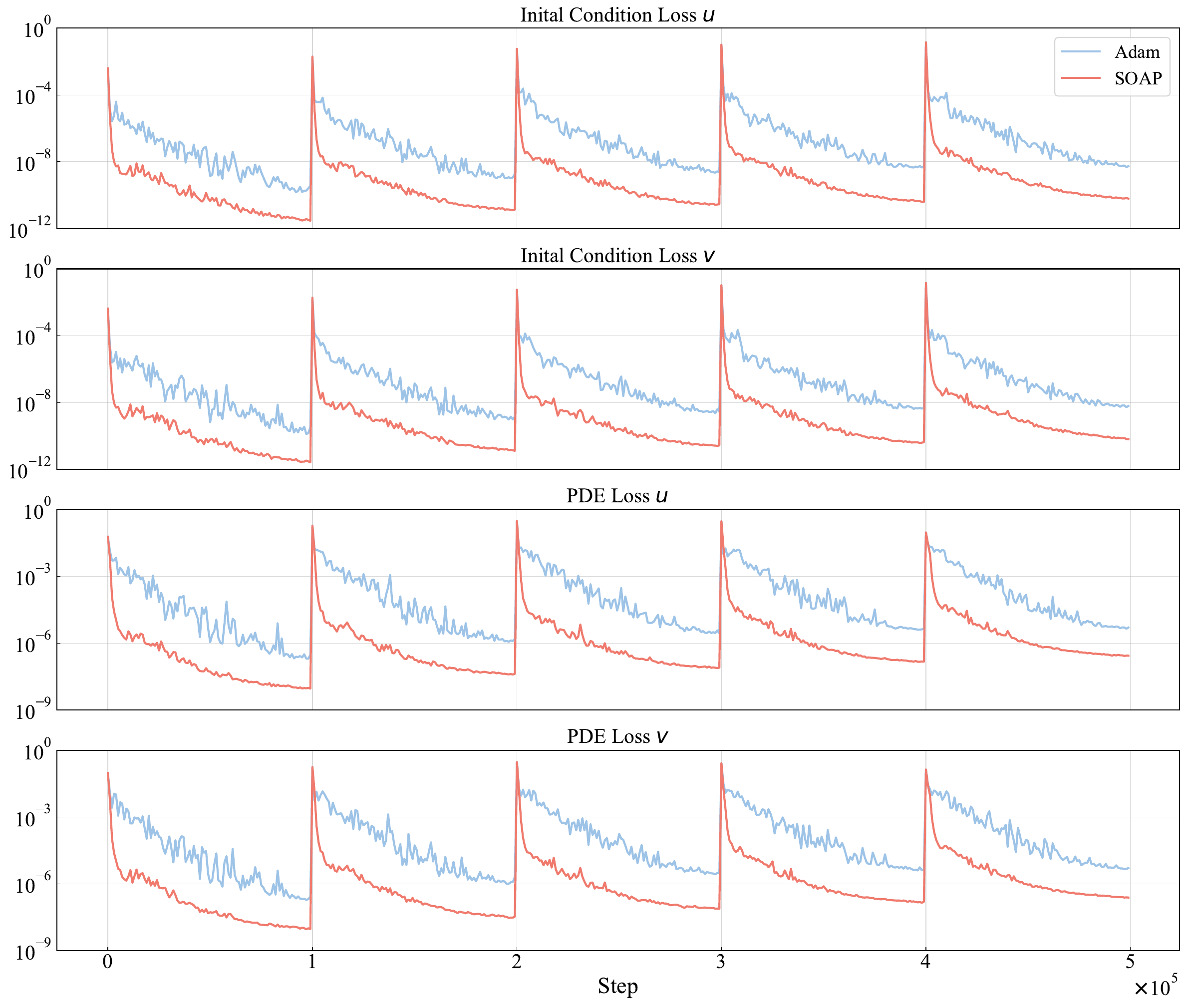}
 \caption{{\em Ginzburg-Landau equation.} Training loss trajectories for the  Adam and SOAP optimizers.}
    \label{fig:gl_loss}
\end{figure}

\paragraph{Lid-driven Cavity.} We study the incompressible Navier-Stokes equations in non-dimensional form for a two-dimensional domain:
\begin{align*}
    \mathbf{u} \cdot \nabla \mathbf{u}+\nabla p-\frac{1}{R e} \Delta \mathbf{u}&=0\,, \quad  (x,y) \in (0,1)^2\,, \\
    \nabla \cdot \mathbf{u}&=0\,, \quad  (x,y) \in (0,1)^2\,,
\end{align*}
where $\mathbf{u} = (u,v)$ represents the steady-state velocity field, $p$ is the pressure field, and $Re$ is the Reynolds number which characterizes the ratio of inertial to viscous forces.  This system models the equilibrium state of the flow, which is driven by the top boundary moving at a constant velocity while the other walls are stationary, leading to the formation of characteristic vortical structures whose complexity increases with the Reynolds number.

To ensure continuity at the corner boundaries, we implement a smoothed top-lid boundary condition:
\begin{align}
& u(x, y)=1-\frac{\cosh \left(C_0(x-0.5)\right)}{\cosh \left(0.5 C_0\right)}\,, \quad v(x, y)=0\,,
\end{align}
where $x \in [0, 1], y=1, C_0 = 50$. For the other three walls, we enforce a no-slip boundary condition. Our goal is to obtain the velocity and pressure field corresponding to a Reynolds number of $5,000$.  The accuracy of our method is evaluated using the velocity magnitude $\sqrt{u^2 + v^2}$, with results presented in Table \ref{tab: sota}.

\begin{figure}
    \centering
    \includegraphics[width=0.8\linewidth]{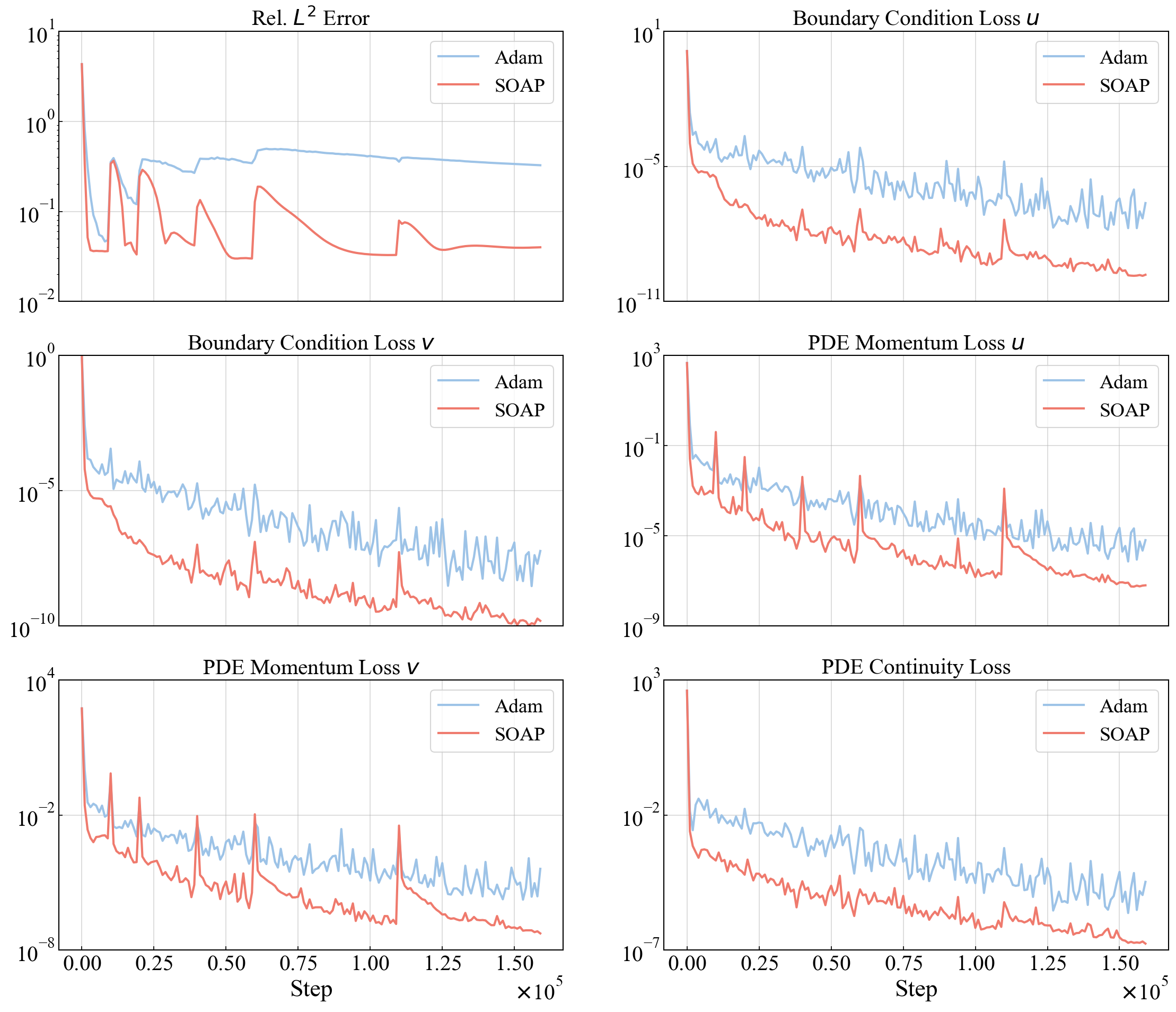}
  \caption{{\em Lid-driven Cavity.} Training loss and test error trajectories for the Adam and SOAP optimizers.}
    \label{fig:ldc_loss}
\end{figure}

\paragraph{Kolmogorov flow.}  We study the two-dimensional Kolmogorov flow governed by the incompressible Navier-Stokes equations:
\begin{align*}
\mathbf{u}_t + \mathbf{u} \cdot \nabla \mathbf{u} & =- \nabla p + \frac{1}{R e} \Delta \mathbf{u} + \mathbf{f}, \\
\nabla \cdot \mathbf{u} & =0,
\end{align*}
on the unit square domain $(x,y) \in [0, 1]^2$.  

Here $\mathbf{u} = (u,v)$ represents the time-varying velocity field, and $\mathbf{f}$ denotes the external forcing term that maintains the flow structure. The system evolves from a random initial state and develops characteristic flow patterns, where energy transfers between different spatial scales through nonlinear interactions and viscous dissipation.

For our study, the system is driven by a sinusoidal forcing $\mathbf{f} =(2 \sin(4 \pi y), 0)$. The numerical experiment initializes with a random initial condition and evolves until $T=2$. The model's performance is quantified by the relative $L^2$ error of vorticity (Table \ref{tab: sota}).



\begin{figure}
  \centering
  \begin{subfigure}[b]{0.45\textwidth}
    \includegraphics[width=\textwidth]{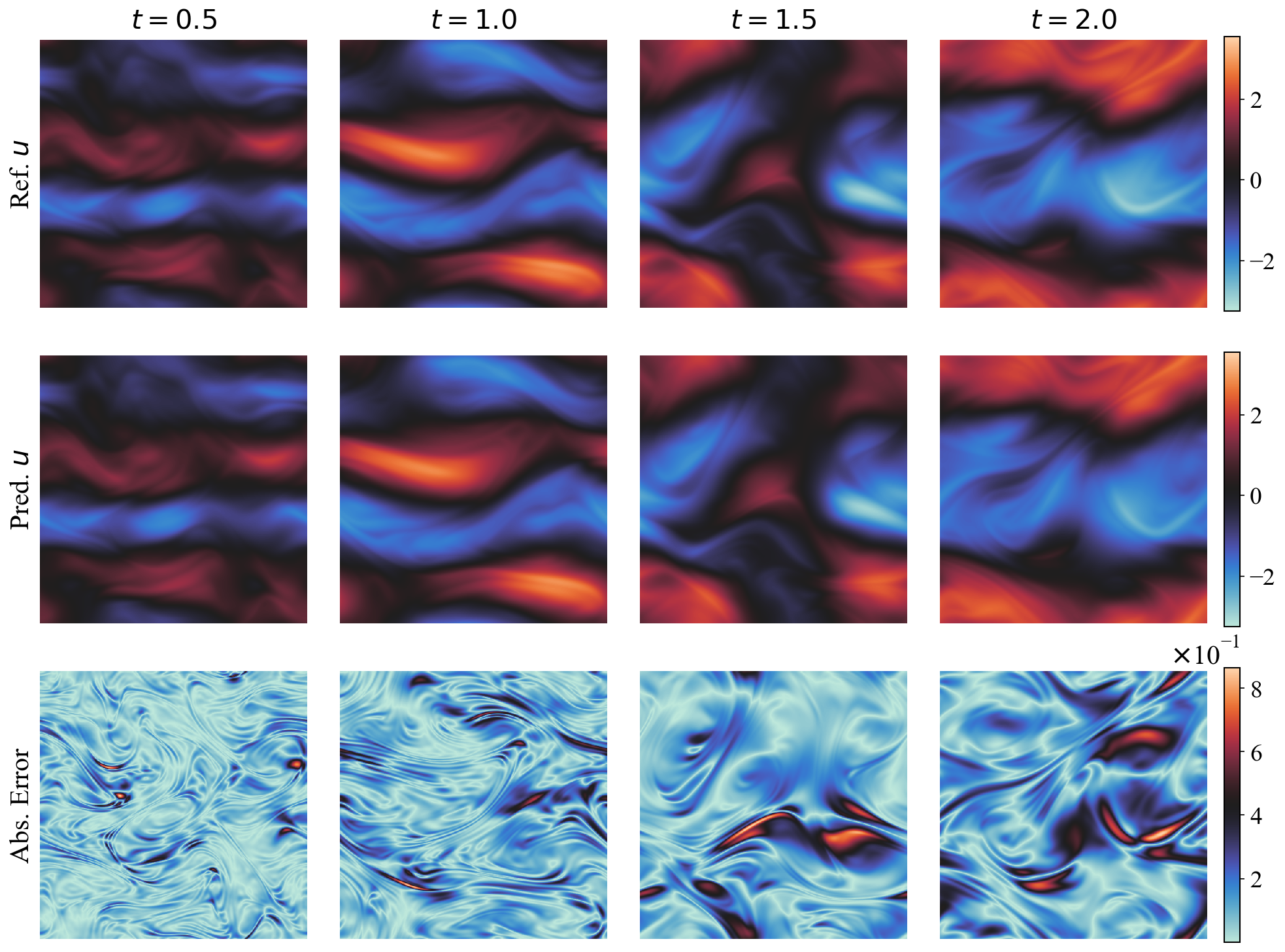}
  \end{subfigure}
  \hspace{2mm}
  \begin{subfigure}[b]{0.45\textwidth}
    \includegraphics[width=\textwidth]{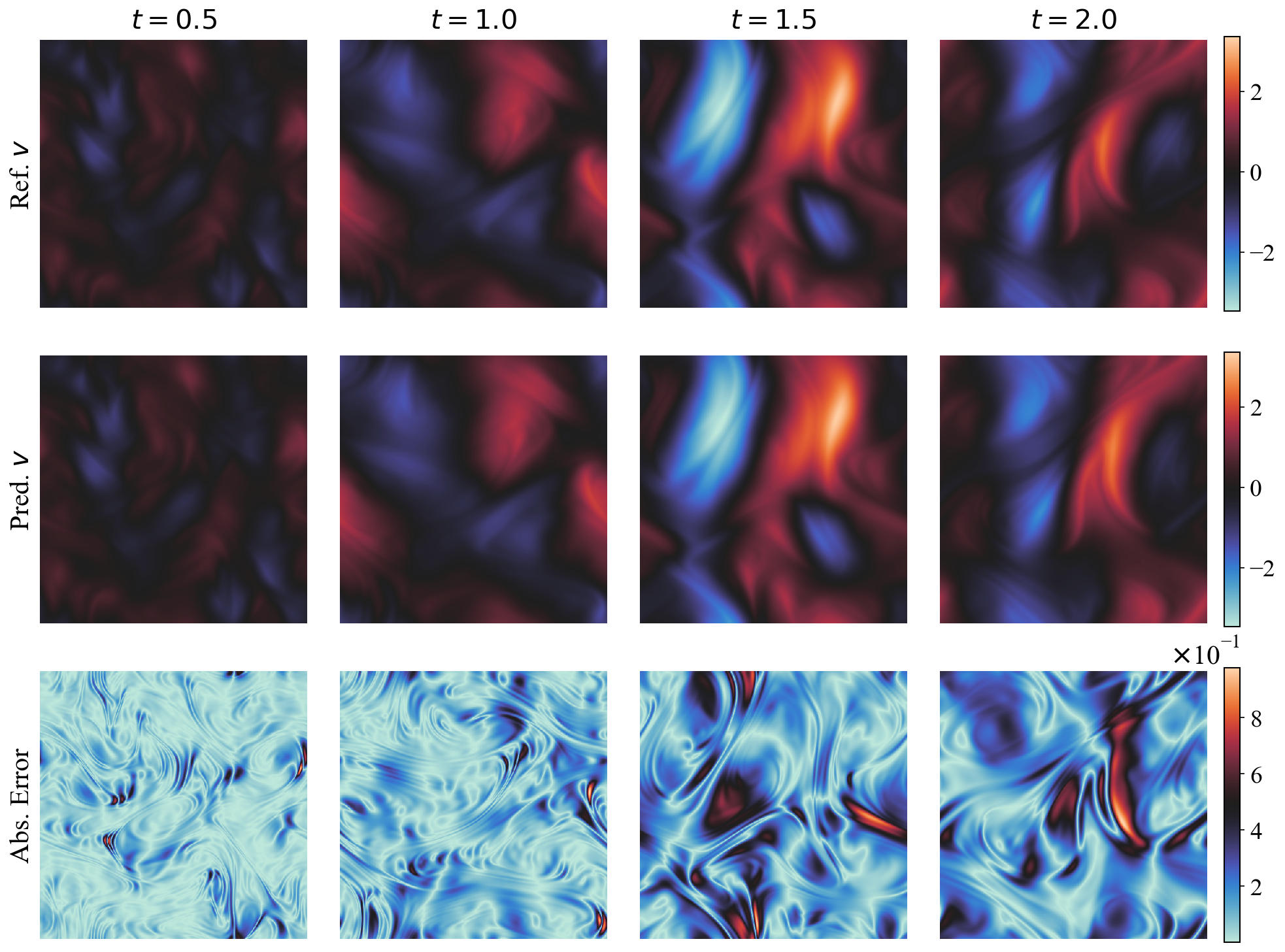}
  \end{subfigure}
\caption{{\em Kolmogorov flow.} Comparison between reference solution and model predictions.}
  \label{fig:ns_tori_uv_preds}
\end{figure}

\begin{figure}
    \centering
    \includegraphics[width=0.7\linewidth]{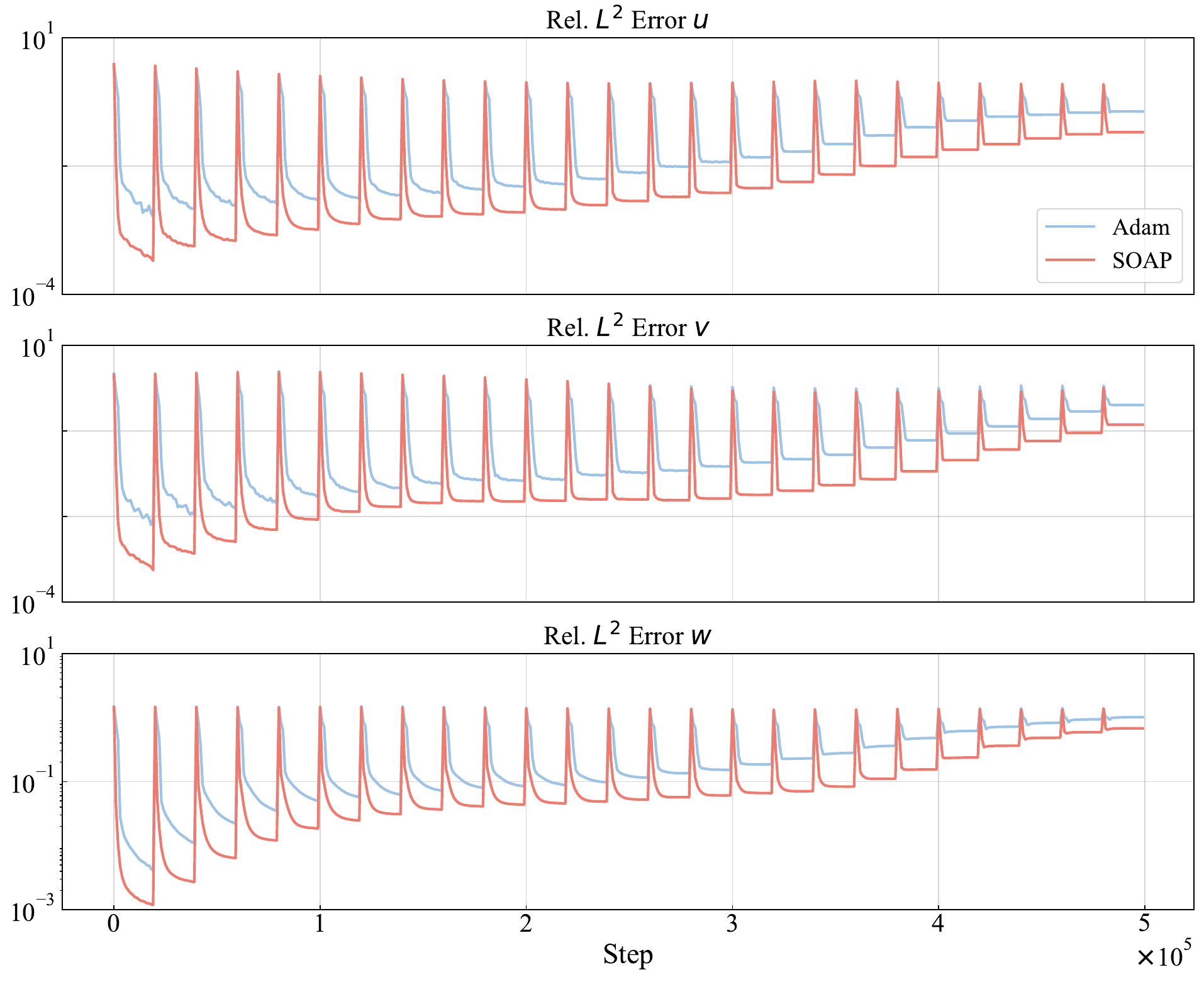}
 \caption{{\em Kolmogorov flow.} Test error trajectories for the Adam and SOAP optimizers.}
    \label{fig:ns_tori_error}
\end{figure}

\begin{figure}
    \centering
    \includegraphics[width=0.7\linewidth]{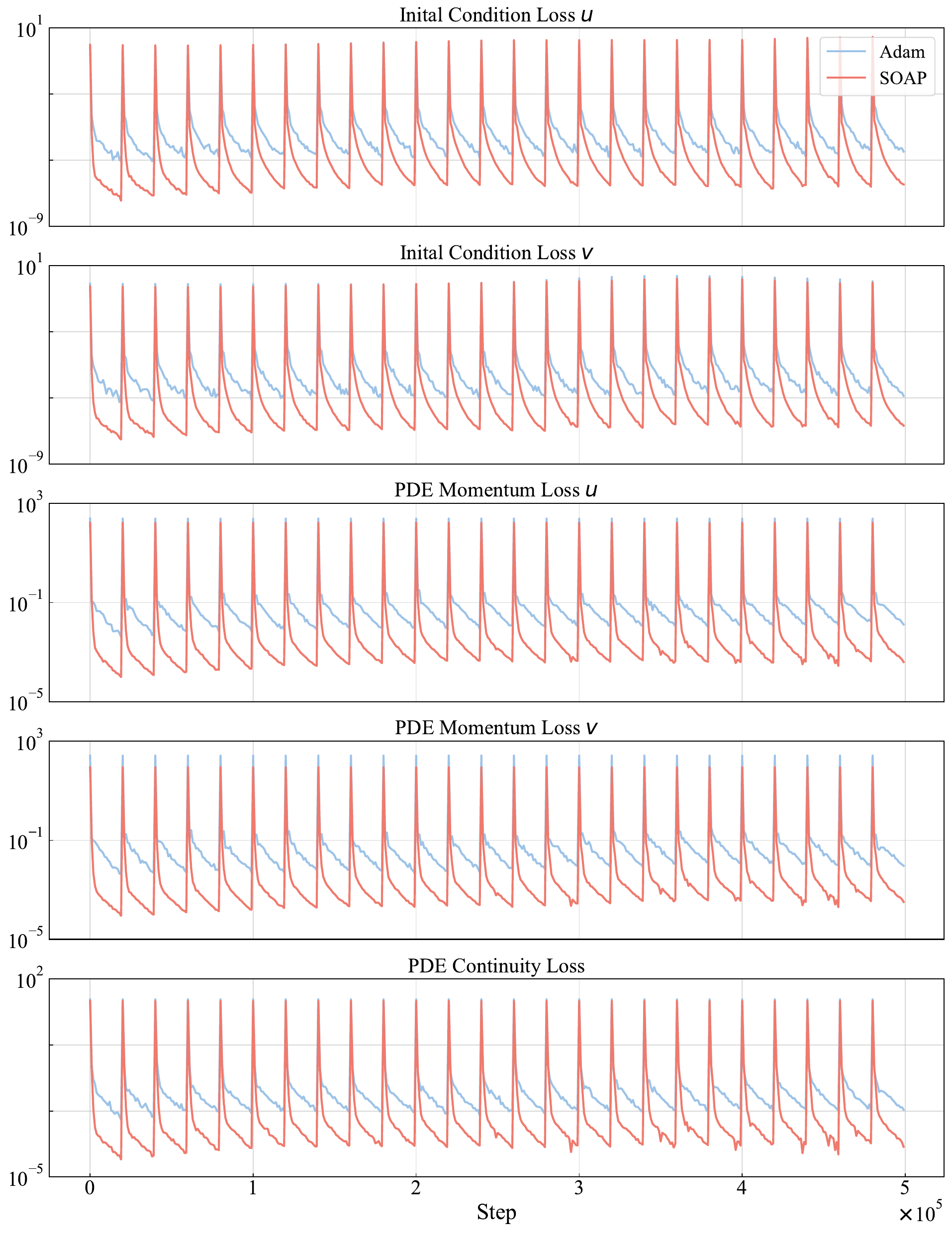}
 \caption{{\em Kolmogorov flow.} Training loss trajectories for the Adam and SOAP optimizers.}
    \label{fig:ns_tori_loss}
\end{figure}

\paragraph{Rayleigh-Taylor instability.} We investigate a coupled flow-temperature system that models buoyancy-driven instability in a rectangular domain $(x, y) \in [0, 1] \times [0, 2]$:
    \begin{align}
        \mathbf{u}_t + \mathbf{u} \cdot \nabla \mathbf{u}  & = - \nabla p +  \sqrt{\frac{Pr}{Ra} }\Delta \mathbf{u} + T \mathbf{e}_y,  \\
        \nabla \cdot \mathbf{u} & =0,  \\
        T_t + \nabla \cdot (\mathbf{u} T)  & =  \frac{1}{\sqrt{Pr Ra}} T_{tt} 
    \end{align}
where $T$ is the temperature field (acting as a density proxy through the Boussinesq approximation),  Pr is the Prandtl number (ratio of momentum to thermal diffusivity), and Ra is the Rayleigh number (measuring buoyancy-driven flow strength). This system captures the characteristic mushroom-shaped plumes that develop as the heavier fluid penetrates into the lighter fluid below.

We set the Prandtl number $\text{Pr}=0.71$ and Rayleigh number $\text{Ra}=10^6$. The boundary conditions are periodic in the horizontal direction for both $\mathbf{u}$ and $T$, with  Dirichlet conditions $\mathbf{u} = T = 0$ imposed on the top and bottom boundaries. The accuracy of our method is evaluated using the temperature field, with results presented in Table \ref{tab: sota}.




\begin{figure}
    \centering
    \includegraphics[width=0.9\linewidth]{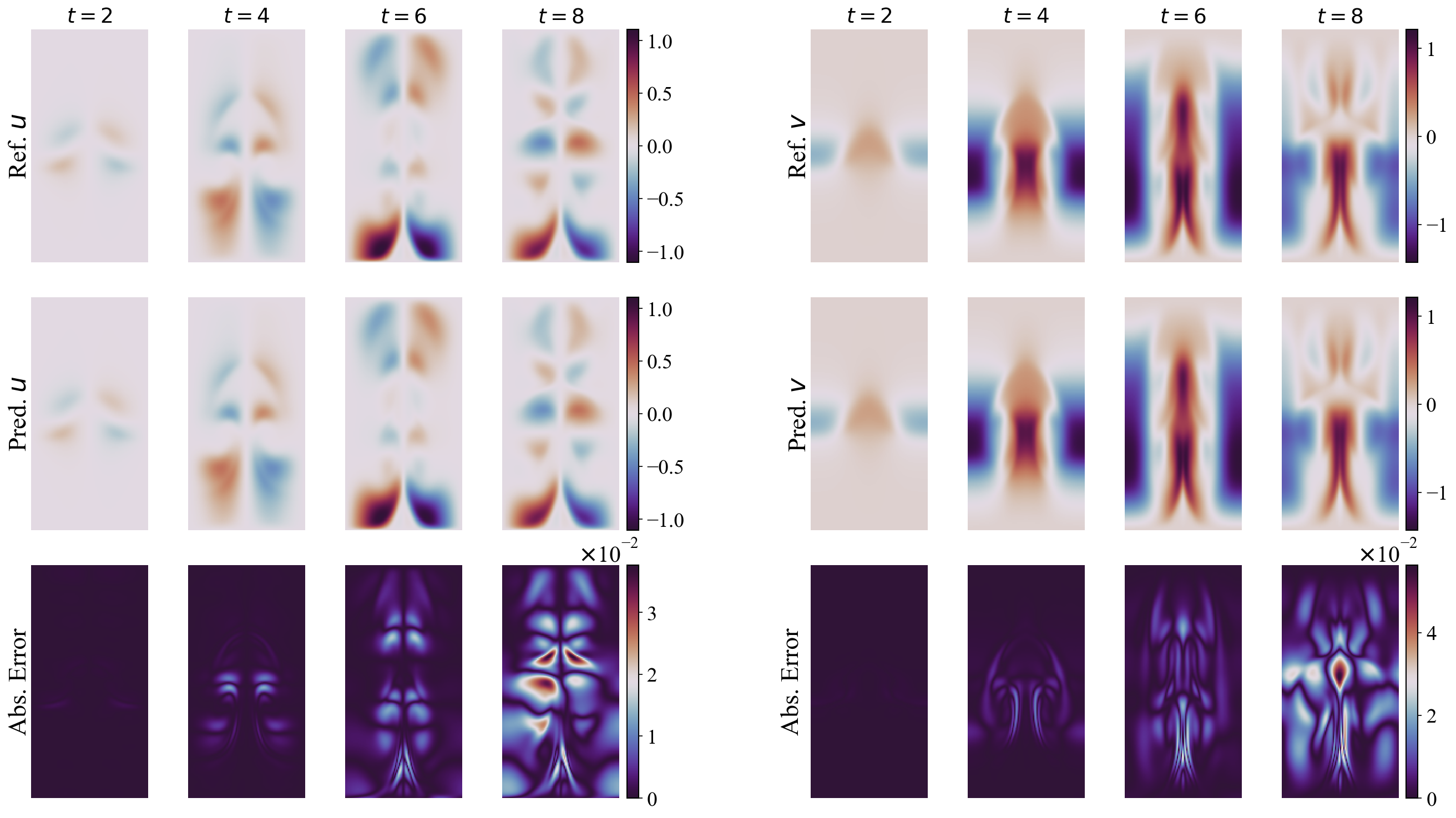}
\caption{{\em Rayleigh-Taylor instability.} Comparison between reference solution and model predictions.}
    \label{fig:rayleigh_taylor_pred_uv}
\end{figure}

\begin{figure}
    \centering
    \includegraphics[width=0.6\linewidth]{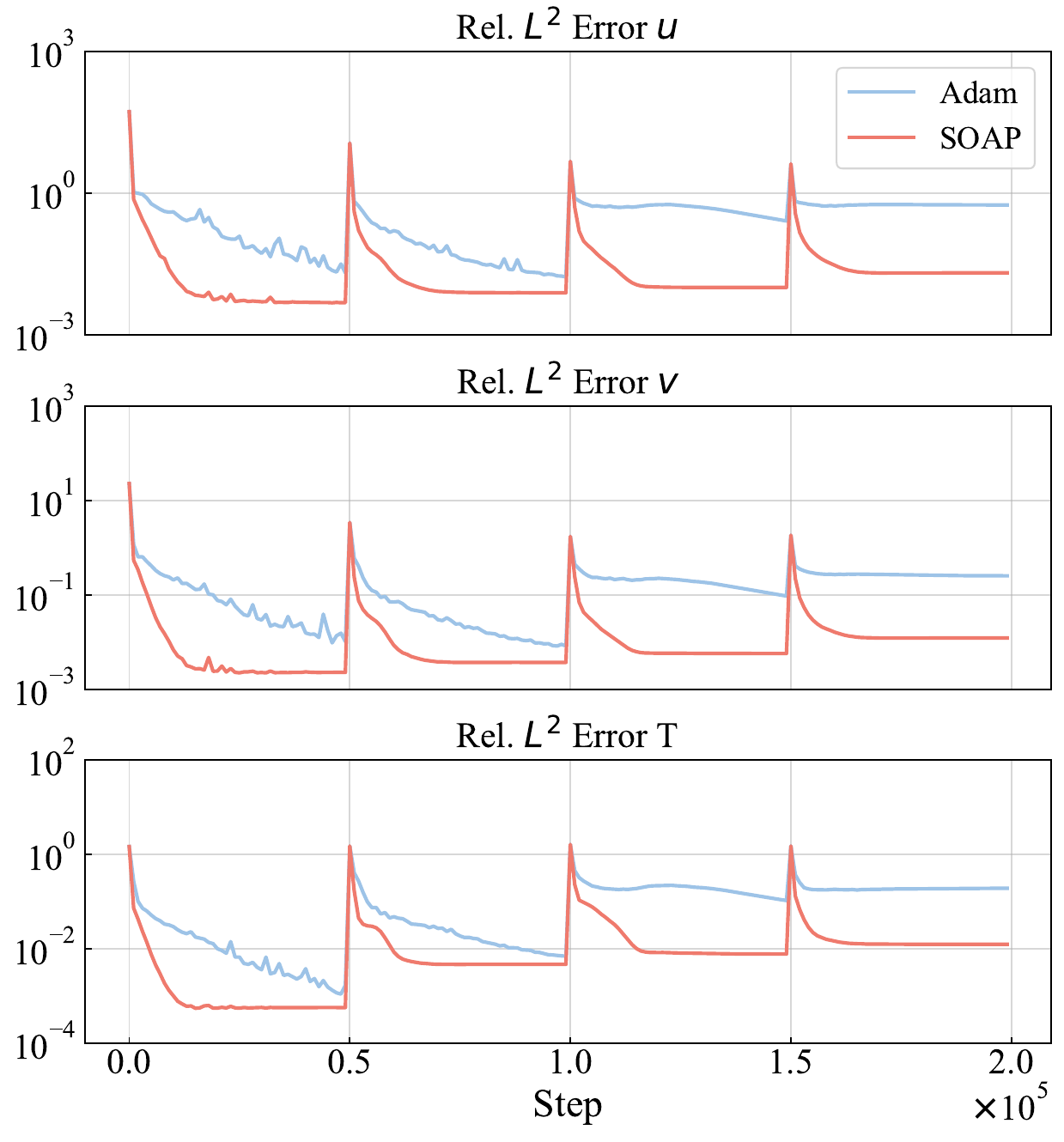}
 \caption{{\em Rayleigh-Taylor instability.} Test error trajectories for the Adam and SOAP optimizers.}
    \label{fig:rayleigh_taylor_error}
\end{figure}

\begin{figure}
    \centering
    \includegraphics[width=0.7\linewidth]{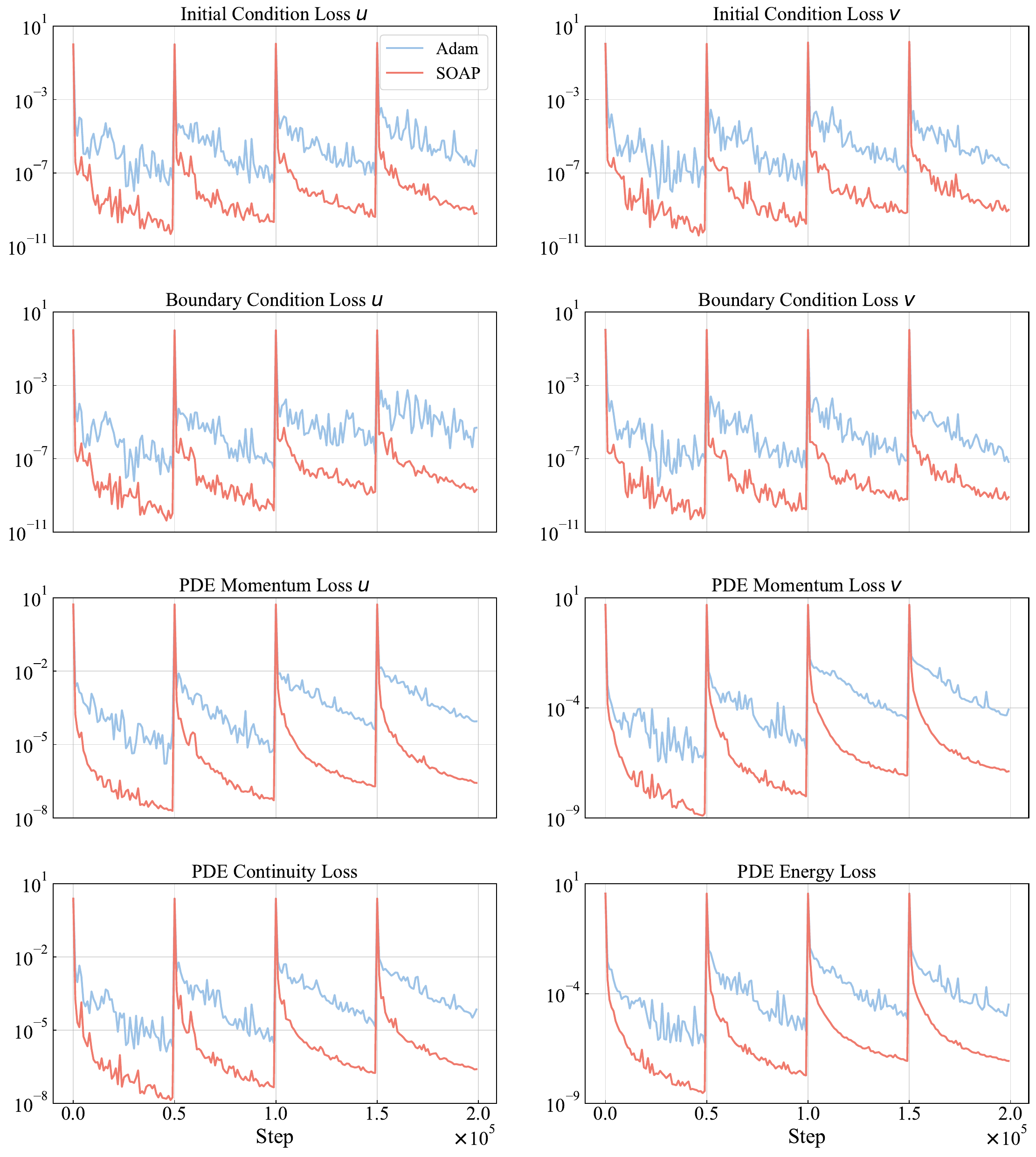}
 \caption{{\em Rayleigh-Taylor instability.} Training loss trajectories for the Adam and SOAP optimizers.}
    \label{fig:rayleigh_taylor_loss}
\end{figure}


\newpage
\section*{NeurIPS Paper Checklist}

\begin{enumerate}

\item {\bf Claims}
    \item[] Question: Do the main claims made in the abstract and introduction accurately reflect the paper's contributions and scope?
    \item[] Answer: \answerYes{} 
    \item[] Justification: The abstract and introduction accurately reflect the paper's contributions, including: introducing a gradient alignment score to quantify directional conflicts (Sec. 1-2), demonstrating how second-order optimizers enhance gradient alignment (Sec. 3), establishing SOAP's connection to Newton's method (Sec. 4), and showing improved performance across 10 PDE benchmarks (Sec. 5).

\item {\bf Limitations}
    \item[] Question: Does the paper discuss the limitations of the work performed by the authors?
    \item[] Answer: \answerYes{} 
    \item[] Justification: The paper discusses computational costs in Section 5 (paragraph "Computational costs"), noting that SOAP requires approximately 2x longer training time compared to baselines, though it achieves better convergence. The conclusion also acknowledges the need for "more efficient preconditioned algorithms that maintain their effectiveness with reduced computational cost.


\item {\bf Theory assumptions and proofs}
    \item[] Question: For each theoretical result, does the paper provide the full set of assumptions and a complete (and correct) proof?
    \item[] Answer: \answerYes{} 
    \item[] Justification: The paper includes theoretical results with clear assumptions and proofs. Proposition 1 in Section 3 establishes the connection between alignment score and cosine similarity. Proposition 2 analyzes gradient alignment at initialization. Proposition 3 connects preconditioned gradient descent to Newton's method. Complete proofs are provided in the appendix (referenced throughout Section 3-4).

    \item {\bf Experimental result reproducibility}
    \item[] Question: Does the paper fully disclose all the information needed to reproduce the main experimental results of the paper to the extent that it affects the main claims and/or conclusions of the paper (regardless of whether the code and data are provided or not)?
    \item[] Answer: \answerYes{} 
    \item[] Justification: The paper provides sufficient information to reproduce the main experimental results.  Section 5 describes the baseline setup, and the appendices (referenced in Section 5) contain complete information on experimental settings, hyperparameters, and data generation.

\item {\bf Open access to data and code}
    \item[] Question: Does the paper provide open access to the data and code, with sufficient instructions to faithfully reproduce the main experimental results, as described in supplemental material?
    \item[] Answer: \answerNA{} 
    \item[] Justification: We will release the code once the paper is accepted.

\item {\bf Experimental setting/details}
    \item[] Question: Does the paper specify all the training and test details (e.g., data splits, hyperparameters, how they were chosen, type of optimizer, etc.) necessary to understand the results?
    \item[] Answer: \answerYes{} 
    \item[] Justification: The paper specifies training and test details in Section 5.

\item {\bf Experiment statistical significance}
    \item[] Question: Does the paper report error bars suitably and correctly defined or other appropriate information about the statistical significance of the experiments?
    \item[] Answer: \answerYes{} 
    \item[] Justification: The paper includes multiple performance comparisons across different PDEs and optimizers, with ablation studies in Figure 4 showing the impact of different hyperparameter settings. The consistent performance improvements across diverse benchmarks (Table 2) and multiple experimental configurations support the statistical significance of the results.

\item {\bf Experiments compute resources}
    \item[] Question: For each experiment, does the paper provide sufficient information on the computer resources (type of compute workers, memory, time of execution) needed to reproduce the experiments?
    \item[] Answer: \answerYes{} 
    \item[] Justification: The paper provides information about computational resources in Appendix G.5.
    
\item {\bf Code of ethics}
    \item[] Question: Does the research conducted in the paper conform, in every respect, with the NeurIPS Code of Ethics \url{https://neurips.cc/public/EthicsGuidelines}?
    \item[] Answer: \answerYes{} 
    \item[] Justification:  The research conforms to the NeurIPS Code of Ethics. It focuses on fundamental advances in scientific machine learning with no apparent ethical concerns. The work acknowledges all related research properly and presents results transparently.

\item {\bf Broader impacts}
    \item[] Question: Does the paper discuss both potential positive societal impacts and negative societal impacts of the work performed?
    \item[] Answer: \answerYes{} 
    \item[] Justification: The conclusion discusses broader positive impacts, noting that the principles of gradient alignment could benefit applications involving competing objectives beyond scientific computing. The research enables more accurate simulation of complex physical systems, which has positive societal impact for scientific and engineering applications.
    As with any tool that furthers our understanding and ability to predict the outcomes of complex systems, there may be ill-intentioned use cases, but we do not expect any specific negative impact from this work.
    
    
\item {\bf Safeguards}
    \item[] Question: Does the paper describe safeguards that have been put in place for responsible release of data or models that have a high risk for misuse (e.g., pretrained language models, image generators, or scraped datasets)?
    \item[] Answer: \answerNA{} 
    \item[] Justification: The paper focuses on optimization techniques for solving PDEs and does not involve models or data with high risk for misuse. The methods presented are used for scientific computing applications without foreseeable harmful applications.

\item {\bf Licenses for existing assets}
    \item[] Question: Are the creators or original owners of assets (e.g., code, data, models), used in the paper, properly credited and are the license and terms of use explicitly mentioned and properly respected?
    \item[] Answer: \answerYes{} 
    \item[] Justification: The paper properly cites existing work and acknowledges software used (JAX, Matplotlib, Chebfun, and NumPy) in the Acknowledgments section. The baseline implementation builds on prior work that is appropriately cited.

\item {\bf New assets}
    \item[] Question: Are new assets introduced in the paper well documented and is the documentation provided alongside the assets?
    \item[] Answer: \answerYes{} 
    \item[] Justification: We will release code implementing the methods described in this paper upon acceptance. 

\item {\bf Crowdsourcing and research with human subjects}
    \item[] Question: For crowdsourcing experiments and research with human subjects, does the paper include the full text of instructions given to participants and screenshots, if applicable, as well as details about compensation (if any)? 
    \item[] Answer: \answerNA{} 
    \item[] Justification: The paper does not involve crowdsourcing or research with human subjects. It focuses on algorithmic improvements and numerical experiments for solving PDEs.

\item {\bf Institutional review board (IRB) approvals or equivalent for research with human subjects}
    \item[] Question: Does the paper describe potential risks incurred by study participants, whether such risks were disclosed to the subjects, and whether Institutional Review Board (IRB) approvals (or an equivalent approval/review based on the requirements of your country or institution) were obtained?
    \item[] Answer: \answerNA{} 
    \item[] Justification: The research does not involve human subjects, so IRB approval is not applicable.

\item {\bf Declaration of LLM usage}
    \item[] Question: Does the paper describe the usage of LLMs if it is an important, original, or non-standard component of the core methods in this research? Note that if the LLM is used only for writing, editing, or formatting purposes and does not impact the core methodology, scientific rigorousness, or originality of the research, declaration is not required.
    \item[] Answer: \answerNA{} 
    \item[] Justification: The paper does not indicate the use of LLMs as part of the core methods or research.

\end{enumerate}

\end{document}